\documentclass{article}
\usepackage{hyphenat}
\usepackage{microtype}
\usepackage{graphicx}
\usepackage{subcaption}
\usepackage{booktabs} 
\usepackage{multirow} 
\usepackage{longtable}
\usepackage{booktabs}
\usepackage{float}
\usepackage{marvosym}
\newenvironment{text}{\ttfamily\small}{} 

\usepackage{hyperref}
\usepackage[ruled,vlined]{algorithm2e}

\usepackage[table]{xcolor}
\usepackage{colortbl}
\definecolor{bestcolor}{RGB}{219, 208, 237}
\definecolor{secondcolor}{RGB}{241, 237, 248}
\definecolor{line-blue}{RGB}{243, 248, 252}
\definecolor{taskSpatial}{RGB}{255, 204, 170}   
\definecolor{taskTransform}{RGB}{210, 190, 230} 
\definecolor{taskLogic}{RGB}{160, 200, 255}     
\definecolor{taskAbstract}{RGB}{245, 170, 185}  
\definecolor{taskPercept}{RGB}{180, 225, 200}   
\definecolor{taskPhysics}{RGB}{30, 60, 120}

\usepackage{xspace}
\newcommand{\name}{VBVR-Pro\xspace}
\newcommand{\benchname}{VBVR-Pro-Bench\xspace}
\newcommand{\dataname}{VBVR-Pro-Dataset\xspace}

\usepackage{booktabs}
\usepackage{multirow}
\usepackage{adjustbox}
\usepackage{pifont}
\usepackage{xcolor}
\newcommand{\cmark}{\ding{51}}
\newcommand{\xmark}{\textcolor{gray!60}{--}}
\usepackage[most]{tcolorbox}

\newtcolorbox{promptbox}[1][]{
    breakable,
    colback=gray!6,
    colframe=gray!25,
    boxrule=0.5pt,
    arc=2pt,
    left=8pt,
    right=8pt,
    top=7pt,
    bottom=7pt,
    before skip=8pt,
    after skip=8pt,
    fontupper=\small\rmfamily,
    #1
}

\usepackage[pr]{icml2026}

\usepackage{amsmath}
\usepackage{amssymb}
\usepackage{mathtools}
\usepackage{amsthm}

\usepackage[capitalize,noabbrev]{cleveref} 

\theoremstyle{plain}

\theoremstyle{definition}

\theoremstyle{remark}

\usepackage[textsize=tiny]{todonotes}
\usepackage{enumitem}

\usepackage{makecell}

\usepackage{wrapfig}
\usepackage[table]{xcolor}
\usepackage{colortbl}
\definecolor{line-blue}{RGB}{243, 248, 252}
\usepackage{tabularx} 
\usepackage{array} 
\usepackage[table]{xcolor} 
\newcolumntype{Y}{>{\raggedright\arraybackslash}X}
\newcolumntype{P}[1]{>{\raggedright\arraybackslash}p{#1}}

\begin{document}
\icmltitlerunning{\name: A Scalable and Verifiable Suite for Native Visual Reasoning}
\twocolumn[
  \icmltitle{
    \name: A Scalable and Verifiable Suite for \\ Native Visual Reasoning
  }



  \icmlsetsymbol{equal}{*}

\icmlsetaffiliationorder{ntu,ucb,ucsd,vbvr,utokyo,cuhk,umich,jhu,davis,ucla,cmu,columbia,toronto,stanford,mila,oxford,INSAIT,bristol,fau,hkust}
\begin{icmlauthorlist}
    \icmlauthor{Junxiang Xu}{equal,ntu}
    \icmlauthor{Ruisi Wang}{equal,ntu}
    \icmlauthor{Fanyi Pu}{equal,ntu}
    \icmlauthor{Maijunxian Wang}{equal,ucb}
    \icmlauthor{Ran Ji}{equal,ucsd}
    \icmlauthor{Tongxi Zhou}{equal,vbvr}

    \icmlauthor{Chenyang Gu}{ntu}
    \icmlauthor{Jing Zuo}{vbvr}
    \icmlauthor{Hongcan Xiao}{vbvr}
    \icmlauthor{Yimeng Geng}{vbvr}
    \icmlauthor{Wanqi Yin}{utokyo}
    \icmlauthor{Wei Chen}{ntu}
    \icmlauthor{Oscar Qian}{ntu}
    \icmlauthor{Zhengan Yan}{ntu}
    \icmlauthor{Ziqi Huang}{ntu}
    \icmlauthor{Haiwen Diao}{ntu}
    \icmlauthor{Liang Pan}{ntu}
    \icmlauthor{Bo Li}{ntu}
    \icmlauthor{Xiangyu Fan}{cuhk}
    \icmlauthor{Dezhi Luo}{umich}
    \icmlauthor{Fengyuan Yu}{ntu}
    \icmlauthor{Zehong Zhao}{ucsd}
    \icmlauthor{Qingying Gao}{jhu}
    \icmlauthor{Tinghui Zhu}{davis}
    \icmlauthor{Yilan Zhang}{ucla}
    \icmlauthor{Jingqi Tong}{cmu}
    \icmlauthor{Pinyuan Feng}{columbia}
    \icmlauthor{Zhengze Jiang}{columbia}
    \icmlauthor{Letian Wang}{toronto}
    \icmlauthor{Ziyu Guo}{cuhk}
    \icmlauthor{Renrui Zhang}{cuhk}
    \icmlauthor{Jieneng Chen}{stanford}
    \icmlauthor{Sonia Joseph}{mila}
    \icmlauthor{Constantin Venhoff}{oxford}
    \icmlauthor{Saman Motamed}{INSAIT}
    \icmlauthor{Mengyue Yang}{bristol}
    \icmlauthor{Chandra Sripada}{umich}
    \icmlauthor{Alan Yuille}{jhu}
    \icmlauthor{Philip Torr}{oxford}
    \icmlauthor{Lvmin Zhang}{stanford}
    \icmlauthor{Vikash Kumar}{cmu}
    \icmlauthor{Daniel Khashabi}{jhu}
    \icmlauthor{Nikolaus Kriegeskorte}{columbia}
    \icmlauthor{Raphaël Millière}{oxford}
    \icmlauthor{Vincent C. Müller}{fau}   
    \icmlauthor{Anyi Rao}{hkust}
    \icmlauthor{Quan Wang}{vbvr}
    \icmlauthor{Ziwei Liu}{ntu}
    \icmlauthor{Dahua Lin}{cuhk}
    \icmlauthor{Lei Yang}{cuhk}
    \icmlauthor{Hokin Deng}{correspondence,cmu}
    \icmlauthor{Zhongang Cai}{correspondence,ntu}
\end{icmlauthorlist}

\icmlaffiliation{ucb}{University of California, Berkeley}
\icmlaffiliation{ntu}{Nanyang Technological University}
\icmlaffiliation{ucsd}{University of California, San Diego} 
\icmlaffiliation{umich}{University of Michigan}
\icmlaffiliation{oxford}{University of Oxford}
\icmlaffiliation{cuhk}{The Chinese University of Hong Kong}
\icmlaffiliation{hkust}{Hong Kong University of Science and Technology}
\icmlaffiliation{stanford}{Stanford University}
\icmlaffiliation{cmu}{Carnegie Mellon University} 
\icmlaffiliation{bristol}{University of Bristol} 
\icmlaffiliation{jhu}{Johns Hopkins University}
\icmlaffiliation{ucla}{University of California, Los Angeles}
\icmlaffiliation{INSAIT}{INSAIT, Sofia University `St. Kliment Ohridski'}
\icmlaffiliation{mila}{Mila - Institut québécois d'IA}
\icmlaffiliation{fau}{Friedrich-Alexander-Universität Erlangen}
\icmlaffiliation{davis}{University of California, Davis}
\icmlaffiliation{toronto}{University of Toronto}
\icmlaffiliation{columbia}{Columbia University}
\icmlaffiliation{vbvr}{VBVR Community Contributors}
\icmlaffiliation{utokyo}{The University of Tokyo}

\newcommand{\icmlaffiliationlayout}{%
  \icmlaffiliationitem{ntu}
  \icmlaffiliationitem{ucb}
  \icmlaffiliationitem{ucsd}
  \icmlaffiliationbreak

  \icmlaffiliationitem{vbvr}
  \icmlaffiliationitem{utokyo}
  \icmlaffiliationitem{cuhk}
  \icmlaffiliationbreak
  
  \icmlaffiliationitem{umich}
  \icmlaffiliationitem{jhu}
  \icmlaffiliationitem{davis}
  \icmlaffiliationbreak

  \icmlaffiliationitem{ucla}
  \icmlaffiliationitem{cmu}
  \icmlaffiliationitem{columbia}
  \icmlaffiliationbreak
  
  \icmlaffiliationitem{toronto}
  \icmlaffiliationitem{stanford}
  \icmlaffiliationitem{mila}
  \icmlaffiliationitem{oxford}
  \icmlaffiliationbreak

  \icmlaffiliationitem{INSAIT}
  \icmlaffiliationitem{bristol}
  \icmlaffiliationitem{fau}
  \icmlaffiliationbreak

  \icmlaffiliationitem{hkust}
}



\icmlcorrespondingauthor{Zhongang Cai}{caiz0023@e.ntu.edu.sg}
\icmlcorrespondingauthor{Hokin Deng}{hokind@andrew.cmu.edu}

\icmlkeywords{Visual Reasoning, Video Reasoning, Image Generation, Video Generation, Interleaved Reasoning}
]



\printAffiliationsAndNotice{\icmlEqualContribution}

\begin{figure*}[!h]
\vspace{8pt}
  \centering
  \label{fig:vbvr_pro_teaser}
  \includegraphics[width=0.98\textwidth]{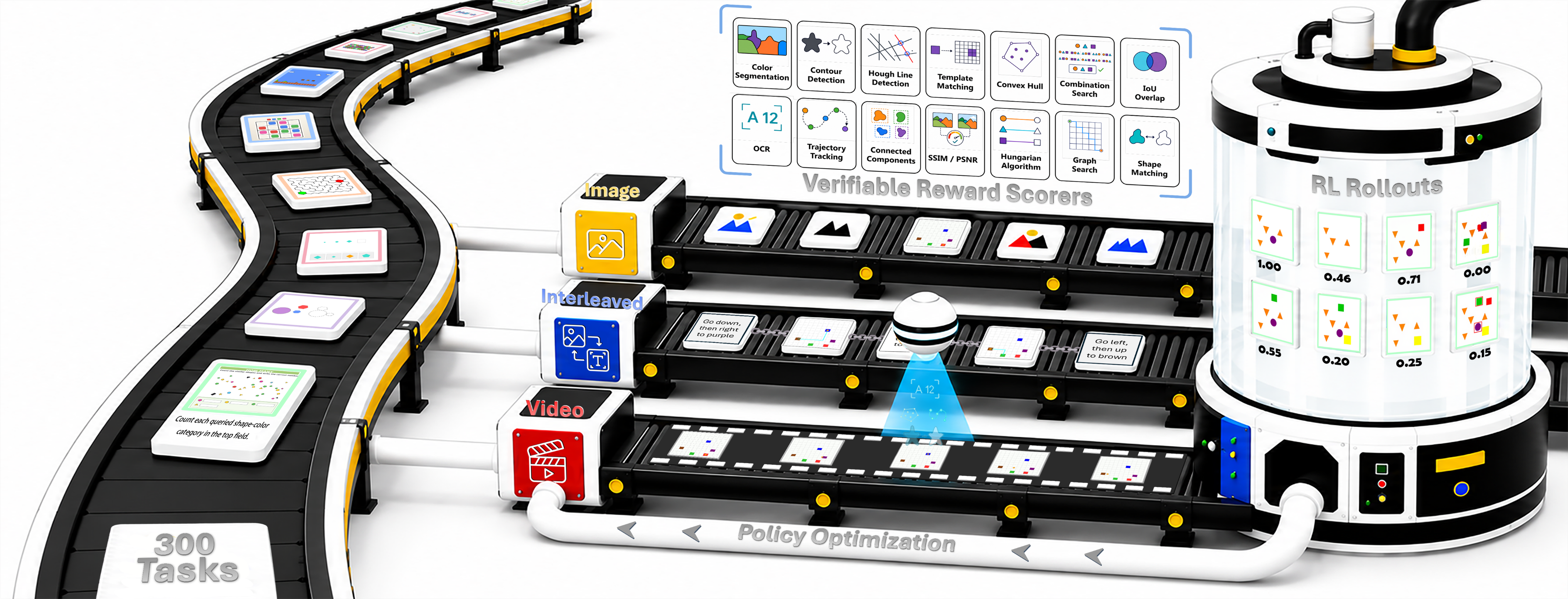}
  \caption{
    \textbf{Overview of \name.} \name establishes a closed-loop testbed for native visual reasoning, providing \textit{300} visual reasoning tasks that support controlled studies of image, interleaved, and video generators under a unified task distribution. Its \textit{verifiable reward scorers} provide reproducible, human-aligned evaluation and further serve as task-grounded reward functions for large-scale multi-task reinforcement learning. 
  }
\end{figure*}
\begin{abstract}

Native visual reasoning treats visual generation as the medium of reasoning itself: visual states (\ie images and videos) are not merely inputs to be understood or outputs to be rendered, but first-class substrates for problem solving beyond language. Yet progress remains bottlenecked by the lack of scalable training tasks, reliable feedback, and controlled comparisons across generative substrates. In this work, we introduce \textbf{\name}, a closed-loop testbed that makes native visual reasoning through generation trainable, verifiable, optimizable, and experimentally controllable.
\textbf{1) Task scaling.} \name turns visual reasoning into a controlled task space of \textit{300} procedurally generated tasks. Models trained on \name show strong transfer beyond the proposed suite across \textit{seven} external visual reasoning benchmarks such as RISE-Video, MME-CoF-Pro, and BabyVision. Further analysis validates that these gains reflect visual reasoning rather than instruction-pattern fitting.
\textbf{2) Verifiable rewards.} \name provides verifiable reward scorers for task-grounded evaluation. Through a systematic study of leading MLLMs as judges, we identify recurring failure modes of the prevalent \textit{VLM-as-a-judge} paradigm. In contrast, the proposed scorers are grounded in deterministic, task-specific rules, achieve fine-grained alignment with human judgments. Importantly, they serve as reliable reward signals for large-scale multi-task reinforcement learning and demonstrate stronger post-RL performance across visual reasoning tasks.
\textbf{3) Mechanism study.} \name enables controlled modality studies across more than \textit{30} image, video, and interleaved generators. Our analysis shows that video generation remains strongest for tasks requiring persistent spatiotemporal state tracking, while interleaved generation provides a compute-efficient alternative by externalizing intermediate visual states. Critically, ablations and probing suggest the presence of vision-native trajectories that are a more crucial substrate than explicit linguistic chains of thought for visual reasoning. We release all data, models, scorers, and code to facilitate future research.
\end{abstract}

\section{Introduction}

The success of large language models has made language the dominant substrate for studying machine reasoning. Yet many forms of intelligence in the physical world are not naturally linguistic: they involve spatial transformations, temporal continuity, object persistence, and dynamic interactions. Recent studies suggest that such reasoning may be better exposed through generation~\cite{he2025diffthinker, wang2026demystifying}, motivating a new paradigm of \textit{native visual reasoning}: models solve problems by constructing, updating, and optimizing visual states (\eg, images and videos), where visual states serve as \textit{first-class} substrates for reasoning, rather than merely inputs to a language-based reasoner or final rendered outputs.
Despite growing interest, native visual reasoning remains difficult to study systematically at the current stage: it remains unclear whether visual reasoning through generation can be trained at scale, reliably evaluated, optimized with reward feedback, or compared across different generative substrates.
We argue that progress requires a closed-loop suite: tasks must be procedurally generated, outputs must be verifiably scored, models must be optimized with reliable rewards, and modality choices must be tested under controlled conditions.
In this work, we introduce \name as the infrastructure that makes native visual reasoning through generation trainable, verifiable, optimizable, and experimentally controllable.

First, native visual reasoning lacks scalable and trainable task sources. Existing benchmarks cover diverse capabilities but are primarily designed for evaluation and often provide little or no training data. Conversely, existing large-scale synthetic sources~\cite{vbvr2026} tend to focus on simplified symbolic settings, leaving open whether they teach transferable visual reasoning skills.
\name turns visual reasoning through generation into a controlled task space. Each task is implemented as a procedural generator with randomized parameters. Importantly, multiple aligned modalities, including videos, keyframe images, and interleaved textual annotations, are generated simultaneously to enable fairer comparison among image generators, video generators, and interleaved generators trained from the same underlying task distribution. Metadata needed for verifiable evaluation are also recorded.
This yields 300 tasks and the resulting training data demonstrates strong transferability beyond the proposed suite. In our capability transfer study, across image, video, and interleaved visual reasoners, models trained on \name obtain consistent gains (often more than 20 percentage points) across a wide range of held-out visual reasoning benchmarks, including RISE-Video~\cite{risebench_zhao2026envisioning}, V-ReasonBench~\cite{v-reasonbench_luo2025v}, RULER-Bench~\cite{ruler-bench_he2025ruler}, MME-CoF-Pro~\cite{mme-cof-pro_qi2026mme}, VideoThinkBench~\cite{videothinkbench_tong2026thinking}, BabyVision~\cite{babyvision_chen2026babyvision}, and IntelligentVBench~\cite{intelligentvbench_pan2026omniweaving}.
This shows that broad coverage of visual reasoning mechanisms is more important for transfer than simply increasing the number of instances from a narrower set of tasks.
A central concern, however, is whether such gains reflect genuine visual reasoning or merely instruction-pattern fitting from larger synthetic data. We therefore conduct a suite of diagnostic analyses. (1) nearest-neighbor analyses in visual and textual embedding spaces show that held-out benchmark cases are not explained by close matches to training samples. (2) same-prompt visual counterfactuals reveal that models trained on \name respond correctly to changes in visual states even when the instruction remains fixed. (3) the model demonstrates visual reasoning behaviors (\eg, multi-path exploration~\cite{wang2026demystifying}) that suggest that the transfer gains arise from learning reusable visual reasoning operations rather than fitting superficial instruction templates. Together, these diagnostics suggest that \name provides a scalable curriculum rather than merely an enlarged task collection.

Second, native visual reasoning requires reliable feedback. The prevalent \textit{VLM-as-a-judge} paradigm is attractive for open-ended outputs but is often unreliable for visual reasoning, where correctness depends on exact counts, fine-grained spatial relations, temporal consistency, and rule satisfaction.
We systematically analyze leading MLLMs (\eg, GPT-5.5~\cite{gpt55}, Qwen-3.7-plus~\cite{qwen37}, and Gemini-3.1-Pro~\cite{gemini31pro}) as judges and identify recurring failure modes, including numerical imprecision, neglect of fine-grained evidence, and misunderstanding of task rules.
To address this, \name pairs evaluated tasks with verifiable reward scorers based on structured semantic extraction and task-grounded verification. Unlike prior model-level validation based on aggregate rank correlations, we evaluate reward quality at the instance level through an arena-style human study. Under this stricter protocol, our scorers achieve strong agreement with human judgments and substantially outperform VLM-based judges in per-instance accuracy. 
Crucially, these scorers enable optimization rather than evaluation alone: they  offer task-grounded and learning signals that are unambiguous and directionally meaningful. Experiments show that using them as rewards yields steady improvements even on strong baselines, demonstrating that verifiable rewards can serve as a practical foundation for multi-task reinforcement learning in large-scale visual reasoning.

Third, it remains unclear which generative substrate best supports visual reasoning. Video generation offers continuous temporal trajectories, image generation offers efficiency in condensed single visual output, and interleaved generation offers integrated visual-textual reasoning~\cite{gu2025thinkmorph,zebracot_li2025zebra}. However, these paradigms have rarely been trained and evaluated under a shared task distribution and verification protocol.
With the same task suite and scorers, \name enables controlled comparison across more than 30 image, video, and interleaved generators, varying scale, output format, and modality configuration.
Our results show that video generation remains strongest for tasks requiring persistent spatiotemporal state tracking, while interleaved generation offers a compute-efficient alternative by externalizing intermediate visual states. In contrast, image-only generation often lacks the capacity to express procedural or temporal reasoning. 
Moreover, degrading language reasoning in interleaved models has limited effect compared with removing visual states~\cite{xu2026visualplanning}. Further middle-state intervention experiments show that intermediate visual states are causally used by the model: corrupting or replacing them can predictably degrade or redirect the final answer. These results suggest that visual trajectories, rather than explicit linguistic chains of thought, form a critical substrate for native visual reasoning.

In summary, \name establishes a closed-loop foundation for scaling native visual reasoning through generation. It provides: 1) a scalable curriculum of procedurally generated visual reasoning environments; 2) verifiable reward scorers that are human-aligned and usable for reinforcement learning; 3) a modality-controlled model suite spanning image, video, and interleaved generators, with diagnostic evidence, validating visual-state reasoning. Together, these components make native visual reasoning trainable, verifiable, optimizable, and experimentally controllable.

\section{\name-Dataset: Task Scaling for Transferability}
\label{sec:dataset}

\begin{figure}[t!]
  \centering
  \includegraphics[width=\columnwidth]{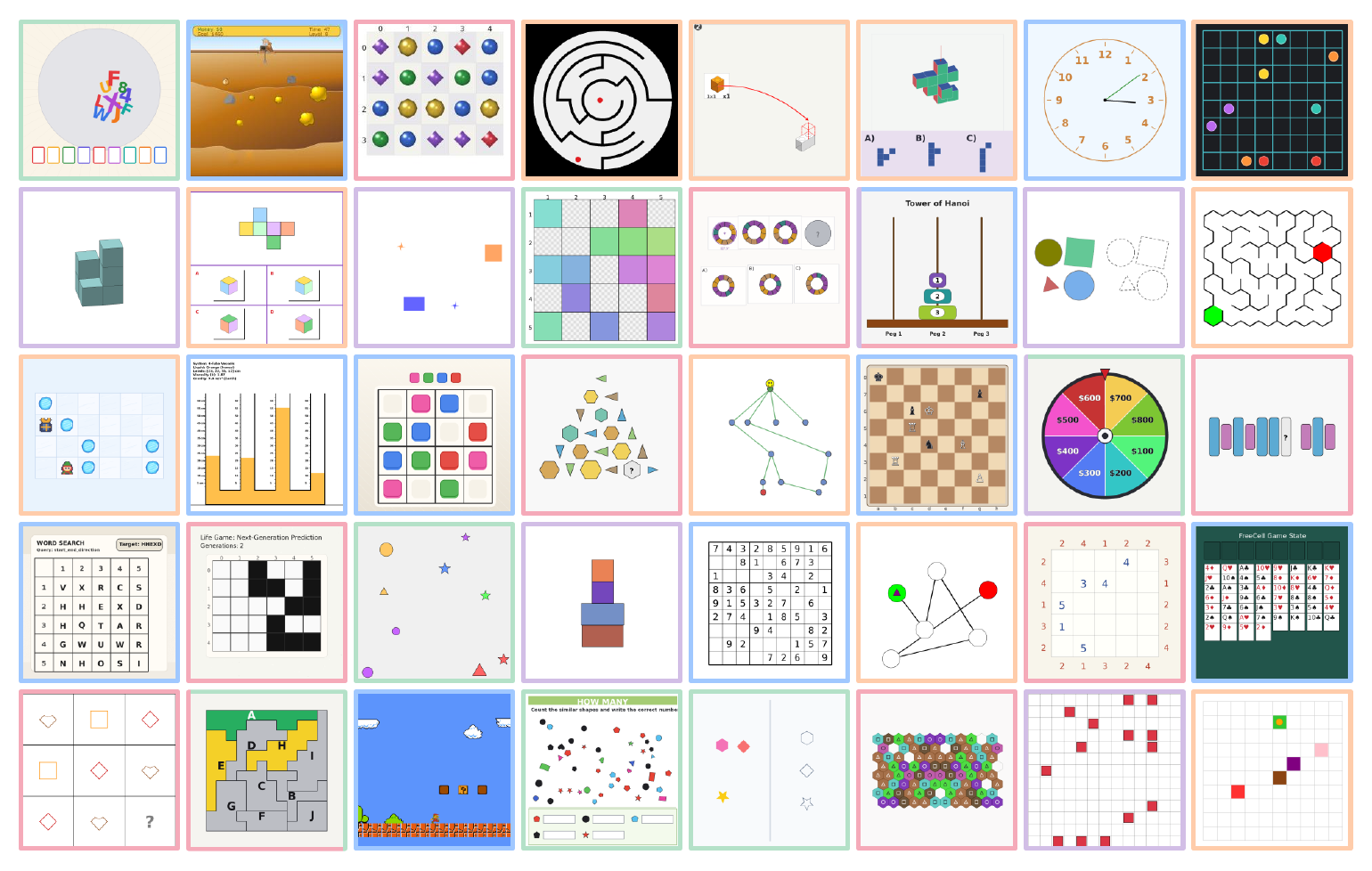}
  \caption{\textbf{Overview of \name tasks.} We show representative tasks spanning our five cognitive faculties, each outlined by the color of its underlying capability: \textcolor{taskPercept}{\textbf{Perception}}, \textcolor{taskTransform}{\textbf{Transformation}}, \textcolor{taskSpatial}{\textbf{Spatiality}}, \textcolor{taskAbstract}{\textbf{Abstraction}}, and \textcolor{taskLogic}{\textbf{Knowledge}}. Tasks requiring multiple capabilities are outlined with multiple colors.
  }
  \label{fig:data_overview}
\end{figure}

\begin{table}[t]
\centering
\renewcommand{\arraystretch}{1.08}
\setlength{\tabcolsep}{3pt}
\caption{\textbf{Comparison of \name with existing visual reasoning benchmarks and datasets.} We report task coverage, data scale, split composition, supported modalities, and evaluation protocols for assessing generated outputs. \textit{VLM} denotes fully relying on VLM judges, \textit{Mixed} primarily relies on VLM judgment, with programmatic checks limited to selected tasks or output aspects, and \textit{Verifiable} uses task-specific verifiable scorers for all tasks. \name leads in task numbers and data scale, provides dedicated data split for RL research, supports video, image, and interleaved generators, and adopts fully verifiable evaluation.}
\label{tab:datasets_comparison}
\begin{adjustbox}{max width=\textwidth}
\begin{tabular}{lrrrrrrcccc}
\toprule
\multirow{2}{*}{\textbf{Dataset}} & \multirow{2}{*}{\textbf{\#Tasks}}
& \multirow{2}{*}{\textbf{\#Images}} & \multirow{2}{*}{\textbf{\#Videos}}
& \multicolumn{3}{c}{\textbf{Split}}
& \multicolumn{3}{c}{\textbf{Modality}}
& \multirow{2}{*}{\textbf{Eval}} \\
\cmidrule(lr){5-7} \cmidrule(lr){8-10}
& & & & \textbf{\#SFT} & \textbf{\#RL} & \textbf{\#Test}
& \textbf{Image} & \textbf{Interleaved} & \textbf{Video} & \\
\midrule
VideoThinkBench~\cite{videothinkbench_tong2026thinking} & 48 & 6,845 & 0 & 0 & 0 & 4,149 & \xmark & \xmark & \cmark & VLM \\
V-ReasonBench~\cite{v-reasonbench_luo2025v}         & 13 & 652 & 0 & 0 & 0 & 326 & \xmark & \xmark & \cmark & Mixed \\
Ruler-Bench~\cite{ruler-bench_he2025ruler}          & 40 & 493 & 0 & 0 & 0 & 622 & \xmark & \xmark & \cmark & VLM \\
RISE-Video~\cite{risevideo_liu2026rise}     & 8 & 477 & 0 & 0 & 0 & 467 & \xmark & \xmark & \cmark & VLM \\
MME-CoF-Pro~\cite{mme-cof-pro_qi2026mme}            & 16 & 370 & 0 & 0 & 0 & 303 & \xmark & \xmark & \cmark & VLM \\
IntelligentVBench~\cite{intelligentvbench_pan2026omniweaving} & 4 & 1,694 & 210 & 0 & 0 & 1,030 & \xmark & \xmark & \cmark & VLM \\
VGI-Bench~\cite{vgibench_he2026vgi}             & 27 & -- & 0 & 0 & 0 & 810 & \xmark & \xmark & \cmark & VLM \\
RISEBench~\cite{risebench_zhao2026envisioning}      & 16 & 415 & 0 & 0 & 0 & 360 & \cmark & \xmark & \xmark & VLM \\
KRIS-Bench~\cite{kris_wu2026kris}                   & 22 & 1,670 & 0 & 0 & 0 & 1,267 & \cmark & \xmark & \xmark & VLM \\
BabyVision-Gen~\cite{babyvision_chen2026babyvision} & 21 & 560 & 0 & 0 & 0 & 280 & \cmark & \xmark & \xmark & VLM \\
GRADE~\cite{grade_liu2026grade}                     & 10 & 1,040 & 0 & 0 & 0 & 520 & \cmark & \xmark & \xmark & VLM \\
OpenING~\cite{opening_zhou2025opening}              & 56 & 17,603 & 0 & 3,240 & 0 & 2,160 & \xmark & \cmark & \xmark & VLM \\
Zebra-CoT~\cite{zebracot_li2025zebra}               & 18 & 921,039 & 0 & 182,384 & 0 & 0 & \xmark & \cmark & \xmark & -- \\
StructCoT~\cite{structcot_wang2026deltav}           & 44 & -- & 0 & 840,000 & 0 & 5,600 & \xmark & \cmark & \xmark & Mixed \\
RealUnify~\cite{realunify_shi2026realunify}         & 32 & 587 & 0 & 0 & 0 & 1,000 & \xmark & \cmark & \xmark & VLM \\
Uni-MMMU~\cite{uni-mmmu_zou2026uni}                 & 8 & 3,544 & 0 & 0 & 0 & 885 & \xmark & \cmark & \xmark & Mixed \\
ThinkMorph~\cite{gu2025thinkmorph}       & 4 & 50,780 & 0 & 24,990 & 0 & 400 & \xmark & \cmark & \xmark & Mixed \\
MIRA~\cite{mira_zhou2026visualizing}                & 20 & 1,482 & 0 & 0 & 0 & 546 & \xmark & \cmark & \xmark & Mixed \\
VBVR~\cite{vbvr2026}                        & 150 & 2,015,000 & 1,007,500 & 1,000,000 & 0 & 500 & \xmark & \xmark & \cmark & Verifiable \\
\midrule
\textbf{\name} & \textbf{300} & \textbf{3,471,558} & \textbf{1,300,500} & \textbf{1,250,000} & \textbf{50,000} & \textbf{500} & \cmark & \cmark & \cmark & \textbf{Verifiable} \\
\bottomrule
\end{tabular}
\end{adjustbox}
\end{table}

A central challenge in visual reasoning is the scarcity of large-scale, diverse training data that transfers beyond the patterns seen during training. Prior work~\cite{vbvr2026} has shown that large-scale synthetic supervision can improve video reasoning, but its transfer to broader downstream visual reasoning benchmarks remains limited. We hypothesize that an important bottleneck is task coverage rather than instance count alone: repeatedly sampling from a fixed set of task generators creates many variations of similar reasoning patterns, but does not substantially expand the range of capabilities being trained.
Motivated by this observation, we construct \dataname, a large-scale synthetic dataset with 300 tasks spanning diverse visual environments and reasoning capabilities (\cref{sec:dataset:tasks}).
We then describe the implementation and curation protocol used to build \name (\cref{sec:dataset:implementation}).
Finally, we analyze \name against existing visual reasoning resources, highlighting its scale, modality coverage, and task diversity (\cref{sec:dataset:analysis}).

\subsection{Task Design}
\label{sec:dataset:tasks}

\name{} is built through a large-scale collaborative effort involving more than 50 researchers and engineers worldwide. Rather than aiming to enumerate isolated tasks, we seek to construct a broad and trainable task space that covers visual reasoning capabilities of practical interest to the community.

\paragraph{Taxonomy.}
We adapt the cognitive framework of VBVR~\cite{vbvr2026}, which organizes visual reasoning into five faculties: \textit{perception}, \textit{spatiality}, \textit{transformation}, \textit{abstraction}, and \textit{knowledge}. \emph{Perception} extracts information directly from visual inputs, covering OCR, symbol recognition, counting, sorting, and visual comparison. \emph{Spatiality} reasons about locations and geometric relationships, including relative positions, 3D structures, and navigation. \emph{Transformation} manipulates visual representations through operations such as 2D translation and rotation. \emph{Abstraction} infers categories or rules from observations, including pattern induction, symmetry and shape completion, and reasoning under complex or newly specified rules. Finally, \emph{knowledge} draws on intrinsic or acquired knowledge, such as conventional icon meanings, physical phenomena, common sense, rules and strategies of classic games, and the mechanics of everyday objects.
Recognizing that a comprehensive task could demonstrate multiple capabilities, tasks in \name can carry several labels.

\paragraph{Design Principles.}
Each task is designed according to four principles. First, the core reasoning process should be vision-native: the solution should depend on visual state transformations, spatial relations, or dynamics rather than purely linguistic deduction. Second, task outcomes should be unambiguous and rigorously inspectable; prompts and visual inputs are iteratively refined through human inspection of model responses. Third, tasks should exhibit substantial visual diversity through randomized appearances, layouts, object attributes, scene configurations, and paraphrased textual prompts. Fourth, contributors are encouraged to design non-trivial tasks that require multi-step reasoning rather than single-step perception or pattern matching.

\subsection{Implementation}
\label{sec:dataset:implementation}

\paragraph{Generators.}
\name is built from 300 task-specific generators under a unified framework: 150 are reimplemented and revised from VBVR~\cite{vbvr2026}, and 150 are newly designed to broaden task coverage. Each generator is a parameterized program that samples a configuration from a structured parameter space, such as grid size, object count, layout, visual appearance, and difficulty level, and then instantiates a corresponding problem instance. A task-specific solver computes the correct solution, providing programmatically derived supervision without manual annotation.

\begin{figure}[t]
  \centering
  \includegraphics[width=\columnwidth]{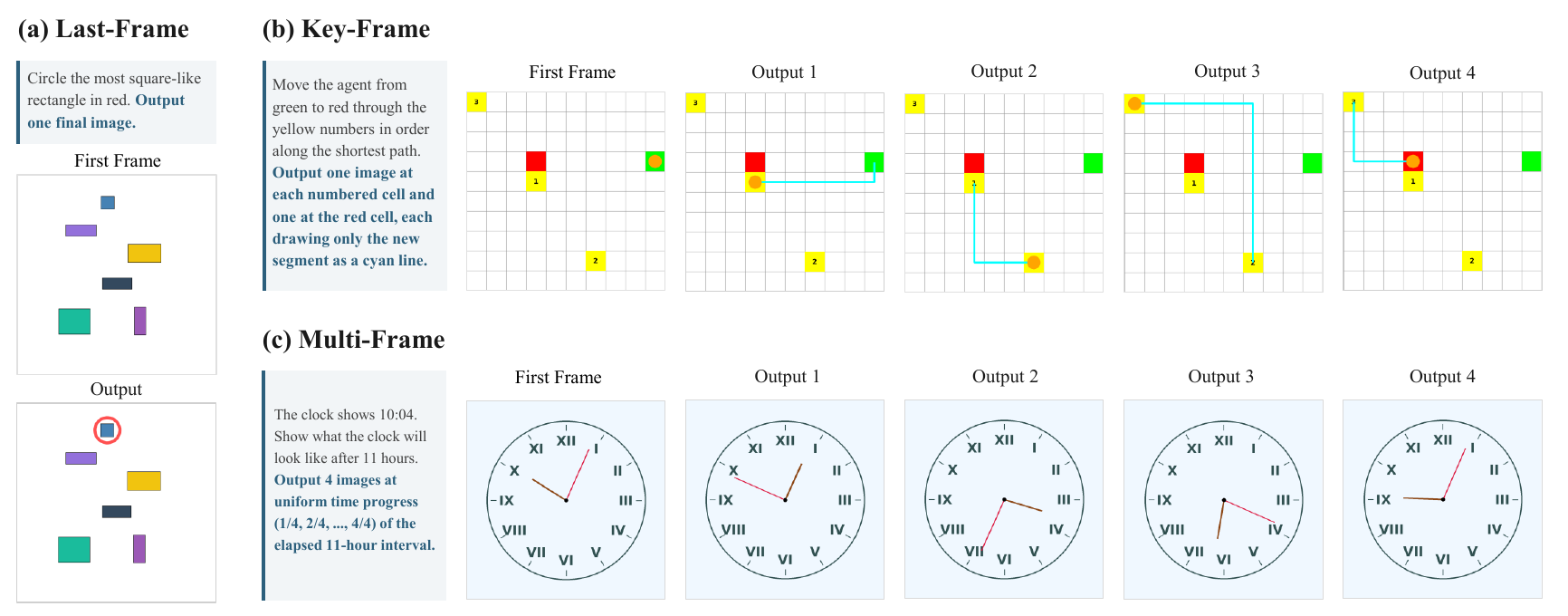}
  \caption{
    \textbf{Three output regimes for the image and interleaved modalities.} Intermediate text annotation is omitted here for clarity. \textit{Last-Frame} is used when the final answer alone is sufficient for evaluation; \textit{Key-Frame} is used for more complex tasks, such as multi-step planning in maze navigation, where the solution depends on a few essential intermediate frames; and \textit{Multi-Frame} is used when the solution process itself matters, such as temporal continuity or per-frame validity.
  }
  \label{fig:image_modality_taxonomy}
\end{figure}

\paragraph{Metadata.}
During generation, each solved instance is accompanied by a metadata file containing its random seed, problem specification, complete solution, and key element attributes. This information provides structured ground truth for the verifiable reward scorers introduced in~\cref{sec:4_verifiable_reward_scorer}. It is also used for de-duplication and split construction, ensuring a clear separation between training and evaluation instances.

\paragraph{Multi-modality.}
A key feature of \name is that each instance is rendered into aligned visual modalities, including video and image. Constructing the image modality requires more than simply sampling frames from the video: different tasks require different levels of trajectory information for the answer to remain complete.
As illustrated in \cref{fig:image_modality_taxonomy}, the image modality adopts three output regimes. \emph{Last-Frame} shows only the final state for tasks whose solutions can be represented by a single image, such as selecting the correct option. \emph{Key-Frame} presents selected states that capture the essential transitions in the solution process. For path-based tasks such as maze navigation, the complete solution trajectory is additionally drawn on the selected frames. \emph{Multi-Frame} uniformly samples intermediate states for tasks that evaluate process integrity, validity, or temporal continuity.
Together, these regimes preserve the essential solution information of the video while providing compact image-based representations. We emphasize that videos and images are not separate corpora, but different renderings of the same underlying problem with equivalent solutions. This design enables controlled comparison between video and image generation under identical task conditions.
Further design details and examples are provided in \cref{app:multimodal_examples}.

\paragraph{Curation.}
Since configurations are sampled procedurally, each generator can produce a large number of diverse instances. To ensure reliability, we review all 300 generators by inspecting rendered samples and verifying their solver implementations. We also check the resolution, frame rate, and frame count of every task before release. For the interleaved setting, we additionally use Gemini-3.1-Pro-Preview~\cite{gemini31pro} to generate textual descriptions of intermediate solution steps. We design a dedicated prompt for each task and apply it uniformly to all instances from that task. Further details are provided in \cref{app:interleaved_generation}.

\subsection{Statistics and Analysis}
\label{sec:dataset:analysis}

\cref{tab:datasets_comparison} compares \name with existing visual reasoning benchmarks and datasets in terms of task coverage, dataset scale, supported modalities, and evaluation protocols. Amongst 300 tasks, we hold out 50 tasks for out-of-domain evaluation, and sample 5,000 instances per task
from the remaining 250 tasks, resulting in 1.25M training instances. \benchname contains the 50 out-of-domain tasks and another 50 tasks from the 250 tasks with training data as in-domain tasks (total 100 tasks). In addition, \name includes a dedicated training set for reinforcement learning that covers the 50 in-domain tasks, which is used in the experiments in \cref{sec:rl}. Among the compared resources, \name provides the largest training set and uniquely supports video, image, and interleaved modalities under a unified task distribution. 

\cref{fig:data_overview} shows representative tasks from \name, colored according to their corresponding cognitive faculties. Each faculty spans a wide range of visual environments, from primitives on blank canvases to structured game states, board configurations, and sprite-based scenes. 
Compared with the 150 reworked tasks adapted from VBVR, the 150 newly developed tasks are designed to provide broader and more challenging coverage of visual reasoning scenarios. We analyze this expansion along two dimensions.
First, in terms of visual complexity, averaged over 50 sampled question frames per task, the median new task contains 80 distinct connected color regions, compared with 12 for the reworked tasks. Other metrics, such as spatial information and encoded image size, exhibit the same trend  (\cref{app:visual_complexity}). 
Second, in terms of reasoning depth, the reworked tasks are often solvable by applying a single rule to the input, whereas many new tasks, such as Sudoku and Gomoku, require multi-step search over candidate solutions or actions. In a blinded pairwise comparison, the new task was judged to require deeper reasoning in 113 of the 150 pairs. Moreover, only 7\% of the reworked VBVR tasks require multi-step reasoning, compared with 47\% of the new tasks (\cref{app:reasoning_depth}).
As shown in~\cref{sec:baselines:transferability}, training on \dataname consequently yields stronger and more consistent performance on downstream visual reasoning benchmarks.

\section{Verifiable Reward Scorers}
\label{sec:4_verifiable_reward_scorer}

A trustworthy benchmark requires evaluators that are accurate, inexpensive, and reproducible. The mainstream \textit{VLM-as-a-judge} paradigm often fails to satisfy these requirements, especially for visual reasoning tasks that demand precise verification (\cref{subsec:vlm_limitations}). To address this limitation, we build a suite of \textit{verifiable reward scorers} for all 100 tasks (50 In-Domain and 50 Out-of-Domain) in \benchname (\cref{subsec:implementation}). We further validate these scorers against large-scale human preferences (\cref{subsec:alignment}), showing that they achieve stronger human alignment at substantially lower cost than VLM-based judges. These properties make the scorers a critical foundation not only for reliable evaluation, but also for large-scale multi-task reinforcement learning for visual reasoning, as studied in~\cref{sec:rl}.

\subsection{Limitations of VLM-as-a-Judge}
\label{subsec:vlm_limitations}

\begin{figure}[t]
  \centering
  \includegraphics[width=\columnwidth]{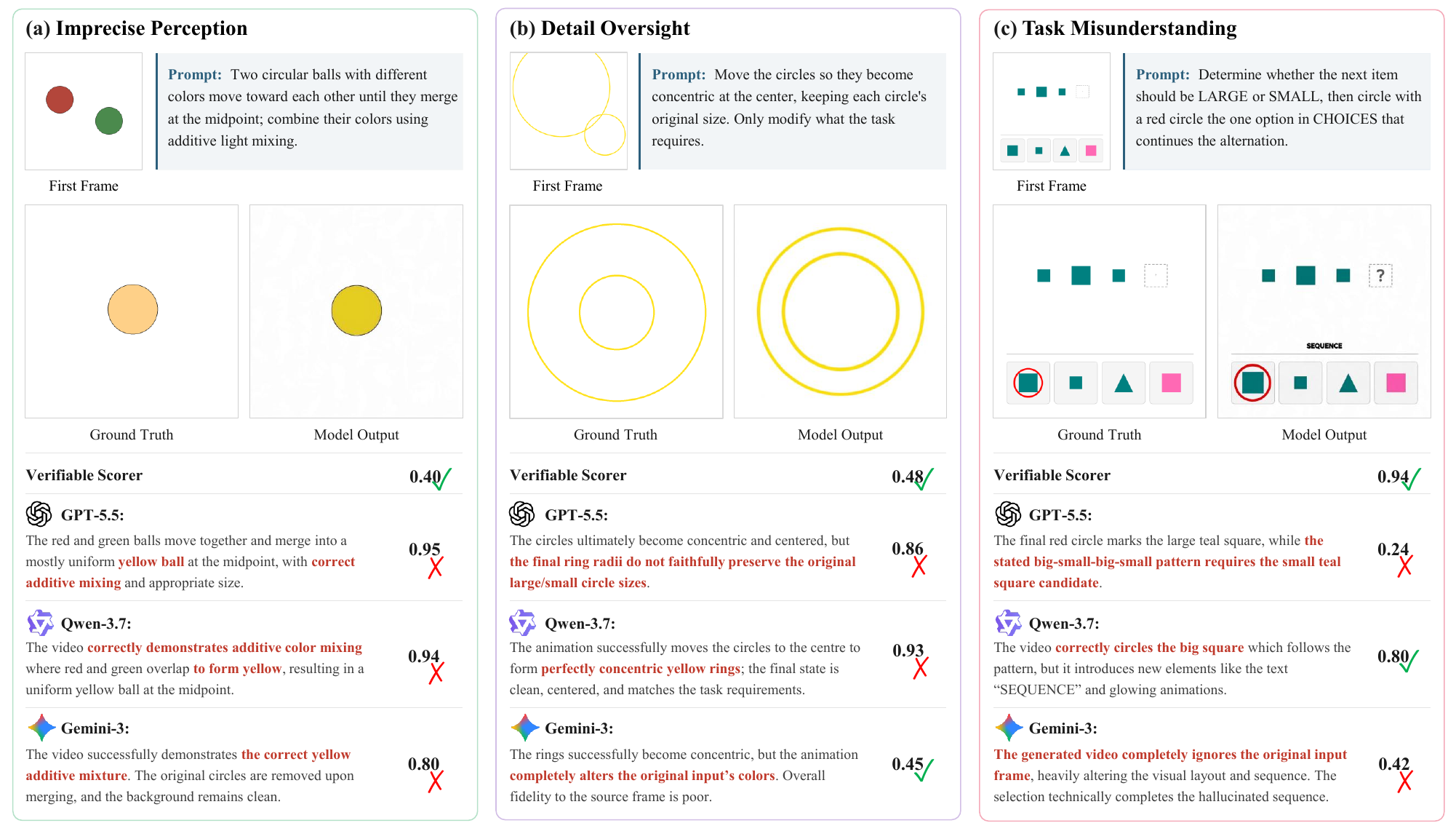}
  \caption{Three failure modes of VLM-as-a-judge (\cref{subsec:vlm_limitations}). For each mode we show an example consisting of the ground-truth and a model's output, together with the scores and rationales of three strong VLM judges.}
  \label{fig:vlm_failure}
\end{figure}

VLM-as-a-judge has become a common evaluation choice for many recent visual reasoning benchmarks (\cref{tab:datasets_comparison}) because it is easy to deploy and applicable to open-ended outputs. However, we find that this paradigm is problematic when used either as an evaluation metric or as a reward function for reinforcement learning \cite{liu2025your, motamed2025travl,  motamed2026generative}.

First, VLM judges can be inaccurate. Illustrated in~\cref{fig:vlm_failure}, we observe three representative failure modes. \textbf{(a)} Judges can be imprecise on fine-grained visual perception. For example, in an additive color-mixing task, the generated ball has an incorrect hue, yet all three VLM judges assign scores of at least $0.80$, compared with $0.40$ from our scorer. \textbf{(b)} Judges can overlook decisive errors. In a concentric-ring task, the output changes circle sizes and colors that should be preserved, but the judges still assign high scores, such as $0.78$ or $0.93$. \textbf{(c)} Judges can misunderstand the task itself. In a pattern-completion task, a correct answer receives $0.94$ from our scorer but only $0.24$--$0.80$ from VLM judges. These examples show that VLM judges can make mistakes in both directions: overestimating incorrect outputs and underestimating correct ones.

Second, VLM judges are expensive. This limitation becomes particularly severe in reinforcement learning, where a large number of rollouts must be scored over many optimization steps. Proprietary judges incur substantial API costs, while open-source judges require GPU inference that competes with model training under a fixed compute budget. In contrast, our task-specific scorers are lightweight and substantially cheaper per evaluation (\cref{fig:scorer_vlm_cost}).

Third, VLM judges are not reliably reproducible. Even at temperature~$0$, repeated evaluations on the same videos can produce different scores for $55$--$93\%$ of samples, whereas our deterministic scorer produces identical results across runs (\cref{tab:reproducible}). This reproducibility is especially important when scores are used as per-sample rewards: inconsistent feedback can introduce noisy or conflicting optimization signals and hinder effective learning.

\begin{table}[h!]
  \centering
  \small
  \setlength{\tabcolsep}{4pt}
  \caption{\textbf{Reproducibility of evaluators.} We re-run each evaluator on the same videos. Our verifiable scorer produces identical scores across runs, while VLM judges change scores for a large fraction of cases.}
  \label{tab:reproducible}
  \resizebox{\linewidth}{!}{%
  \begin{tabular}{lccccccc}
    \toprule
    & \textbf{Scorer} & GLM-4.6V-Flash~\cite{glm_hong2025glm} & Gemma4-31B~\cite{gemma_team2026gemma} & Qwen3.6-27B~\cite{qwen3.6-27b} & Qwen3.7-plus~\cite{qwen37} & InternVL3.5-38B~\cite{internvl_wang2025internvl3} & Gemini-3.1-Pro~\cite{gemini31pro} \\
    \midrule
    Cases with Changed Score (\%) $\downarrow$ & \textbf{0.0}   & 54.6  & 69.5  & 74.5  & 80.8  & 82.9  & 92.8 \\
    Max. Score Change (0--1) $\downarrow$        & \textbf{0.000} & 0.125 & 0.110 & 0.117 & 0.091 & 0.158 & 0.221 \\
    \bottomrule
  \end{tabular}%
  }
\end{table}

\subsection{Design of Verifiable Reward Scorers}
\label{subsec:implementation}

\begin{figure}[t]
  \centering
  \includegraphics[width=\columnwidth]{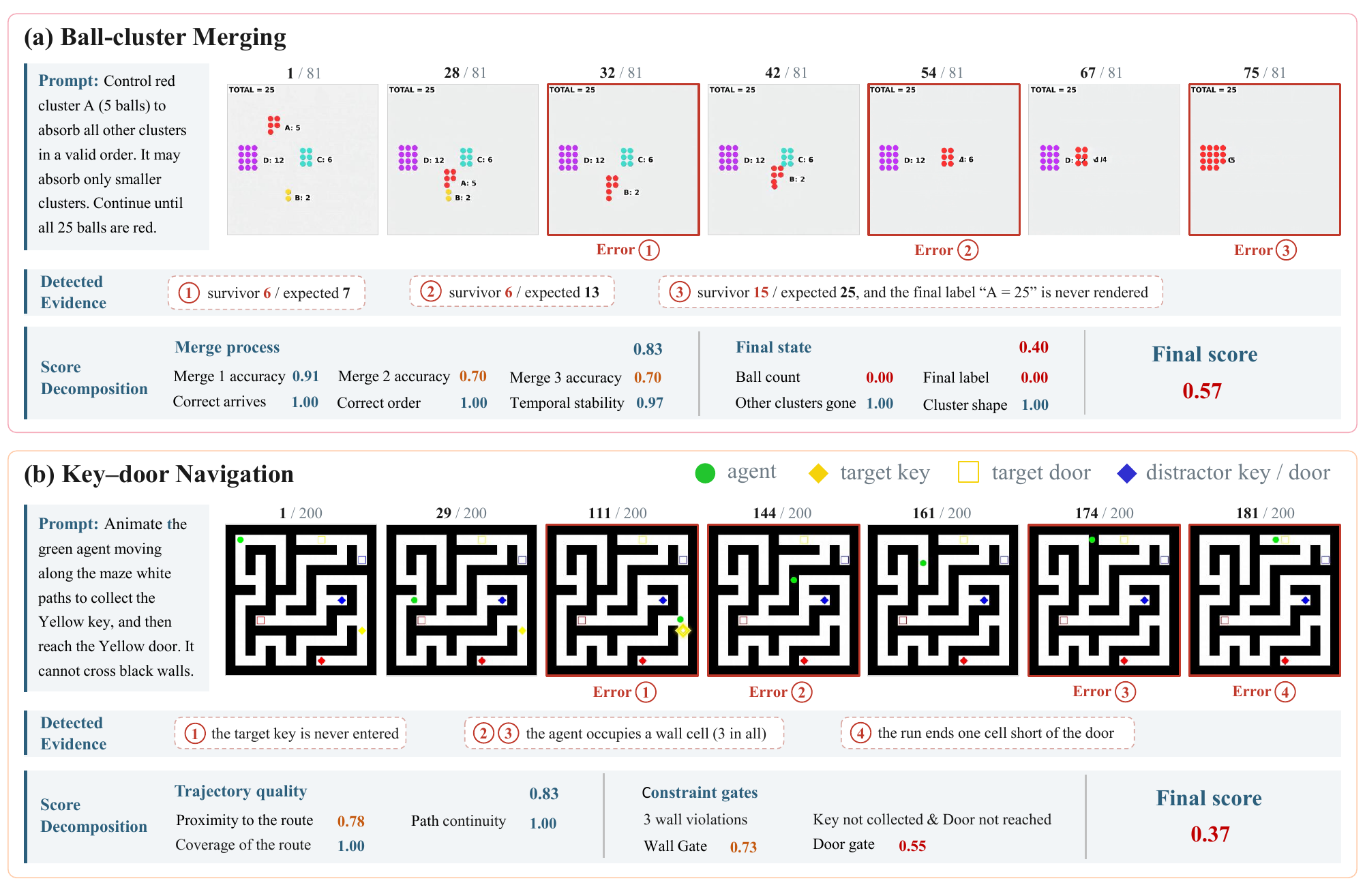}
  \caption{\textbf{Examples of task-specific scoring.} We show how verifiable reward scorers evaluate generated videos by decomposing each output into task-relevant checks and aggregating them into a final score.}
  \label{fig:scorer_example}
\end{figure}

We implement a dedicated verifiable reward scorer for each of the 100 tasks in \benchname. Whenever possible, the scorers avoid relying on pixel-level differences and instead operate on task-relevant semantic entities. Each scorer locates the objects of interest and extracts their attributes, such as color, shape, position, and count, using classical computer-vision methods including HSV color segmentation, contour detection, OCR for rendered digits and labels, and trajectory tracking across frames. Grading is then performed over these extracted attributes rather than raw pixels. This semantic design makes the scores more interpretable and robust: a pixel-level comparison may penalize a correct answer due to slight spatial shifts or rendering-style differences, whereas attribute-level verification focuses on task-relevant correctness.

\cref{fig:scorer_example} illustrates two evaluators end to end. Given the extracted attributes, each evaluator applies a set of task-specific checks with human-designed and carefully calibrated weights, and aggregates them into a final score. For tasks with soft criteria, the final score is computed as a weighted sum, as shown in ~\cref{fig:scorer_example}(a). For tasks with hard constraints, such as avoiding wall crossings, checks are combined multiplicatively so that a single violation sharply reduces the final score, as shown in ~\cref{fig:scorer_example}(b). Because the scorers are implemented as lightweight computations, they require no API calls, use GPU computation only optionally for OCR, and are fully deterministic. They also provide interpretable error decomposition, indicating which task-specific checks a sample fails.

\subsection{Human Preference Alignment}
\label{subsec:alignment}

To evaluate both accuracy and efficiency, we validate our scorers against a large collection of human preference annotations and compare them with multiple VLM-as-a-judge baselines under the same evaluation setting.
\benchname contains 100 tasks with five instances per task. We evaluate eight video generation models, yielding $4{,}000$ videos in total. To assess evaluator reliability, we sample one instance per task, compare all pairs of models, and collect ten independent human annotations for each pair. Annotators rate the absolute quality of both videos using a standardized rubric, and we convert these ratings into pairwise preferences, equivalent to a standard four-way arena format. We map the scores produced by each automatic evaluator to the same preference format and measure agreement with human judgments. We also estimate evaluation cost based on API token usage for proprietary judges and GPU runtime for open-source judges using cloud-service pricing. Full annotation criteria, preference mappings, and cost calculations are provided in~\cref{app:human_preference_alignmrnt}.

\begin{figure}[t]
\centering
\includegraphics[width=\columnwidth]{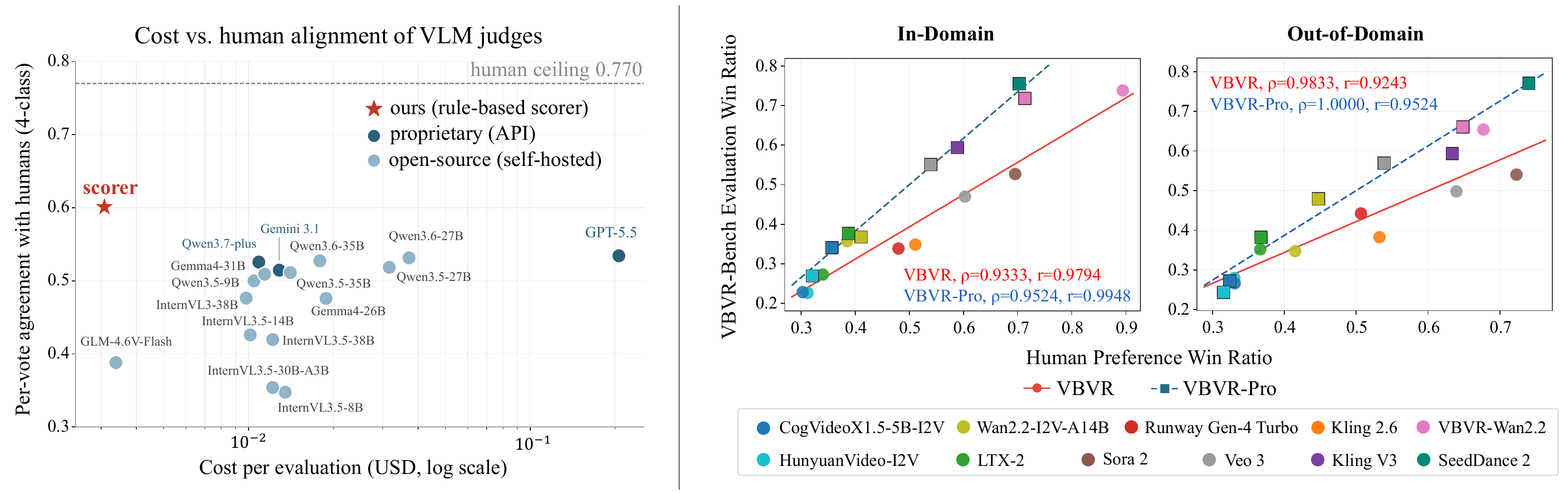}
\caption{\textbf{Left:} Per-vote agreement with human preferences ($y$) versus cost per evaluation ($x$, log scale) for each evaluator. \textbf{Right:} Model-level agreement. Each point denotes one generation model, with its scorer-based win ratio on the $y$-axis and human win ratio on the $x$-axis, for in-domain and out-of-domain tasks.}
\label{fig:scorer_vlm_cost}
\end{figure}

As shown in~\cref{fig:scorer_vlm_cost} (left), our scorer achieves the highest agreement with human preferences while remaining the least expensive evaluator. Its per-vote agreement exceeds $0.60$, outperforming all VLM judges, including GPT-5.5~\cite{gpt55} ($0.54$) and Gemini-3.1-Pro~\cite{gemini31pro} ($0.52$), while reaching approximately $78\%$ of the human agreement ceiling of $0.77$. We note that perfect agreement is not expected, since human annotators themselves can disagree on close comparisons. As a lightweight rule-based evaluator, our scorer requires only minimal or no GPU computation, resulting in the lowest cost per evaluation. Its deterministic procedure further ensures reproducible scoring.

Beyond agreement on individual comparisons, a benchmark should also reliably rank generation models. We therefore compute each model's win ratio under human evaluation and under our scorer, and measure their correlation across models (\cref{fig:scorer_vlm_cost}, right). Compared with the earlier VBVR scorer, the VBVR-Pro scorer improves correlation on every split, most notably out-of-domain, where Spearman's $\rho$ increases from $0.93$ to $1.00$ and Pearson's $r$ from $0.92$ to $0.95$. Overall, the final scorer reaches $\rho=0.95$ in-domain and $\rho=1.00$ out-of-domain, with $r\geq0.95$ on both splits, demonstrating that it can reliably approximate human evaluation when comparing visual generation models.

\section{Benchmarking Native Visual Reasoning}
\label{sec:baselines}

In this section, we benchmark leading proprietary and open-source generative models on \benchname. We first report overall performance and analyze the behavior of different generative paradigms, including image, video, and interleaved generation (\cref{sec:baselines:leaderboard}). We then conduct counterfactual diagnostics to examine whether trained models rely on reusable reasoning trajectories rather than superficial shortcuts (\cref{sec:baselines:counterfactual}). Finally, we visualize native visual reasoning behaviors through Chain-of-Step analysis (\cref{sec:baselines:cos}) and evaluate transferability to a wide range of unseen benchmarks (\cref{sec:baselines:transferability}).

\subsection{Leading Model Performance}
\label{sec:baselines:leaderboard}

\subsubsection{Training Open-source Models}



\begin{wraptable}{l}{0.4\textwidth}
\vspace{-8pt}
\centering
\scriptsize
\setlength{\tabcolsep}{3pt}
\renewcommand{\arraystretch}{1.10}
\captionsetup{skip=2pt} 
\caption{Configurations of the open-source generative foundation models. ``A" indicates activated parameters in MoE and MoT. }
\label{tab:base_models}
\begin{tabular}{lcc}
\toprule
\textbf{Model} & \textbf{Parameters} & \textbf{Architecture} \\
\midrule

\rowcolor{gray!20}
\multicolumn{3}{l}{\textbf{Image Generation Models}} \\
BAGEL-7B-MoT \cite{deng2025bagel}
    & 14BA7B & MoT \\
FLUX.2-dev \cite{flux-2-2025}
    & 32B & Dense \\
Qwen-Image-Edit \cite{wu2025qwenimagetechnicalreport}
    & 20B & Dense \\

\midrule
\rowcolor{gray!20}
\multicolumn{3}{l}{\textbf{Interleaved Text--Image Generation Models}} \\
ThinkMorph-7B \cite{gu2025thinkmorph}
    & 14BA7B & MoT \\
SenseNova-U1-8B-MoT \cite{sensenova_u1}
    & 16BA8B & MoT \\

\midrule
\rowcolor{gray!20}
\multicolumn{3}{l}{\textbf{Video Generation Models}} \\
Wan2.1-I2V-14B-720P \cite{wan2025wan}
    & 14B & Dense \\
Wan2.2-TI2V-5B \cite{wan2025wan}
    & 5B & Dense \\
Wan2.2-I2V-A14B \cite{wan2025wan}
    & 27BA14B & MoE \\
LTX-2.3-I2AV \cite{ltx23}
    & 22B & Dense \\
\bottomrule
\end{tabular}
\end{wraptable}
We select nine open-source generative foundation models spanning image generation, video generation, and interleaved text-image generation to investigate how reasoning behaviors differ across generative modalities after training. The selected models exhibit substantial diversity in scale, architecture, cross-modal interaction mechanism, and output representation, covering both sparse and dense architectures as well as latent- and pixel-space generation paradigms. Further architectural details are provided in \cref{sec:architecture_details}.
All models are trained on \dataname for one epoch at a resolution of $512\times512$. Rank-32 LoRA is applied to all video generators and to image and interleaved text-image models with more than 10B parameters, whereas smaller image and interleaved text-image models are fully fine-tuned. Video models are trained at $16$ fps. Additional training and implementation details are provided in \cref{sec:training_details}. Fine-tuned models are denoted with the prefix \textit{VBVR-Pro-}.

\begin{table*}[!t]
\centering
\scriptsize
\setlength{\tabcolsep}{3pt}
\caption{Benchmarking results on \textbf{\benchname}. Overall, In-Domain (ID), and Out-of-Domain (OOD) scores are reported alongside category-wise performance. \textbf{Bold}: best in subgroup; \underline{underline}: second best.}
\resizebox{1.0\linewidth}{!}{
\begin{tabular}{l|c|c|ccccc|c|ccccc}
\toprule
& \multicolumn{1}{c|}{}
& \multicolumn{6}{c|}{\textbf{In-Domain by Category}}
& \multicolumn{6}{c}{\textbf{Out-of-Domain by Category}} \\
\cmidrule(lr){3-3}\cmidrule(lr){4-8}\cmidrule(lr){9-9}\cmidrule(lr){10-14}
\textbf{Models} & \textbf{Overall} & \textbf{Avg.} & \textbf{Abst.} & \textbf{Know.} & \textbf{Perc.} & \textbf{Spat.} & \textbf{Trans.} & \textbf{Avg.} & \textbf{Abst.} & \textbf{Know.} & \textbf{Perc.} & \textbf{Spat.} & \textbf{Trans.} \\
\midrule
\rowcolor{gray!20}
\multicolumn{14}{l}{\textbf{Image Generation Models}} \\
\midrule
\rowcolor{line-blue}
\multicolumn{14}{l}{\textbf{Proprietary Models}} \\
Qwen-Image-2.0 \cite{zhao2026qwen}
& \underline{0.313} & \underline{0.248} & \underline{0.350} & \underline{0.220} & \underline{0.300} & \underline{0.187} & \underline{0.186} & \underline{0.378} & \underline{0.463} & \underline{0.292} & \underline{0.440} & \underline{0.384} & \underline{0.088} \\
Seedream-5.0-Pro \cite{bytedance2026seedream50pro}
& \textbf{0.557} & \textbf{0.485} & \textbf{0.673} & \textbf{0.349} & \textbf{0.679} & \textbf{0.441} & \textbf{0.307} & \textbf{0.629} & \textbf{0.689} & \textbf{0.565} & \textbf{0.744} & \textbf{0.559} & \textbf{0.223} \\
\midrule
\rowcolor{line-blue}
\multicolumn{14}{l}{\textbf{Open-source Models}} \\
BAGEL-7B-MoT \cite{deng2025bagel}
& 0.089 & \underline{0.066} & 0.050 & \underline{0.095} & 0.089 & 0.050 & 0.038 & 0.111 & \textbf{0.272} & 0.039 & 0.082 & 0.028 & \textbf{0.133} \\
FLUX.2-dev \cite{flux-2-2025}
& \textbf{0.157} & \textbf{0.108} & \underline{0.114} & \textbf{0.122} & \underline{0.096} & \underline{0.110} & \textbf{0.093} & \textbf{0.206} & \underline{0.268} & \textbf{0.205} & \textbf{0.207} & \textbf{0.241} & 0.085 \\
Qwen-Image-Edit \cite{wu2025qwenimagetechnicalreport}
& \underline{0.134} & \textbf{0.108} & \textbf{0.120} & 0.092 & \textbf{0.134} & \textbf{0.120} & \underline{0.079} & \underline{0.159} & 0.239 & \underline{0.078} & \underline{0.158} & \underline{0.182} & \underline{0.091} \\
\midrule
\rowcolor{line-blue}
\multicolumn{14}{l}{\textbf{Strong Baselines}} \\
VBVR-Pro-BAGEL
& 0.172 & 0.168 & 0.259 & 0.117 & 0.146 & 0.234 & 0.078 & 0.176 & 0.345 & 0.130 & 0.167 & 0.015 & \underline{0.160} \\
VBVR-Pro-FLUX.2
& \textbf{0.407} & \textbf{0.484} & \textbf{0.628} & \textbf{0.361} & \textbf{0.490} & \textbf{0.494} & \textbf{0.475} & \textbf{0.330} & \textbf{0.490} & \textbf{0.338} & \textbf{0.287} & \textbf{0.454} & 0.141 \\
VBVR-Pro-Qwen-Image-Edit
& \underline{0.322} & \underline{0.332} & \underline{0.387} & \underline{0.243} & \underline{0.257} & \underline{0.474} & \underline{0.313} & \underline{0.311} & \underline{0.464} & \underline{0.297} & \underline{0.262} & \underline{0.413} & \textbf{0.200} \\
\midrule
\rowcolor{gray!20}
\multicolumn{14}{l}{\textbf{Interleaved Text-image Generation Models}} \\
\midrule
\rowcolor{line-blue}
\multicolumn{14}{l}{\textbf{Proprietary Models}} \\
GPT-Image-2 \cite{openai2026chatgptimages}
& \underline{0.507} & \underline{0.428} & \underline{0.593} & \underline{0.356} & \underline{0.571} & \underline{0.227} & \textbf{0.424} & \underline{0.587} & \underline{0.540} & \underline{0.513} & \underline{0.712} & \underline{0.480} & \textbf{0.335} \\
Nano Banana Pro \cite{raisinghani2025nanobananapro}
& \textbf{0.564} & \textbf{0.480} & \textbf{0.674} & \textbf{0.473} & \textbf{0.682} & \textbf{0.313} & \underline{0.246} & \textbf{0.648} & \textbf{0.751} & \textbf{0.619} & \textbf{0.739} & \textbf{0.585} & \underline{0.243} \\
\midrule
\rowcolor{line-blue}
\multicolumn{14}{l}{\textbf{Open-source Models}} \\
ThinkMorph-7B \cite{gu2025thinkmorph}
& \underline{0.154} & \underline{0.113} & \underline{0.130} & \underline{0.092} & \underline{0.134} & \underline{0.162} & \underline{0.044} & \underline{0.195} & \underline{0.238} & \underline{0.206} & \underline{0.183} & \underline{0.253} & \underline{0.114} \\
VBVR-SenseNova-U1 \cite{vbvr2026}
& \textbf{0.408} & \textbf{0.469} & \textbf{0.463} & \textbf{0.351} & \textbf{0.497} & \textbf{0.425} & \textbf{0.674} & \textbf{0.347} & \textbf{0.395} & \textbf{0.393} & \textbf{0.309} & \textbf{0.480} & \textbf{0.263} \\
\midrule
\rowcolor{line-blue}
\multicolumn{14}{l}{\textbf{Strong Baselines}} \\
VBVR-Pro-ThinkMorph
& \underline{0.373} & \underline{0.402} & \underline{0.524} & \underline{0.385} & \underline{0.317} & \underline{0.500} & \underline{0.259} & \underline{0.344} & \underline{0.499} & \underline{0.278} & \underline{0.268} & \underline{0.535} & \underline{0.284} \\
VBVR-Pro-SenseNova-U1
& \textbf{0.638} & \textbf{0.811} & \textbf{0.843} & \textbf{0.778} & \textbf{0.828} & \textbf{0.847} & \textbf{0.764} & \textbf{0.464} & \textbf{0.651} & \textbf{0.407} & \textbf{0.387} & \textbf{0.558} & \textbf{0.451} \\
\midrule
\rowcolor{gray!20}
\multicolumn{14}{l}{\textbf{Video Generation Models}} \\
\midrule
\rowcolor{line-blue}
\multicolumn{14}{l}{\textbf{Proprietary Models}} \\
Veo 3.1 \cite{deepmind2025veo3}
& 0.309 & 0.312 & \underline{0.358} & 0.335 & 0.335 & 0.294 & 0.222 & 0.305 & \underline{0.414} & 0.290 & 0.283 & \underline{0.312} & 0.242 \\
Kling VIDEO 3.0 \cite{kuaishou2025kling26}
& \underline{0.392} & \underline{0.356} & 0.277 & \underline{0.365} & \underline{0.426} & \underline{0.390} & \underline{0.323} & \underline{0.427} & 0.399 & \textbf{0.700} & \underline{0.421} & 0.242 & \underline{0.455} \\
Seedance 2.0 \cite{seedance2026seedance}
& \textbf{0.499} & \textbf{0.451} & \textbf{0.439} & \textbf{0.405} & \textbf{0.471} & \textbf{0.515} & \textbf{0.435} & \textbf{0.547} & \textbf{0.501} & \underline{0.634} & \textbf{0.538} & \textbf{0.538} & \textbf{0.588} \\
\midrule
\rowcolor{line-blue}
\multicolumn{14}{l}{\textbf{Open-source Models}} \\
HunyuanVideo-I2V \cite{kong2024hunyuan}
& 0.054 & 0.054 & 0.030 & 0.072 & 0.020 & 0.092 & 0.046 & 0.053 & 0.120 & 0.018 & 0.032 & 0.062 & 0.061 \\
CogVideoX1.5-5B-I2V \cite{yang2024cogvid}
& 0.085 & 0.100 & 0.079 & 0.132 & 0.092 & 0.101 & 0.085 & 0.070 & 0.169 & 0.047 & 0.058 & 0.040 & 0.026 \\
Wan2.1-I2V-14B-720P \cite{wan2025wan}
& 0.100 & 0.105 & 0.068 & 0.141 & 0.121 & 0.113 & 0.073 & 0.095 & 0.152 & 0.090 & 0.080 & 0.123 & 0.049 \\
Wan2.2-TI2V-5B \cite{wan2025wan}
& 0.094 & 0.066 & 0.037 & 0.082 & 0.067 & 0.091 & 0.044 & 0.122 & 0.212 & 0.064 & 0.119 & 0.063 & 0.110 \\
Wan2.2-I2V-A14B \cite{wan2025wan}
& \underline{0.182} & \underline{0.157} & \underline{0.106} & \underline{0.147} & \underline{0.147} & \underline{0.178} & \underline{0.221} & \underline{0.207} & \underline{0.304} & \underline{0.172} & \underline{0.157} & \underline{0.195} & \underline{0.301} \\
LTX-2.3-I2AV \cite{ltx23}
& 0.112 & 0.106 & 0.080 & 0.122 & 0.093 & 0.146 & 0.077 & 0.119 & 0.219 & 0.168 & 0.096 & 0.091 & 0.056 \\
VBVR-Wan2.2 \cite{vbvr2026}
& \textbf{0.517} & \textbf{0.548} & \textbf{0.308} & \textbf{0.558} & \textbf{0.446} & \textbf{0.623} & \textbf{0.834} & \textbf{0.486} & \textbf{0.421} & \textbf{0.426} & \textbf{0.389} & \textbf{0.732} & \textbf{0.756} \\
\midrule
\rowcolor{line-blue}
\multicolumn{14}{l}{\textbf{Strong Baselines}} \\
VBVR-Pro-LTX2.3
& 0.425 & 0.527 & 0.531 & 0.571 & 0.461 & 0.506 & 0.551 & 0.324 & 0.518 & 0.134 & 0.201 & 0.477 & 0.427 \\
VBVR-Pro-Wan2.1-I2V-14B
& \underline{0.562} & \underline{0.730} & \underline{0.802} & \underline{0.649} & \underline{0.603} & \underline{0.744} & \underline{0.880} & \underline{0.395} & \underline{0.557} & \underline{0.379} & \underline{0.259} & \underline{0.617} & \underline{0.486} \\
VBVR-Pro-Wan2.2-TI2V-5B
& 0.470 & 0.641 & 0.687 & 0.623 & 0.498 & 0.621 & 0.786 & 0.300 & 0.453 & 0.158 & 0.181 & 0.505 & 0.452 \\
VBVR-Pro-Wan2.2-I2V-A14B
& \textbf{0.670} & \textbf{0.808} & \textbf{0.821} & \textbf{0.768} & \textbf{0.742} & \textbf{0.826} & \textbf{0.898} & \textbf{0.532} & \textbf{0.651} & \textbf{0.519} & \textbf{0.393} & \textbf{0.679} & \textbf{0.763} \\
\bottomrule
\end{tabular}
}
\label{tab:vbvr_results}
\end{table*}

\subsubsection{Main Results}

\cref{tab:vbvr_results} compares more than 30 proprietary and open-source models on \benchname. We find that top-tier proprietary models, such as Seedance 2.0 and GPT-Image-2, achieve strong performance, indicating that current generative models already exhibit nontrivial visual reasoning capabilities. In contrast, most academic open-source models lag substantially behind, for which we believe the availability of task-relevant training data is an important bottleneck. Consistent with this observation, task-specific training on \dataname improves all nine models across image, interleaved text-image, and video generation, with an average overall gain of $0.290$. Nevertheless, even the strongest trained model remains well below human-level performance, highlighting the considerable room for improvement.

To distinguish within-task generalization from cross-task transfer, we evaluate models under both in-domain (ID) and out-of-domain (OOD) settings. ID evaluation uses unseen instances from the same task distributions as training and measures generalization within known task families, whereas OOD evaluation introduces different task distributions and measures whether learned visual reasoning capabilities transfer to unseen tasks. Training yields pronounced gains on ID tasks, with an average improvement of $+0.401$, while OOD performance also improves consistently, with an average gain of $+0.179$. This provides evidence that \dataname improves not only instance-level generalization within known tasks, but also transfer across task families. We further examine transferability to external reasoning benchmarks, including real-world and embodied scenarios, in \cref{sec:baselines:transferability}.

\subsubsection{Image, Interleaved Text-Image, and Video Generation}
\begin{figure}[t]
  \centering
  \includegraphics[width=\linewidth]{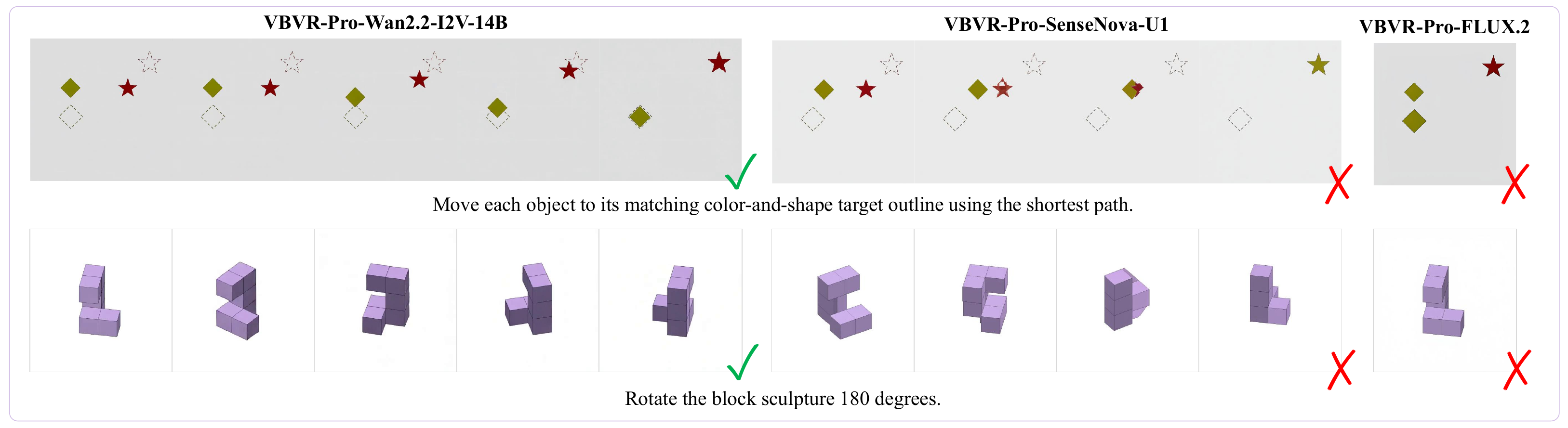}
  \caption{Qualitative comparison across visual reasoning modalities on transformation tasks.
  }
  \label{fig:video_vs_interleaved}
\end{figure}

A common intuition is that video generation should outperform interleaved text-image generation because videos provide dense temporal continuity. After task-specific training, however, we find that the strongest interleaved text-image model, VBVR-Pro-SenseNova-U1, performs on par with the strongest video model, VBVR-Pro-Wan2.2-I2V-A14B, across many ID tasks. This suggests that, once the underlying task structure is learned, a sparse sequence of discrete but task-relevant visual states can be sufficient for solving many familiar visual reasoning problems. Moreover, interleaved models incur substantially lower generation and inference costs, yielding a favorable accuracy--efficiency trade-off.

Video generation nevertheless retains a clear advantage when successful reasoning requires accurate modeling of state transitions. Video models perform better on ID transformation tasks and across OOD categories, suggesting that dense temporal representations become particularly valuable when models must track fine-grained changes over time or generalize learned transition dynamics beyond the training distribution. This distinction is qualitatively illustrated in~\cref{fig:video_vs_interleaved}: the video model generates coherent intermediate states and reaches the correct solution on both the shortest-path object-manipulation task and the $180^\circ$ block-rotation task, whereas the interleaved text-image and single-image models fail.

Single-image generation remains substantially weaker than both trajectory-based representations, particularly on transformation tasks. This might because a single endpoint image does not explicitly capture the intermediate states involved in the solution process, it provides limited support for modeling multi-step state transitions.

\subsubsection{Visual Trajectory Supervision Contributes More Than Textual Semantics}
\label{sec:visual_vs_text}

\begin{table*}[h]
\centering
\scriptsize
\setlength{\tabcolsep}{3pt}
\caption{
Ablation on \textbf{interleaved modalities}. \textbf{Text} indicates whether the output contains meaningful reasoning (\emph{Full}) or semantically uninformative placeholders (\emph{Placeholder}), while \textbf{Images} indicates whether the intermediate visual trajectory contains multiple states (\emph{Multi}) or a single visual representation (\emph{Single}).
}
\resizebox{\linewidth}{!}{
\begin{tabular}{c|c|cc!{\color{pr@accent}\vrule width 1.2pt}c|cccccc|cccccc}
\toprule
& & \multicolumn{2}{c!{\color{pr@accent}\vrule width 1.2pt}}{\textbf{Modalities}}
& \multicolumn{1}{c|}{}
& \multicolumn{6}{c|}{\textbf{In-Domain by Category}}
& \multicolumn{6}{c}{\textbf{Out-of-Domain by Category}} \\
\cmidrule(lr){3-4}
\cmidrule(lr){6-11}
\cmidrule(lr){12-17}

\textbf{\#}
& \textbf{Models}
& \textbf{Text}
& \textbf{Images}
& \textbf{Overall}
& \textbf{Avg.}
& \textbf{Abst.}
& \textbf{Know.}
& \textbf{Perc.}
& \textbf{Spat.}
& \textbf{Trans.}
& \textbf{Avg.}
& \textbf{Abst.}
& \textbf{Know.}
& \textbf{Perc.}
& \textbf{Spat.}
& \textbf{Trans.} \\
\midrule

1-1
& \multirow{3}{*}{ThinkMorph}
& Full
& Multi
& \underline{0.373}
& \textbf{0.402}
& \textbf{0.403}
& \underline{0.344}
& 0.238
& \textbf{0.454}
& \underline{0.184}
& \underline{0.344}
& \underline{0.367}
& \textbf{0.224}
& \underline{0.238}
& \underline{0.535}
& \underline{0.257} \\

1-2
& 
& Full
& Single
& 0.349
& 0.382
& \underline{0.402}
& 0.290
& \underline{0.244}
& 0.391
& \textbf{0.219}
& 0.316
& 0.333
& 0.143
& 0.225
& \textbf{0.538}
& 0.235 \\

1-3
&
& Placeholder
& Multi
& \textbf{0.374}
& \underline{0.399}
& 0.384
& \textbf{0.350}
& \textbf{0.250}
& \underline{0.440}
& \underline{0.184}
& \textbf{0.348}
& \textbf{0.372}
& \underline{0.187}
& \textbf{0.262}
& 0.479
& \textbf{0.266} \\

\midrule

2-1
& \multirow{3}{*}{SenseNova-U1}
& Full
& Multi
& \textbf{0.638}
& \underline{0.811}
& \underline{0.648}
& \textbf{0.695}
& \underline{0.621}
& \textbf{0.770}
& \underline{0.541}
& \textbf{0.464}
& \textbf{0.480}
& \underline{0.328}
& \textbf{0.344}
& \underline{0.558}
& \textbf{0.408} \\

2-2
&
& Full
& Single
& 0.527
& 0.643
& 0.637
& 0.464
& 0.608
& 0.560
& 0.323
& 0.412
& 0.457
& 0.276
& 0.319
& \textbf{0.578}
& 0.199 \\

2-3
&
& Placeholder
& Multi
& \underline{0.629}
& \textbf{0.821}
& \textbf{0.661}
& \underline{0.683}
& \textbf{0.660}
& \underline{0.734}
& \textbf{0.577}
& \underline{0.436}
& \underline{0.469}
& \textbf{0.351}
& \underline{0.329}
& 0.468
& \underline{0.339} \\

\bottomrule
\end{tabular}
}
\label{tab:vbvr_ablation}
\end{table*}
To disentangle the relative contributions of textual and visual reasoning trajectory supervision to the interleaved models, we conduct an ablation study in~\cref{tab:vbvr_ablation} under three training settings. Exp.~1-1 and Exp.~2-1 serve as the reference settings and correspond to the original VBVR-Pro-ThinkMorph and VBVR-Pro-SenseNova-U1 models, respectively. Both models are trained on complete interleaved trajectories comprising meaningful reasoning text and multiple intermediate images.

Removing intermediate visual states leads to clear performance degradation. Exp.~1-2 and Exp.~2-2 retain the original reasoning text but compress each visual trajectory into a single image, thereby isolating the contribution of intermediate visual supervision. Compared with their respective reference models, these variants reduce the overall score by $0.024$ and $0.111$.

In contrast, replacing meaningful intermediate text with placeholders has a much smaller effect on overall performance. Exp.~1-3 and Exp.~2-3 preserve the multi-image trajectory while replacing reasoning text with semantically uninformative placeholders, isolating the contribution of textual semantics during intermediate reasoning.

These findings suggest that, for visual reasoning tasks in \name, explicit visual state transitions contribute more than intermediate textual reasoning. Many tasks in the benchmark, such as maze navigation, object translation and rotation, are inherently spatial and require tracking locations, order, directions, paths, and orientation changes. A sequence of intermediate visual states makes these transformations directly observable and therefore provides a natural representation of the underlying reasoning process.

\subsection{Counterfactual Diagnostics}
\label{sec:baselines:counterfactual}














\begin{wraptable}[12]{l}{0.5\textwidth}
\centering
\vspace{-6pt}
\footnotesize
\setlength{\tabcolsep}{4pt}
\caption{
Counterfactual diagnostics on a \textit{trained} \textbf{VBVR-Pro-SenseNova-U1} at the inference.
}
\label{tab:counterfactual_diagnostics}
\resizebox{\linewidth}{!}{
\begin{tabular}{c|l|ccc}
\toprule
\textbf{\#}
& \textbf{Intervention}
& \textbf{Overall}
& \textbf{ID Avg.}
& \textbf{OOD Avg.} \\
\midrule

\rowcolor{gray!20}
\multicolumn{5}{l}{\textbf{Experiment 1: Input Modalities}} \\
\midrule

1
& Original input text and image
& 0.638
& 0.820
& 0.455 \\

2
& Input text rephrased, semantically unchanged
& 0.640
& 0.811
& 0.468 \\

3
& Input text with task semantics removed
& 0.317
& 0.435
& 0.199 \\

4
& Input image removed
& 0.064
& 0.077
& 0.051 \\


5
& Horizontally flipped or rotated by $180^\circ$
& 0.613
& 0.786
& 0.439 \\

\midrule
\rowcolor{gray!20}
\multicolumn{5}{l}{\textbf{Experiment 2: Intermediate Reasoning States}} \\
\midrule

6
& Intermediate text removed
& 0.533
& 0.661
& 0.406 \\

7
& Intermediate text contradicted
& 0.530
& 0.670
& 0.390 \\

8
& Intermediate image removed
& 0.099
& 0.128
& 0.069 \\

9
& Intermediate image corrupted with noise
& 0.190
& 0.307
& 0.074 \\

\bottomrule
\end{tabular}
}
\end{wraptable}

The strong performance of trained models raises an important question: does task-specific training induce reusable visual reasoning, or does the model instead exploit superficial regularities in the training distribution, such as recurring instruction templates or memorized associations between prompts and outputs? We conduct three complementary counterfactual diagnostics to examine this question. The first perturbs the input text or image to test whether predictions depend on the underlying problem instance rather than its surface formulation. The second intervenes in intermediate reasoning states to determine which modality plays a more important role during the reasoning process. All experiments in this section are conducted on VBVR-Pro-SenseNova-U1.

\subsubsection{Intervention on Input Modalities}

As shown in \cref{tab:counterfactual_diagnostics}, \textbf{Intervention \#1} uses the original input text and image as the clean reference. \textbf{Intervention \#2} semantically rephrases the input text while preserving the task requirements: the performance is essentially unchanged, indicating that successful reasoning does not depend on the particular linguistic form of the instruction. 
In contrast, \textbf{Intervention \#3} removes the task-specific semantics from the input text, retaining only generic output requirements. This causes a substantial degradation, showing that the initial text still provides necessary task objectives rather than merely serving as a superficial trigger.
\textbf{Intervention \#4} removes the input image entirely while keeping the input text unchanged, which produces by far the largest degradation in this experiment, reducing performance to 0.064. The solution therefore cannot be recovered from the textual input alone.
\textbf{Intervention \#5} constructs geometric counterfactuals that preserve the underlying taskbut change its visual realization. Specifically, the input image is either horizontally flipped or rotated by $180^\circ$, and direction-dependent language in the input text is transformed consistently with the corresponding geometric transformation. Performance remains substantial under these transformations. When the spatial arrangement changes, the generated reasoning trajectory changes accordingly and can still reach the correct final state.

Taken together, these results argue against instruction-template fitting or question memorization: performance is insensitive to a semantically equivalent reformulation of the input text, but strongly dependent on both the task semantics and,  more importantly, the specific visual state on which the task must be executed. The gains from \dataname training are therefore difficult to explain as memorized mappings from recurring question patterns to outputs; they are more consistent with learning reusable visual relations and operations.

\subsubsection{Intervention on Intermediate Reasoning States}

While \cref{sec:visual_vs_text} studies which forms of supervision are most useful during training, here we examine which information is actually used during the interleaved reasoning trajectory of a trained model. Specifically, we intervene directly on intermediate outputs at inference time and measure their causal contribution to the subsequent rollout.

For the textual pathway, \textbf{Intervention \#6} removes the intermediate text before the subsequent generation step, while \textbf{Intervention \#7} replaces it with a contradictory version (\eg., changing ``left" to ``right"). Both interventions lead to similar, moderate degradation. This suggests that intermediate language provides useful context for the rollout, but the reasoning trajectory remains partially functional even when that information is removed or made inconsistent.

The visual pathway is considerably more sensitive. \textbf{Intervention \#8} removes the intermediate image before it is fed back into the subsequent generation step, while \textbf{intervention \#9} corrupts the intermediate image with strong noise. Both interventions cause substantially larger performance drops than the textual interventions, with complete removal producing the strongest degradation. Since the final output image itself is not directly modified, these effects arise specifically from disrupting the intermediate visual state available to subsequent reasoning steps.

The stronger sensitivity to intermediate visual interventions mirrors the training-time result in \cref{sec:visual_vs_text}. Together, the two analyses provide complementary evidence: visual trajectories are not only a stronger source of supervision during training, but are also actively used during inference. Explicit textual reasoning can support the rollout, but the intermediate visual state appears to serve as the more consequential working representation. This pattern is consistent with a reasoning process that maintains and updates visual states, with language playing a secondary supporting role rather than serving as the primary substrate of computation.

\subsection{Chain-of-Step: Visualization of Native Visual Reasoning}
\label{sec:baselines:cos}

\begin{figure}[t]
  \centering
  \includegraphics[width=\columnwidth]{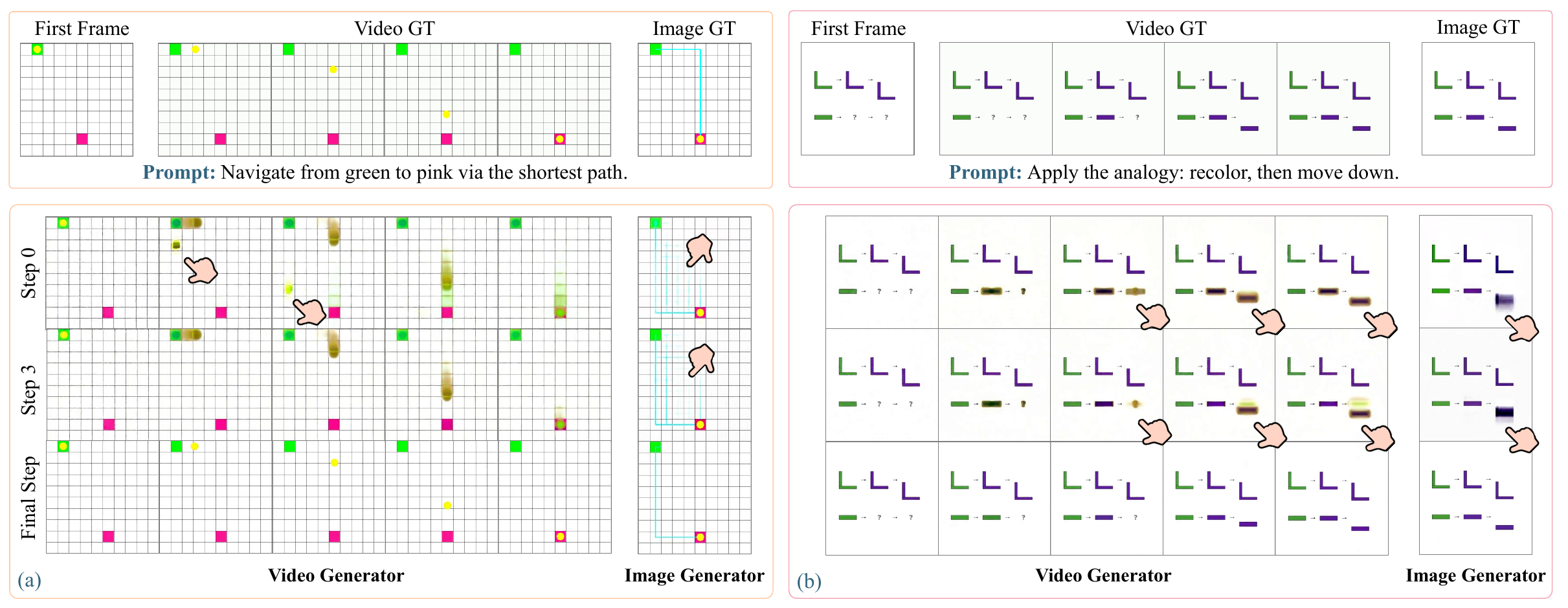}
  \caption{
    \textbf{Chain-of-Step visualization.} Following Wang \etal~\cite{wang2026demystifying}, we visualize how generative models reason through diffusion steps. Both video and image generators exhibit Chain-of-Step behaviors (see pointed), such as \textit{Multi-Path Exploration} (left) and \textit{Superposition-Based Exploration} (right), where multiple plausible solution hypotheses emerge in early denoising steps and gradually converge toward the final answer.
  }
  \label{fig:vbvrpro_cos_analysis}
\end{figure}

Following Wang \etal~\cite{wang2026demystifying}, who introduce tools for visualizing reasoning behaviors in video diffusion models, we apply the same analysis to both video and image generation models. Specifically, we decode noisy diffusion latents from different denoising steps to inspect the intermediate reasoning behaviors of models trained on \dataname. We analyze two representative models: the strongest video generation model, VBVR-Pro-Wan2.2-I2V-A14B, and the interleaved text-image generation model, VBVR-Pro-SenseNova-U1.
Across both modalities, we observe clear \textit{Chain-of-Step} behaviors~\cite{wang2026demystifying}. As shown in \cref{fig:vbvrpro_cos_analysis}(a), both models explore multiple candidate paths across denoising steps before converging to the shortest feasible route, corresponding to \textit{Multi-Path Exploration}. The video model represents this process through continuous object motion, whereas the interleaved text-image model directly draws the candidate paths in image space. In \cref{fig:vbvrpro_cos_analysis}(b), both models further exhibit \textit{Superposition-Based Exploration}, where multiple plausible visual states are overlaid during intermediate denoising steps before the model converges to a final solution.

\subsection{Transferability}
\label{sec:baselines:transferability}

\begin{figure}[th!]
  \centering
  \includegraphics[width=\columnwidth]{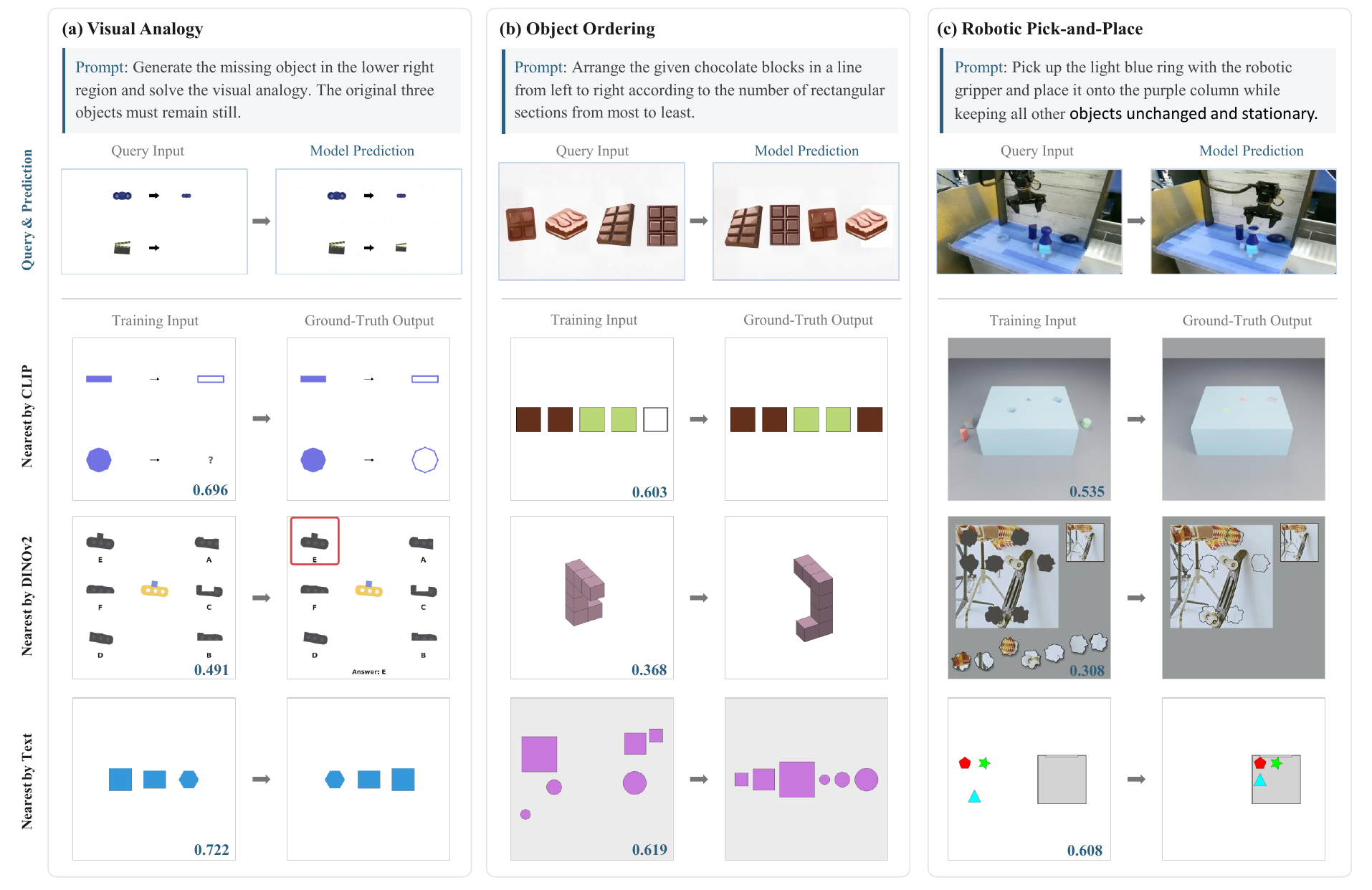}
  \caption{\textbf{Nearest-neighbor analysis against the training set.} For each query, we show the model prediction and the nearest training instances retrieved using CLIP, DINOv2, and text features. The retrieved instances are visually and semantically distinct from the queries, suggesting that the model's transferable capabilities cannot be attributed to direct memorization. Blue numbers indicate feature-space similarity.}
  \label{fig:nearest_neighbor_analysis}
\end{figure}

Among all models fine-tuned on \dataname, we evaluate the strongest model, VBVR-Pro-Wan2.2-I2V-A14B, on a suite of seven external reasoning benchmarks in~\cref{tab:video_reasoning_benchmark_results}. VBVR-Pro-Wan2.2-I2V-A14B achieves consistent improvements of up to 20 percentage points across a wide variety of unseen benchmarks. Notably, these benchmarks contain many real-world visual problems that go beyond the scope of \name, yet the model fine-tuned on \dataname still acquires transferable capabilities. For example, compared with the Wan2.2-I2V-A14B base model, VBVR-Pro-Wan2.2-I2V-A14B shows remarkable improvements on shape fitting and analogy solving in V-ReasonBench, and on embodied reasoning in MME-CoF-Pro. These tasks involve real-world scenes with everyday objects and robotic manipulators, which differs substantially from the procedurally generated abstractions in \dataname. Representative cases are shown in~\cref{fig:nearest_neighbor_analysis}, with detailed results reported in~\cref{sec:appendix_transferability}.

\definecolor{line-blue}{RGB}{235,243,250}

\begin{table*}[t]
\centering
\scriptsize
\setlength{\tabcolsep}{3pt}
\caption{\textbf{Results on external video reasoning benchmarks.} Higher scores indicate better performance. \textbf{Bold} denotes the best result, and \underline{underline} denotes the second-best result. Training on \dataname yields transferable improvements across diverse visual reasoning scenarios.
}
\resizebox{1.0\linewidth}{!}{
\begin{tabular}{l|ccccccc}
\toprule
\textbf{Models}
& \textbf{RISE-Video~\cite{risevideo_liu2026rise}}
& \textbf{V-ReasonBench~\cite{v-reasonbench_luo2025v}}
& \textbf{RULER-Bench~\cite{ruler-bench_he2025ruler}}
& \textbf{MME-COF-Pro~\cite{mme-cof-pro_qi2026mme}}
& \textbf{\shortstack{VideoThinkBench\\mini}~\cite{videothinkbench_tong2026thinking}}
& \textbf{BabyVision-Gen\cite{babyvision_chen2026babyvision}}
& \textbf{\shortstack{intelligentVBench\\(Implicit I2V)}\cite{intelligentvbench_pan2026omniweaving}} \\
\midrule

\rowcolor{gray!20}
\multicolumn{8}{l}{\textbf{Video Generation Models}} \\
\midrule

\rowcolor{line-blue}
\textbf{Proprietary Models}
& & & & & & & \\

Veo-3.1
& \textbf{76.40}
& \underline{24.25}
& \underline{65.19}
& \textbf{55.90}
& 27.69
& --
& 3.74/5 \\

\midrule

\rowcolor{line-blue}
\textbf{Open-source Models}
& & & & & & & \\

CogVideoX1.5-5B-I2V
& 46.99
& 7.87
& 42.27
& 28.46
& 22.83
& 1.07
& 3.50/5 \\

LTX-2.3-I2AV
& 55.23
& 4.51
& 49.14
& 34.05
& 22.40
& 2.14
& 3.88/5 \\

Wan2.2-I2V-A14B
& 62.83
& 10.21
& 57.89
& 24.76
& 25.71
& 0.36
& 3.93/5 \\

VBVR-Wan2.2
& 65.94
& 18.05
& 56.28
& 34.13
& \underline{30.72}
& \underline{4.64}
& \underline{4.08/5} \\

\midrule

\rowcolor{line-blue}
\textbf{Ours}
& & & & & & & \\

\textbf{VBVR-Pro-Wan2.2-I2V-A14B}
& \underline{66.18}
& \textbf{38.22}
& \textbf{66.96}
& \underline{48.39}
& \textbf{52.86}
& \textbf{12.50}
& \textbf{4.11/5} \\

\bottomrule
\end{tabular}
}
\label{tab:video_reasoning_benchmark_results}
\end{table*}

To validate if the improvements arise from memorizing training samples similar to the test set, or instead from generalizable reasoning acquired from conceptually transferable synthetic examples, we conduct a nearest-neighbor analysis. For the examples in~\cref{fig:nearest_neighbor_analysis}, we retrieve the closest training samples using CLIP~\cite{clip_radford2021learning} and DINOv2~\cite{dinov2_oquab2023dinov2} features for the initial image, and BGE-base-en-v1.5~\cite{bge_xiao2024c} embeddings for the task instruction. In these cases, the model transfers to unseen benchmark problems without visually similar training examples. Although some retrieved training examples have related task instructions, their visual appearances are substantially different from the test examples. This suggests that the model learns abstract visual concepts and operations, such as pattern finding, ordering, and object manipulation, beyond surface appearance, enabling transfer to different real-world instances.

\section{Reinforcement Learning with Verifiable Rewards}
\label{sec:rl}

A key goal of this work is to examine the feasibility of large-scale multi-task reinforcement learning (RL) for visual reasoning. We emphasize that this is not a straightforward extension of RL for general visual generation: while conventional generative tasks often reward pixel-level fidelity or aesthetics, native visual reasoning requires satisfying high-level goals expressed through semantic transformations, and task-specific constraints. In this section, we present an end-to-end RL baseline for visual reasoning, including the optimization algorithm (\cref{sec:rl:algo}), training recipes (\cref{sec:rl:training}), quantitative analysis (\cref{sec:rl:results}), and qualitative case studies (\cref{sec:rl:cases}).

\subsection{Algorithms}
\label{sec:rl:algo}

\begin{figure}[t]
\centering
\includegraphics[width=\columnwidth]{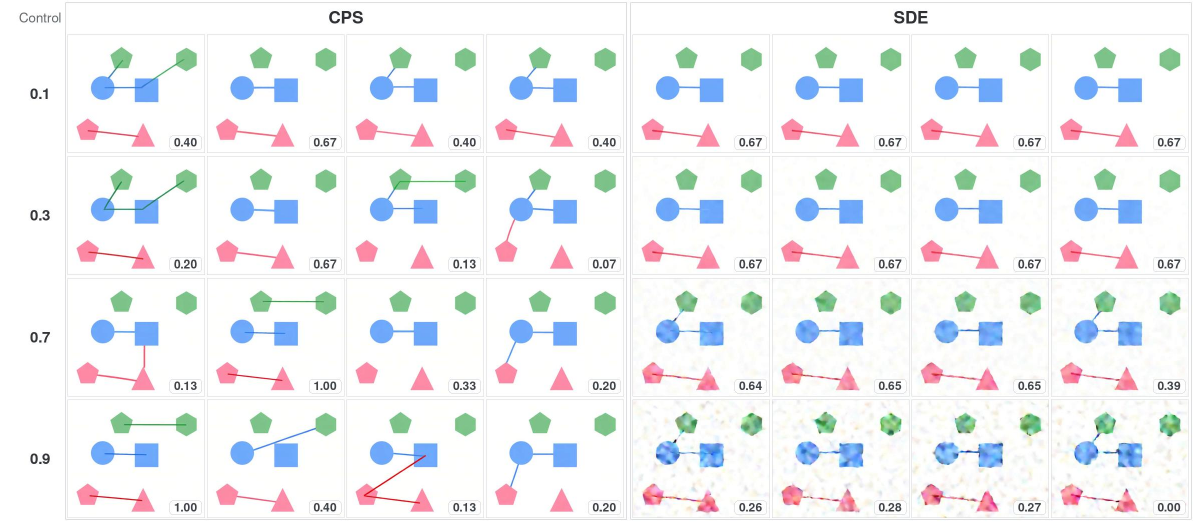}
\caption{Final-frame comparison of CPS and Flow-SDE across different stochasticity settings. Each row shows four selected rollouts generated from the same prompt; the number in each thumbnail denotes its score. Stronger CPS noise enables more diverse connection decisions while largely preserving clean shapes and backgrounds. In contrast, increasing the SDE stochasticity introduces progressively more severe visual corruption. We note that the CPS noise level and the SDE $\eta$ parameterize different sampling rules and should not be interpreted as directly calibrated stochasticity values.}
\label{fig:cps_sde_stochasticity}
\end{figure}

Group-relative optimization methods, such as Flow-GRPO~\cite{NEURIPS2025_3a10c465} and DanceGRPO~\cite{xue2025dancegrpo}, formulate reverse denoising as a sequential decision process and optimize the denoising policy using group-relative terminal rewards together with a clipped policy objective. These methods have become popular choices for RL-enhanced visual generation. A key requirement for such approaches is that sampled trajectories differ in reward-relevant decisions, so that relative rewards provide meaningful optimization signals.

In native visual reasoning, however, useful exploration is primarily \emph{semantic}. The model must make discrete, high-impact decisions, such as turning left or right in a maze, selecting an answer, or determining which objects to connect or mark. Standard stochastic perturbations often change only low-level appearance while preserving the same underlying decision, providing little useful variation for group-relative reward estimation. As illustrated in \cref{fig:cps_sde_stochasticity}, low-stochasticity rollouts frequently converge to similar outputs. We therefore hypothesize that stronger stochasticity is necessary for effective RL exploration in native visual reasoning tasks.

At the same time, naively increasing stochasticity can severely damage visual fidelity. Conventional Flow-SDE sampling exhibits an unfavorable trade-off between exploration and visual quality. As analyzed by~\cite{wang2025coefficients}, newly injected Gaussian noise is not exactly compensated by the scheduler-prescribed predicted-noise component, allowing the effective noise magnitude to exceed its intended level. This behavior is visible in \cref{fig:cps_sde_stochasticity}: increasing the SDE stochasticity to $\eta=0.7$ introduces substantial background artifacts and distorted object boundaries, while $\eta=0.9$ severely corrupts generated frames and reduces rewards.

In contrast, Coefficients-Preserving Sampling (CPS)~\cite{wang2025coefficients} provides a better exploration--fidelity trade-off. Stronger CPS noise induces visibly different task-level decisions and a wider range of rewards while largely preserving clean shapes and backgrounds. In~\cref{fig:cps_sde_stochasticity}, CPS continues to produce visually clean rollouts even at high stochasticity, while supporting diverse semantic decisions. These observations motivate our use of CPS for RL exploration. We therefore adopt CPS in our RL baseline:
\begin{equation}
x_{t-\Delta t}
=
\left(1-(t-\Delta t)\right)\widehat{x}_{0}
+
(t-\Delta t)\cos\left(\frac{\eta\pi}{2}\right)\widehat{x}_{1}
+
(t-\Delta t)\sin\left(\frac{\eta\pi}{2}\right)\epsilon,
\end{equation}
where $\widehat{x}_{0}$ and $\widehat{x}_{1}$ denote the predicted clean sample and predicted noise, respectively, $\epsilon\sim\mathcal{N}(0,I)$, and $\eta\in[0,1]$ controls the stochasticity. When $\eta=0$, CPS reduces to deterministic ODE sampling. Crucially, the coefficients of the predicted and newly sampled noise satisfy
$\cos^2(\eta\pi/2)+\sin^2(\eta\pi/2)=1$.
Thus, increasing $\eta$ replaces part of the predicted noise with newly sampled Gaussian noise without increasing the total noise coefficient prescribed by the scheduler. CPS can therefore introduce sufficient stochasticity for exploring alternative visual decisions while avoiding excessive noise accumulation, thereby preserving rollout quality and providing more reliable reward signals.

Detailed descriptions of the RL implementation, including optimization strategies, training configurations, and specific hyperparameter settings, are provided in \cref{sec:rl_details}.

\subsection{Data and Training Recipes}
\label{sec:rl:training}

All experiments in this section start from VBVR-Pro-Wan2.2-TI2V-5B and use the same 50K RL subset of the VBVR-Pro Dataset, spanning 50 in-domain tasks. We refer to the original VBVR-Pro-Wan2.2-TI2V-5B checkpoint, before applying any of the training strategies below, as the baseline. This controlled setup isolates the effects of the training objective and reward signal.

\textbf{Supervised Fine-Tuning (SFT).}
We fine-tune the model directly on the 50K training instances using a supervised objective, providing a non-RL baseline under the same data budget.

\textbf{Reinforcement Learning with VLM Rewards (RLVLM).}
This variant uses Qwen3.6-27B~\cite{qwen3.6-27b} as a VLM judge. The judge evaluates each generated sample and provides the reward used for policy optimization.

\textbf{Reinforcement Learning with Verifiable Rewards (RLVR).}
This variant uses the proposed VBVR Scorer, which deterministically computes task-specific verifiable rewards from the underlying ground-truth metadata.

RLVLM and RLVR use the same RL algorithm, rollout configuration, optimization hyperparameters, training data, and model initialization; they differ only in the source of the reward signal. Therefore, their comparison directly measures the effect of replacing a general-purpose VLM judge with the proposed task-grounded verifiable reward. We use the terms RLVLM and RLVR to denote these two reward variants throughout the remainder of the paper.

\subsection{Results and Analysis}
\label{sec:rl:results}

\begin{table*}[t]
\centering
\scriptsize
\setlength{\tabcolsep}{3pt}
\caption{
\textbf{Reinforcement learning results}. We report the overall, In-Domain (ID), and Out-of-Domain (OOD) scores together with category-wise performance.
Detailed inference configurations are described in \cref{sec:rl_inference}, and results with additional samplers are provided in \cref{sec:rl_sampler_sensitivity}.
Higher is better.
\textbf{Bold}: best; \underline{underline}: second best.
Baseline refers to VBVR-Pro-Wan2.2-TI2V-5B.
*Note the baseline has already undergone SFT on the main training set; SFT here further uses the RL training set.
}
\label{tab:vbvr_rl_results}

\begin{tabular}{l|c|c|ccccc|c|ccccc}
\toprule
& \multicolumn{1}{c|}{}
& \multicolumn{6}{c|}{\textbf{In-Domain}}
& \multicolumn{6}{c}{\textbf{Out-of-Domain}} \\
\cmidrule(lr){3-3}
\cmidrule(lr){4-8}
\cmidrule(lr){9-9}
\cmidrule(lr){10-14}
\textbf{Model}
& \textbf{Overall}
& \textbf{Avg.}
& \textbf{Abst.}
& \textbf{Know.}
& \textbf{Perc.}
& \textbf{Spat.}
& \textbf{Trans.}
& \textbf{Avg.}
& \textbf{Abst.}
& \textbf{Know.}
& \textbf{Perc.}
& \textbf{Spat.}
& \textbf{Trans.} \\
\midrule


Baseline
& 0.470
& 0.641
& 0.578
& 0.511
& 0.476
& 0.480
& 0.724
& 0.300
& \underline{0.585}
& 0.153
& 0.129
& 0.311
& 0.352 \\

 + SFT*
& 0.503
& \underline{0.679}
& 0.671
& 0.637
& 0.524
& \underline{0.687}
& 0.754
& 0.328
& 0.452
& 0.185
& 0.214
& \textbf{0.555}
& \textbf{0.450} \\

\midrule

 + RLVR
& \textbf{0.548}
& \textbf{0.719}
& \textbf{0.743}
& \textbf{0.689}
& \textbf{0.657}
& \textbf{0.709}
& \underline{0.815}
& \textbf{0.377}
& 0.511
& \textbf{0.282}
& \textbf{0.288}
& \underline{0.551}
& 0.388 \\

 + RLVR-SDE
& 0.504
& 0.668
& 0.663
& 0.599
& \underline{0.650}
& 0.583
& \textbf{0.892}
& 0.340
& \textbf{0.641}
& 0.141
& 0.213
& 0.515
& 0.349 \\

 + RLVLM
& \underline{0.508}
& 0.671
& \underline{0.679}
& \underline{0.667}
& 0.601
& 0.652
& 0.771
& \underline{0.345}
& 0.468
& \underline{0.260}
& \underline{0.256}
& 0.467
& \underline{0.389} \\

\bottomrule
\end{tabular}
\end{table*}

\cref{tab:vbvr_rl_results} compares the strong baseline with models trained using supervised fine-tuning (SFT), RLVLM, and RLVR. The SFT and RL models are trained independently from the same initialization using the same training data, enabling a controlled comparison of their training objectives. For a matched evaluation, both RL models are sampled using CPS with $\eta=0.7$; we defer a comprehensive comparison of inference samplers to \cref{sec:rl_sampler_sensitivity}. Under this shared inference configuration, RLVR achieves the strongest performance, reaching an overall score of 0.548, compared with 0.508 for RLVLM and 0.503 for SFT. RLVLM provides only a modest overall improvement, whereas RLVR produces a substantially larger gain.

The Out-of-Domain results are particularly noteworthy. Although no Out-of-Domain tasks are used during RL training, both RL objectives improve over SFT, with RLVLM and RLVR reaching 0.345 and 0.377, respectively, compared with 0.328 for SFT. These gains suggest that reward-guided optimization can improve generalization beyond the RL training distribution. The additional improvement obtained by RLVR further indicates that task-grounded verifiable feedback can provide a stronger signal for transferable visual reasoning.

\begin{figure}[t]
\centering
\includegraphics[width=\columnwidth]{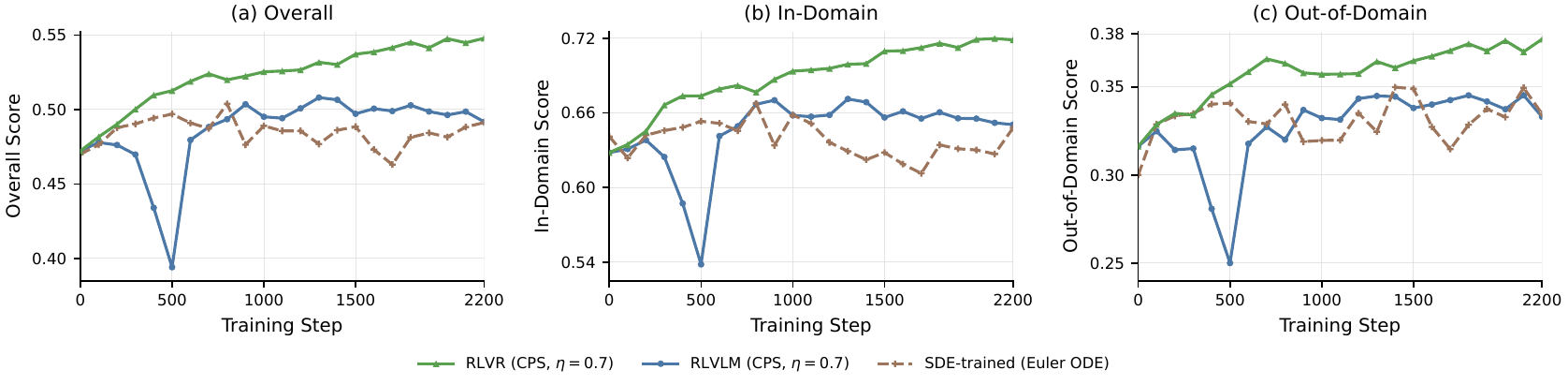}
\caption{\textbf{Evaluation performance during RL.} We report overall, in-domain, and out-of-domain scores every 100 training steps using CPS with $\eta=0.7$. RLVR shows steady performance improvements, whereas RLVLM soon stagnates. The sudden drop of RLVLM at 400-500 steps is investigated in~\cref{sec:vlm_judge_instability}.}
\label{fig:rl_training_eval}
\end{figure}

\cref{fig:rl_training_eval} tracks evaluation performance under the same CPS configuration used in \cref{tab:vbvr_rl_results}. RLVR exhibits a clear and sustained upward trend across the overall, in-domain, and out-of-domain metrics. RLVLM yields smaller and less stable gains, including a temporary performance drop around Steps $400\sim 500$. We investigate this instability in \cref{sec:vlm_judge_instability} and analyze the effects of CPS stochasticity and deterministic ODE solvers separately in \cref{sec:rl_sampler_sensitivity} of the appendix.

\subsection{Case Studies}
\label{sec:rl:cases}

Aggregate benchmark scores show that RLVR improves final-task performance, but they do not reveal how RL changes the underlying visual reasoning trajectory. Following Wang~\etal~\cite{wang2026demystifying}, we visualize the model's instantaneous prediction of the final video at selected points along the denoising trajectory. We compare the SFT-initialized policy used as the starting point for RL, referred to as the pre-RL model, with the resulting RLVR model. The following examples illustrate two complementary changes in reasoning behavior: continued revision across denoising steps and more coherent execution of a multi-step spatial plan.

\begin{figure}[t]
\centering
\includegraphics[width=\columnwidth]{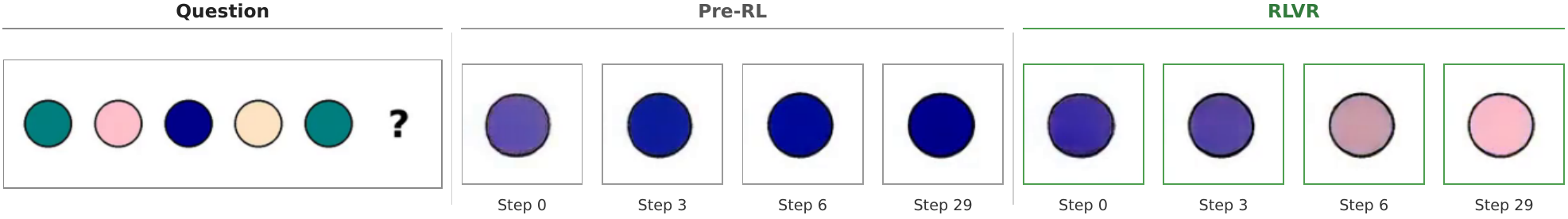}
\caption{\textbf{Self-correction in a color-sequence completion task.} The question is shown on the left, followed by predictions from the pre-RL and RLVR models at Steps 0, 3, 6, and 29. For clarity, video frames are cropped to the predicted answer region. The pre-RL model commits to an incorrect blue answer early in denoising and retains it through the final step. In contrast, RLVR revises its initially incorrect purple prediction toward the target color at Step 6 and reaches the correct pink answer at Step 29.}
\label{fig:rl_better_reasoning}
\end{figure}

In the color-sequence task shown in \cref{fig:rl_better_reasoning}, the pre-RL model determines its answer within the first few denoising steps and does not reconsider it during the remainder of the trajectory. RLVR is also incorrect initially, but its prediction remains revisable: the predicted color changes at Step 6 and subsequently converges to the correct answer. Because both models use the same 30-step sampling schedule, the observed difference cannot be attributed to additional inference computation. Instead, RLVR continues to perform task-relevant refinement beyond the initial denoising steps, providing an example of later-step self-correction.

\begin{figure}[t]
\centering
\includegraphics[width=1.0\columnwidth]{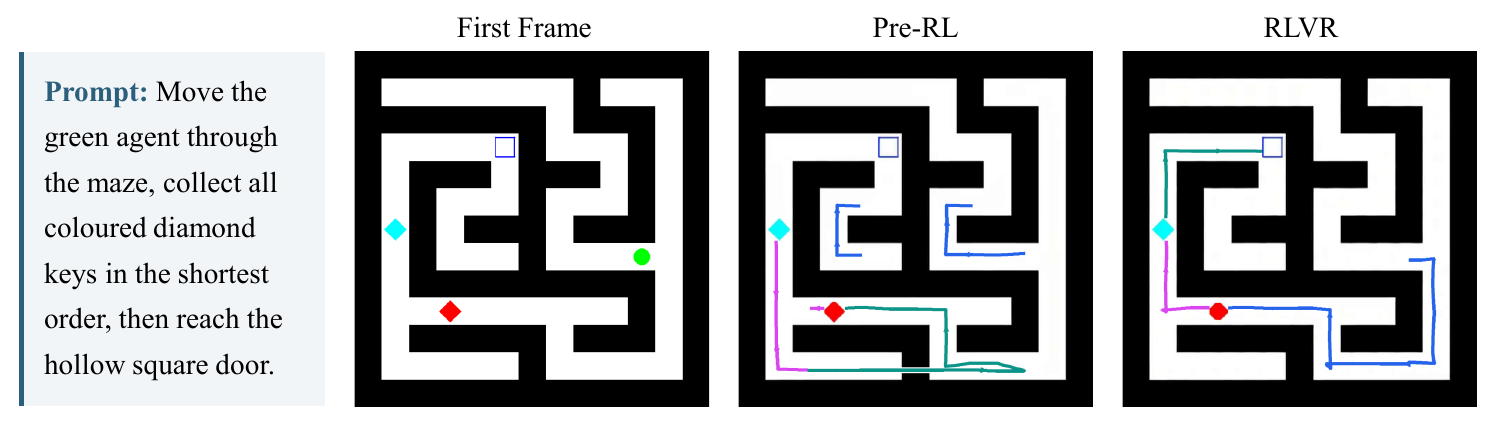}
\caption{\textbf{More coherent maze navigation after RLVR.} Each panel summarizes all 81 frames of the corresponding generated video as a stage-wise directed trajectory. Blue, magenta, and teal denote the Start-to-Red, Red-to-Cyan, and Cyan-to-Goal stages, respectively, while arrowheads indicate motion direction. The pre-RL model produces disconnected and inconsistent movements that do not form a valid solution (score: 0.2249), whereas RLVR follows a coherent route through the maze and reaches the target (score: 0.8692). The trajectory-rendering procedure is described in \cref{sec:temporal_trajectory_visualization}.}
\label{fig:rl_better_reasoning_maze}
\end{figure}

The maze example in \cref{fig:rl_better_reasoning_maze} demonstrates a different form of improvement. The pre-RL model places the agent at disconnected locations across frames and fails to maintain a valid route through the maze. In contrast, RLVR produces a more temporally consistent sequence of agent states, largely follows connected corridors, and eventually reaches the target. The improvement therefore concerns not only the final state but also the coherence of the intermediate trajectory, suggesting a stronger ability to maintain and execute a multi-step spatial plan.

Together, these examples illustrate two complementary effects of RLVR. First, it extends the effective reasoning horizon within a fixed denoising budget, allowing the model to revise decisions that would otherwise become fixed during the initial steps. Second, it improves the temporal consistency of state transitions across generated video frames. Additional qualitative examples are provided in \cref{sec:rl_makes_reasoning_step_better}.
\section{Related Work}

\paragraph{Native visual reasoning through visual generation.}
Native visual reasoning, broadly defined as reasoning through visual generation, has recently received increasing attention. Wiedemer \etal~\cite{wiedemer2025video} observed that video models are reasoners, and identified the \textit{Chain-of-Frames} (CoF) paradigm. Following this direction, OpenCoF~\cite{chen2026opencof} introduces learnable reasoning tokens in the video generators. Wang \etal~\cite{wang2026demystifying} further identify a different mechanism, \textit{Chain-of-Steps} (CoS), where reasoning unfolds through diffusion denoising steps, a phenomenon further supported by later studies~\cite{newman2026videomodelsreasonearly}. Despite the vision-centric nature of this paradigm, MLLMs have also been incorporated to assist video reasoners. Cheng \etal~\cite{cheng2026vlmteacher} use MLLMs for test-time optimization of video reasoners, while VChain~\cite{huang2026vchain} injects MLLM-derived reasoning signals, \ie Chain-of-Visual-Thought, to guide video generation at inference time.
Beyond video, image generation has also been explored as a substrate for native visual reasoning. DiffThinker~\cite{he2025diffthinker} reformulates visual reasoning as a native image-to-image generative task across four visual reasoning tasks, while EndoCoT~\cite{dai2026endocot} refines latent thoughts to guide image generation. However, existing studies are typically validated on a limited set of visual tasks, largely due to the lack of unified infrastructure that supports large-scale diverse training, verifiable evaluation, and systematic comparison across generative modalities. This is the gap we aim to address in this work.

\paragraph{Interleaved visual-textual reasoning.}
The success of language-based reasoning has naturally motivated interleaving intermediate visual states with textual rationales~\cite{dong2026igcotsurvey}. Within the broader paradigm of reasoning through generation, recent methods enable models to sketch with external tools~\cite{hu2024sketchpad}, visualize reasoning traces~\cite{li2025mvot}, incorporate visual editing into a reasoning harness~\cite{zhang2026etchr}, or learn from large-scale interleaved corpora~\cite{zebracot_li2025zebra,structcot_wang2026deltav,gu2025thinkmorph}. A parallel line of work investigates how much explicit visual state is needed, compressing visual reasoning into update tokens~\cite{structcot_wang2026deltav} or latent tokens~\cite{yang2026mirage,dong2026ilvr,chen2026ivtlr}. Beyond interleaved text-image modeling, SC-CMJP~\cite{le2026concurrent} leverages masked diffusion models to jointly generate text and images for editing and visual reasoning tasks. While these approaches significantly expand the design space of visual reasoning, they often inherit a language-centric view in which visual states serve as auxiliary supports for textual reasoning. In contrast, our work systematically investigates visual trajectories themselves as first-class reasoning substrates.

\paragraph{Reinforcement learning for visual reasoning.}
Reinforcement learning for general visual generation has been extensively studied~\cite{NEURIPS2025_3a10c465,xue2025dancegrpo,zheng2025diffusionnft,wang2025coefficients,cheng2025realdpo,bu2026adagrpo}, primarily to improve instruction following and aesthetic quality. However, visual reasoning tasks require models to reach semantically valid solutions; RL for such visual reasoning objectives remains comparatively underexplored. For interleaved generation, recent approaches jointly optimize text and image policies under unified RL objectives~\cite{liu2026unigrpo,hu2026braid,cheng2026omnir1}, but their visual states are still commonly evaluated by VLM-based judges. VPRL~\cite{xu2026visualplanning} improves large vision models on three visual navigation tasks without language, while VideoRLVR~\cite{zhu2026video} demonstrates the promise of verifiable rewards on three visual tasks. Nevertheless, existing studies remain limited in scale and task diversity. In this work, we perform joint RL training across 50 tasks with verifiable rewards and evaluate the resulting models on 100 tasks, enabling a substantially broader study of multi-task native visual reasoning.

\paragraph{Visual reasoning benchmarks and training sources.}
Existing visual reasoning benchmarks span video generation~\cite{videothinkbench_tong2026thinking, v-reasonbench_luo2025v, ruler-bench_he2025ruler, risevideo_liu2026rise,intelligentvbench_pan2026omniweaving}, image editing~\cite{risebench_zhao2026envisioning,kris_wu2026kris,babyvision_chen2026babyvision,grade_liu2026grade}, and unified multimodal generation~\cite{opening_zhou2025opening,uni-mmmu_zou2026uni,mira_zhou2026visualizing,realunify_shi2026realunify}. However, most of these benchmarks are evaluation-only~(\cref{tab:datasets_comparison}): they offer little supervision for training and often rely on VLM-based judges, whose evaluations can be difficult to verify and reproduce. A few interleaved corpora reach training scale~\cite{zebracot_li2025zebra,structcot_wang2026deltav}, but they remain limited in task coverage and, more importantly, are not procedurally generated, making it difficult to obtain verifiable evaluation signals or task-grounded rewards. VBVR~\cite{vbvr2026} provides both large-scale supervision and verifiable evaluation, but its task space can be further expanded toward broader, more challenging, and more transferable visual reasoning capabilities. \name builds on this foundation by providing a larger controlled task space, verifiable reward scorers, multi-task RL training, and systematic modality comparisons across image, video, and interleaved generative models.

\section{Conclusion}
By unifying procedurally generated visual reasoning environments, task-grounded verifiable rewards, reinforcement learning, and modality-controlled mechanism analyses, \name establishes native visual reasoning through generation as a trainable, verifiable, and experimentally controllable research problem. Our transfer studies, counterfactual diagnostics, visual-state interventions, trajectory visualizations, and reward-driven optimization collectively suggest that visual generative models can construct, maintain, and refine intermediate visual trajectories, positioning visual generation not merely as an output format but as a genuine substrate for reasoning. These findings call for greater attention to native visual reasoning as a complementary paradigm to language-centric reasoning, and we hope \name provides a useful foundation for the future research.

\section*{Acknowledgements}
We gratefully acknowledge Xintao Wang, Yinfei Yang and Andrew Dai for their valuable discussions on visual reasoning and insightful feedback on this work, and Huimuk Jang for creating the teaser figure.

We also thank Amazon Web Services for supporting this work through the AWS Trainium for Research program (\url{https://aws.amazon.com/ai/machine-learning/trainium/research/}).

\clearpage


\bibliography{main}
\bibliographystyle{icml2026}

\newpage
\appendix
\onecolumn

\section{Details of baseline model architectures}
\label{sec:architecture_details}

\subsection{Image, Video, and Interleaved Generators}
To evaluate the effectiveness of training on the \dataname, we fine-tune a diverse set of generative foundation models spanning image generation, video generation, and interleaved text-image generation. Their architectural properties are summarized in~\cref{tab:base_model_architectures}, with additional model-specific details provided in~\cref{sec:architecture_details}. They vary substantially in model scale, cross-modal interaction mechanisms, and output representations. This architectural diversity enables us to examine whether the benefits of fine-tuning on \dataname generalize across different generative modalities and model designs, including sparse and dense architectures as well as latent- and pixel-space generation.

\begin{itemize}

\item \textbf{Image generation models.}
\begin{itemize}
    \item \textbf{BAGEL-7B-MoT}~\cite{deng2025bagel} is a unified multimodal model that supports both multimodal understanding and image generation within a Mixture-of-Transformers (MoT) architecture. Its language-understanding and visual-generation pathways use separate Transformer parameters while participating in a common multimodal sequence, allowing the model to specialize computation by token type. The full checkpoint contains approximately 14B parameters, but only about 7B are active for a given token pathway.

    \item \textbf{Qwen-Image-Edit}~\cite{wu2025qwenimagetechnicalreport} is built on the 20B-parameter Qwen-Image foundation model and uses a multimodal diffusion Transformer (DiT) to perform both image generation and instruction-guided editing. Its dual-stream blocks initially maintain distinct representations for visual latents and multimodal conditioning while enabling interaction between them throughout denoising.

    \item \textbf{FLUX.2-dev}~\cite{flux-2-2025}'s Transformer begins with dual-stream blocks that process visual and conditioning tokens separately and then transitions to single-stream blocks that jointly process the concatenated token sequence. Compared with earlier FLUX architectures, FLUX.2 allocates a larger fraction of its depth and parameters to the joint single-stream stage, providing greater capacity for cross-modal composition and reference-image integration.
\end{itemize}

\item \textbf{Interleaved text-image generation models.}
\begin{itemize}
    \item \textbf{ThinkMorph-7B}~\cite{gu2025thinkmorph} is initialized from BAGEL-7B-MoT and fine-tuned on text-image reasoning trajectories. It therefore retains BAGEL's 14B-total, 7B-active Mixture-of-Transformers architecture and VAE-latent image-generation pathway.

    \item \textbf{SenseNova-U1-8B-MoT}~\cite{sensenova_u1} is a native unified model that treats multimodal understanding and generation as operations within a common Transformer rather than as separate encoder and decoder pipelines. Its NEO-unify architecture removes both the conventional visual encoder and the generative VAE. Images are instead converted directly into RGB pixel patches, processed alongside language tokens using modality-specific Transformer parameters, and reconstructed through a lightweight pixel-output head.
\end{itemize}

\item \textbf{Video generation models.}
\begin{itemize}
    \item \textbf{Wan2.2-I2V-A14B}~\cite{wan2025wan} is a sparse Mixture-of-Experts model with separate high-noise and low-noise denoisers. Each expert is a full 14B-class video DiT, but only one expert is active at each point in the denoising trajectory.

    \item \textbf{Wan2.2-TI2V-5B}~\cite{wan2025wan} is a dense video DiT. In contrast to the A14B variant, it uses the same 5B-parameter backbone throughout the denoising trajectory rather than routing computation between noise-specialized experts.

    \item \textbf{Wan2.1-I2V-14B-720P}~\cite{wan2025wan} is a dense 14B-parameter video DiT that applies a single 40-block Transformer backbone at all denoising timesteps and generates videos as compressed spatiotemporal VAE latents.

    \item \textbf{LTX-2.3-I2AV}~\cite{ltx23} is a unified audiovisual diffusion model that jointly generates synchronized video and audio. Its backbone is an asymmetric dual-stream Transformer, assigning substantially more capacity to the video stream than to the audio stream to reflect their differing information densities. The streams operate on separate video and audio VAE latents and exchange information through bidirectional audiovisual cross-attention, temporal positional encoding, and cross-modal adaptive normalization.
\end{itemize}
\end{itemize}

\begin{table*}[t] 
\centering 
\scriptsize
\setlength{\tabcolsep}{2pt} 
\renewcommand{\arraystretch}{1.18}
\caption{Architectural comparison of the base models used for fine-tuning. Parameter counts are reported as total, active parameters where applicable.} 
\label{tab:base_model_architectures} 
\begin{tabularx}{\textwidth}{ P{2.6cm} P{2.2cm} P{3.5cm} Y P{2cm} }
\toprule
\textbf{Model} &
\textbf{Parameters} &
\textbf{Generative backbone} &
\textbf{Transformer configuration} &
\textbf{Generation representation} \\
\midrule 
\rowcolor{line-blue} 
\multicolumn{5}{l}{\textbf{Image generation models}} \\
BAGEL-7B-MoT
\cite{deng2025bagel} &
14B total; 7B active &
MoT with separate understanding
and generation branches &
28 blocks; width 3,584; 28 heads
$\times$ 128 dims/head &
VAE latents \\

FLUX.2-dev
\cite{flux-2-2025} &
32B &
Hybrid dual-stream and single-stream DiT &
8 dual-stream + 48 single-stream blocks;
width 6,144; 48 heads $\times$ 128 dims/head &
VAE latents \\

Qwen-Image-Edit
\cite{wu2025qwenimagetechnicalreport} &
20B &
Dual-stream multimodal diffusion Transformer (MMDiT) &
60 dual-stream blocks; width 3,072; 24 heads
$\times$ 128 dims/head  &
VAE latents \\

\midrule 
\rowcolor{line-blue} 
\multicolumn{5}{l}{\textbf{Interleaved image generation models}} \\

ThinkMorph-7B
\cite{gu2025thinkmorph} &
14B total; 7B active &
MoT with separate understanding
and generation branches &
28 blocks; width 3,584; 28 heads
$\times$ 128 dims/head &
VAE latents \\

SenseNova-U1-8B-MoT
\cite{sensenova_u1} &
16B total; 8B active &
MoT with separate understanding
and generation branches &
42 blocks; width 4,096; 32 heads
$\times$ dims/head &
Raw RGB patches \\

\midrule
\rowcolor{line-blue}
\multicolumn{5}{l}{\textbf{Video generation models}} \\
Wan2.2-I2V-A14B
\cite{wan2025wan} &
27B total; 14B active  &
Two-expert MoE video DiT with separate high-/low-noise experts &
40 blocks per expert; width 5,120; 40 heads
$\times$ 128 dims/head  &
VAE latents \\

Wan2.2-TI2V-5B
\cite{wan2025wan} &
5B &
Dense video DiT &
30 blocks; width 3,072; 24 heads
$\times$ 128 dims/head  &
VAE latents \\

Wan2.1-I2V-14B-720P
\cite{wan2025wan} &
14B &
Dense video DiT &
40 blocks; width 5,120; 40 heads
$\times$ 128 dims/head  &
VAE latents \\

LTX-2.3-I2AV
\cite{ltx23} &
22B &
Asymmetric dual-stream audio--video DiT &
48 blocks; width 4,096; 32 heads
$\times$ 128 dimensions &
VAE latents \\

\bottomrule
\end{tabularx}

\end{table*}

\subsubsection{Training Recipes}
\label{sec:training_details}
All models are fine-tuned for one epoch at a resolution of $512\times512$. We adopt a unified fine-tuning strategy based on model scale and modality. For image generation and interleaved text-image generation models, models with fewer than 10B active parameters are fully fine-tuned, while larger models are adapted with LoRA. For all video generation models, we consistently use LoRA fine-tuning.

For image generation, Qwen-Image-Edit and FLUX.2-dev are adapted using rank-32 LoRA with a learning rate of $1\times10^{-4}$, whereas BAGEL is fully fine-tuned with a smaller learning rate of $5\times10^{-6}$. For Qwen-Image-Edit and FLUX.2-dev, the LoRA adapters are applied to the query, key, value, and output projections, as well as selected multimodal projection, feed-forward, and modulation layers. For interleaved text-image generation, ThinkMorph and SenseNova-U1 are fully fine-tuned using learning rates of $1\times10^{-5}$ and $2\times10^{-5}$, respectively. For video generation, we apply rank-32 LoRA to all evaluated models, including Wan2.2-I2V-A14B, Wan2.2-TI2V-5B, Wan2.1-I2V-14B-720P, and LTX-2.3-I2AV. The adapters are applied to the query, key, value, and output projections, with the feed-forward projections additionally adapted for the Wan models. All video models are trained at $16$ fps with a learning rate of $1\times10^{-4}$.

Specifically, to isolate and assess the contribution of the \dataname training data, VBVR-SenseNova-U1 and VBVR-Pro-SenseNova-U1 are initialized from an unreleased pretrained checkpoint that has never been exposed to VBVR data during pretraining. This setup allows us to more cleanly attribute the resulting performance gains to VBVR/VBVR-Pro fine-tuning.

\subsection{Details of Transferability to Other Benchmarks}
\label{sec:appendix_transferability}

\begin{table*}[t]
\centering
\scriptsize
\setlength{\tabcolsep}{3pt}
\caption{Per-knowledge-type results on \textbf{RISE-Video}. WS: weighted score; Acc.: accuracy. \textbf{Bold}: best result; \underline{underline}: second best.}
\resizebox{1.0\linewidth}{!}{
\begin{tabular}{l|cccccccccccccccc}
\toprule
\multirow{2}{*}{\textbf{Models}} & \multicolumn{2}{c}{\textbf{Commonsense}} & \multicolumn{2}{c}{\textbf{Subject}} & \multicolumn{2}{c}{\textbf{Perceptual}} & \multicolumn{2}{c}{\textbf{Societal}} & \multicolumn{2}{c}{\textbf{Logical}} & \multicolumn{2}{c}{\textbf{Experiential}} & \multicolumn{2}{c}{\textbf{Spatial}} & \multicolumn{2}{c}{\textbf{Temporal}} \\
\cmidrule(lr){2-3} \cmidrule(lr){4-5} \cmidrule(lr){6-7} \cmidrule(lr){8-9} \cmidrule(lr){10-11} \cmidrule(lr){12-13} \cmidrule(lr){14-15} \cmidrule(lr){16-17}
 & \textbf{WS} & \textbf{Acc.} & \textbf{WS} & \textbf{Acc.} & \textbf{WS} & \textbf{Acc.} & \textbf{WS} & \textbf{Acc.} & \textbf{WS} & \textbf{Acc.} & \textbf{WS} & \textbf{Acc.} & \textbf{WS} & \textbf{Acc.} & \textbf{WS} & \textbf{Acc.} \\
\midrule
\rowcolor{gray!20}
\multicolumn{17}{l}{\textbf{Video Generation Models}} \\
\midrule
\rowcolor{line-blue}
\textbf{Open-source Models} & & & & & & & & & & & & & & & & \\
CogVideoX1.5-5B-I2V & 53.54 & 1.20 & 53.01 & 1.28 & 57.33 & 2.27 & 47.69 & 0.00 & 33.38 & 2.38 & 40.80 & 1.32 & 34.50 & 2.00 & 47.86 & 6.56 \\
LTX-2.3-I2AV & 60.85 & \underline{6.02} & 63.62 & \underline{7.69} & 64.75 & 9.09 & 55.45 & 0.00 & 46.92 & 2.38 & 51.44 & 1.32 & 36.05 & 0.00 & 56.04 & 6.56 \\
Wan2.2-I2V-A14B & \textbf{65.24} & \textbf{7.23} & 64.31 & 3.85 & 70.66 & \underline{11.36} & \underline{69.43} & \underline{3.03} & 46.52 & \underline{4.76} & 64.20 & 6.58 & 54.05 & \underline{10.00} & \underline{64.15} & 6.56 \\
VBVR-Wan2.2 & \underline{61.58} & 4.82 & \underline{66.64} & \underline{7.69} & \textbf{71.70} & 6.82 & 68.94 & \textbf{9.09} & \underline{65.24} & \textbf{26.19} & \underline{66.32} & \underline{7.89} & \textbf{66.18} & \textbf{16.00} & \textbf{64.62} & \textbf{14.75} \\
\midrule
\rowcolor{line-blue}
\textbf{Ours} & & & & & & & & & & & & & & & & \\
\textbf{VBVR-Pro-Wan2.2-I2V-A14B} & 61.37 & \textbf{7.23} & \textbf{67.90} & \textbf{11.54} & \underline{71.34} & \textbf{18.18} & \textbf{71.40} & \textbf{9.09} & \textbf{73.12} & \textbf{26.19} & \textbf{68.90} & \textbf{9.21} & \underline{61.65} & \underline{10.00} & 60.22 & \underline{8.20} \\
\bottomrule
\end{tabular}
}
\label{tab:appendix_rise_video}
\end{table*}

\begin{table*}[t]
\centering
\scriptsize
\setlength{\tabcolsep}{3pt}
\caption{Results on \textbf{RULER-Bench} (I2V setting). IF: Instruction Following; VC: Visual Consistency; VF: Visual Fidelity; RC: Rule Coherence.}
\resizebox{1.0\linewidth}{!}{
\begin{tabular}{l|cccccccccccccccccccccccc}
\toprule
\multirow{2}{*}{\textbf{Models}} & \multicolumn{5}{c}{\textbf{Science Rule}} & \multicolumn{5}{c}{\textbf{Game Rule}} & \multicolumn{5}{c}{\textbf{Humanity Rule}} & \multicolumn{4}{c}{\textbf{Vision Rule}} & \multicolumn{5}{c}{\textbf{Overall}} \\
\cmidrule(lr){2-6} \cmidrule(lr){7-11} \cmidrule(lr){12-16} \cmidrule(lr){17-20} \cmidrule(lr){21-25}
 & \textbf{IF} & \textbf{VC} & \textbf{VF} & \textbf{RC} & \textbf{Avg} & \textbf{IF} & \textbf{VC} & \textbf{VF} & \textbf{RC} & \textbf{Avg} & \textbf{IF} & \textbf{VC} & \textbf{VF} & \textbf{RC} & \textbf{Avg} & \textbf{VC} & \textbf{VF} & \textbf{RC} & \textbf{Avg} & \textbf{IF} & \textbf{VC} & \textbf{VF} & \textbf{RC} & \textbf{Avg} \\
\midrule
\rowcolor{gray!20}
\multicolumn{25}{l}{\textbf{Video Generation Models}} \\
\midrule
\rowcolor{line-blue}
\textbf{Proprietary Models} & & & & & & & & & & & & & & & & & & & & & & & & \\
Veo-3.1 & \textbf{65.05} & \textbf{83.18} & \textbf{91.37} & \textbf{50.97} & \textbf{72.64} & \underline{39.75} & 51.45 & 77.95 & 17.70 & 46.71 & \textbf{79.90} & \underline{87.37} & 94.49 & \textbf{61.23} & \textbf{80.75} & 59.53 & 72.67 & \textbf{48.94} & 60.38 & \textbf{61.57} & 70.38 & 84.12 & \textbf{44.71} & \underline{65.19} \\
\midrule
\rowcolor{line-blue}
\textbf{Open-source Models} & & & & & & & & & & & & & & & & & & & & & & & & \\
CogVideoX1.5-5B-I2V & 30.16 & 53.25 & 80.51 & 13.08 & 44.25 & 23.27 & 64.05 & 73.40 & 9.01 & 42.43 & 22.22 & 66.67 & 73.15 & 19.81 & 45.46 & 48.50 & 54.06 & 20.01 & 40.86 & 25.22 & 58.12 & 70.28 & 15.48 & 42.27 \\
LTX-2.3-I2AV & 44.20 & 70.09 & 80.06 & 18.10 & 53.11 & 22.80 & 59.23 & 73.58 & 11.90 & 41.88 & 38.89 & 66.67 & 83.33 & 40.19 & 57.27 & 56.84 & 58.97 & 26.14 & 47.32 & 35.30 & 63.21 & 73.99 & 24.08 & 49.14 \\
Wan2.2-I2V-A14B & \underline{55.37} & 81.64 & 88.14 & \underline{20.44} & \underline{61.39} & 20.32 & 60.81 & 78.93 & 12.92 & 43.25 & \underline{63.89} & 83.33 & \underline{96.30} & 41.11 & 71.16 & 63.03 & 78.57 & 34.87 & 58.82 & 46.52 & 72.20 & \underline{85.48} & 27.34 & 57.89 \\
VBVR-Wan2.2 & 49.58 & 76.13 & 88.77 & 16.88 & 57.84 & 21.24 & \underline{68.93} & \underline{83.38} & \underline{18.79} & \underline{48.09} & 44.44 & 75.00 & 88.89 & 33.70 & 60.51 & \underline{72.67} & \underline{78.81} & \underline{44.77} & \underline{65.42} & 38.42 & \underline{73.18} & 84.96 & 28.54 & 56.28 \\
\midrule
\rowcolor{line-blue}
\textbf{Ours} & & & & & & & & & & & & & & & & & & & & & & & & \\
\textbf{VBVR-Pro-Wan2.2-I2V-A14B} & 54.76 & \underline{82.44} & \underline{88.91} & 17.47 & 60.90 & \textbf{51.12} & \textbf{84.39} & \textbf{89.08} & \textbf{46.89} & \textbf{67.87} & \underline{63.89} & \textbf{91.67} & \textbf{98.61} & \underline{48.70} & \underline{75.72} & \textbf{73.08} & \textbf{80.98} & 42.74 & \textbf{65.60} & \underline{56.59} & \textbf{82.89} & \textbf{89.40} & \underline{38.95} & \textbf{66.96} \\
\bottomrule
\end{tabular}
}
\label{tab:appendix_ruler_bench}
\end{table*}

\begin{table*}[t]
\centering
\scriptsize
\setlength{\tabcolsep}{3pt}
\caption{Results on the vision-centric split of \textbf{VideoThinkBench}.}
\resizebox{1.0\linewidth}{!}{
\begin{tabular}{l|ccccccccccc}
\toprule
\multirow{2}{*}{\textbf{Models}} & \multicolumn{3}{c}{\textbf{Eyeballing}} & \multicolumn{3}{c}{\textbf{Visual Puzzles}} & \multirow{2}{*}{\textbf{ARC-AGI-2}} & \multicolumn{3}{c}{\textbf{Maze}} & \multirow{2}{*}{\textbf{Average}} \\
\cmidrule(lr){2-4} \cmidrule(lr){5-7} \cmidrule(lr){9-11}
 & \textbf{Point} & \textbf{Line} & \textbf{Shape} & \textbf{Symmetry} & \textbf{Gradient} & \textbf{Compos.} & & \textbf{Square} & \textbf{Hexagon} & \textbf{Circle} &  \\
\midrule
\rowcolor{gray!20}
\multicolumn{12}{l}{\textbf{Video Generation Models}} \\
\midrule
\rowcolor{line-blue}
\textbf{Proprietary Models} & & & & & & & & & & & \\
Veo-3.1 & \underline{34.44} & 24.29 & \underline{30.00} & \underline{77.50} & 40.00 & \underline{70.00} & \underline{0.71} & 0.00 & 0.00 & 0.00 & 27.69 \\
\midrule
\rowcolor{line-blue}
\textbf{Open-source Models} & & & & & & & & & & & \\
CogVideoX1.5-5B-I2V & 23.33 & 15.00 & 15.00 & 67.50 & 75.00 & 32.50 & 0.00 & 0.00 & 0.00 & 0.00 & 22.83 \\
LTX-2.3-I2AV & 28.17 & 23.33 & 17.50 & 35.00 & 55.00 & \textbf{90.00} & \textbf{55.00} & 0.00 & \underline{5.88} & 0.00 & 22.40 \\
Wan2.2-I2V-A14B & 23.33 & 23.75 & \textbf{37.50} & 55.00 & \underline{85.00} & 32.50 & 0.00 & 0.00 & 0.00 & 0.00 & 25.71 \\
VBVR-Wan2.2 & 32.22 & \underline{35.00} & 27.50 & 75.00 & 80.00 & 57.50 & 0.00 & 0.00 & 0.00 & 0.00 & \underline{30.72} \\
\midrule
\rowcolor{line-blue}
\textbf{Ours} & & & & & & & & & & & \\
\textbf{VBVR-Pro-Wan2.2-I2V-A14B} & \textbf{45.56} & \textbf{36.62} & 20.00 & \textbf{87.50} & \textbf{95.00} & 67.50 & 0.00 & \textbf{76.47} & \textbf{100.00} & 0.00 & \textbf{52.86} \\
\bottomrule
\end{tabular}
}
\label{tab:appendix_videothinkbench}
\end{table*}

\begin{table*}[t]
\centering
\scriptsize
\setlength{\tabcolsep}{3pt}
\caption{Results on \textbf{BabyVision-Gen}, reported over the four task types defined by the benchmark (accuracy, \%).}
\resizebox{0.7\linewidth}{!}{
\begin{tabular}{l|ccccc}
\toprule
\textbf{Models} & \textbf{\shortstack{Fine-grained\\Discrimination}} & \textbf{\shortstack{Visual\\Tracking}} & \textbf{\shortstack{Spatial\\Perception}} & \textbf{\shortstack{Visual Pattern\\Recognition}} & \textbf{Overall} \\
\midrule
\rowcolor{gray!20}
\multicolumn{6}{l}{\textbf{Video Generation Models}} \\
\midrule
\rowcolor{line-blue}
\textbf{Open-source Models} & & & & & \\
CogVideoX1.5-5B-I2V & 1.56 & 0.00 & 1.69 & 0.00 & 1.07 \\
LTX-2.3-I2AV & 2.34 & 0.00 & 1.69 & \underline{5.26} & 2.14 \\
Wan2.2-I2V-A14B & 0.00 & 0.00 & 0.00 & 2.63 & 0.36 \\
VBVR-Wan2.2 & \underline{4.69} & 0.00 & \underline{5.08} & \textbf{10.53} & \underline{4.64} \\
\midrule
\rowcolor{line-blue}
\textbf{Ours} & & & & & \\
\textbf{VBVR-Pro-Wan2.2-I2V-A14B} & \textbf{15.63} & 0.00 & \textbf{18.64} & \textbf{10.53} & \textbf{12.50} \\
\bottomrule
\end{tabular}
}
\label{tab:appendix_babyvision_gen}
\end{table*}

\begin{table*}[t]
\centering
\scriptsize
\setlength{\tabcolsep}{3pt}
\caption{Per-task results on \textbf{V-ReasonBench} (pass@$k$, $k=5$). Tasks are grouped by the four reasoning dimensions defined by the benchmark.}
\resizebox{1.0\linewidth}{!}{
\begin{tabular}{l|cccccccccccccc}
\toprule
\multirow{2}{*}{\textbf{Models}} & \multicolumn{4}{c}{\textbf{Structured Problem-Solving}} & \multicolumn{3}{c}{\textbf{Spatial Cognition}} & \multicolumn{3}{c}{\textbf{Pattern-based Inference}} & \multicolumn{3}{c}{\textbf{Physical Dynamics}} & \multirow{2}{*}{\textbf{Avg.}} \\
\cmidrule(lr){2-5} \cmidrule(lr){6-8} \cmidrule(lr){9-11} \cmidrule(lr){12-14}
 & \textbf{Arith.} & \textbf{Code} & \textbf{Sudoku} & \textbf{TicTacToe} & \textbf{ShapeFit} & \textbf{Symmetry} & \textbf{ColorConn.} & \textbf{SeqComp.} & \textbf{Analogy} & \textbf{RuleFollow} & \textbf{BlockSlide} & \textbf{Vessels} & \textbf{Temp.} &  \\
\midrule
\rowcolor{gray!20}
\multicolumn{15}{l}{\textbf{Video Generation Models}} \\
\midrule
\rowcolor{line-blue}
\textbf{Open-source Models} & & & & & & & & & & & & & & \\
CogVideoX1.5-5B-I2V & 0.00 & \textbf{2.27} & 0.00 & 0.00 & 0.00 & 0.00 & 0.00 & 0.00 & 0.00 & 0.00 & 0.00 & 10.00 & \textbf{90.00} & 7.87 \\
LTX-2.3-I2AV & 0.00 & 0.00 & 0.00 & \underline{10.00} & 3.57 & 0.00 & \textbf{10.00} & \underline{5.00} & 0.00 & 0.00 & \textbf{10.00} & 0.00 & 20.00 & 4.51 \\
Wan2.2-I2V-A14B & 0.00 & 0.00 & \underline{2.00} & 0.00 & 10.71 & 0.00 & 0.00 & 0.00 & 0.00 & 0.00 & \textbf{10.00} & 30.00 & \underline{80.00} & 10.21 \\
VBVR-Wan2.2 & 0.00 & \underline{2.22} & 0.00 & 0.00 & \underline{60.71} & \underline{16.67} & 0.00 & \underline{5.00} & \underline{10.00} & 0.00 & 0.00 & \underline{90.00} & 50.00 & \underline{18.05} \\
\midrule
\rowcolor{line-blue}
\textbf{Ours} & & & & & & & & & & & & & & \\
\textbf{VBVR-Pro-Wan2.2-I2V-A14B} & 0.00 & 0.00 & \textbf{76.00} & \textbf{26.67} & \textbf{88.89} & \textbf{43.33} & \textbf{10.00} & \textbf{10.00} & \textbf{70.00} & \textbf{12.00} & 0.00 & \textbf{100.00} & 60.00 & \textbf{38.22} \\
\bottomrule
\end{tabular}
}
\label{tab:appendix_vreasonbench}
\end{table*}

\begin{table*}[th!]
\centering
\scriptsize
\setlength{\tabcolsep}{3pt}
\caption{Per-category results on \textbf{MME-CoF-Pro} (no-hint setting). RS: Reasoning Score; CS: Consistency Score; Avg: mean over the five generation-quality dimensions. Following the original paper, each model is reported under all three metrics.}
\resizebox{1.0\linewidth}{!}{
\begin{tabular}{ll|ccccccccccccccccc}
\toprule
\textbf{Models} & \textbf{Metric} & \textbf{Overall} & \textbf{Visual Detail} & \textbf{Visual Trace} & \textbf{Real-world Spatial} & \textbf{3D Geometry} & \textbf{2D Geometry} & \textbf{Physics-based} & \textbf{Rotation} & \textbf{Table/Chart} & \textbf{Object Count.} & \textbf{GUI} & \textbf{Embodied} & \textbf{Medical} & \textbf{4D Dynamics} & \textbf{Natural Sci.} & \textbf{Text-based} & \textbf{Visual Logical} \\
\midrule
\rowcolor{gray!20}
\multicolumn{19}{l}{\textbf{Video Generation Models}} \\
\midrule
\rowcolor{line-blue}
\textbf{Proprietary Models} & & & & & & & & & & & & & & & & & & \\
\multirow{3}{*}{Veo-3.1} & RS & \textbf{55.90} & \textbf{61.50} & \underline{45.50} & \textbf{54.00} & \underline{43.90} & \underline{29.20} & \textbf{55.90} & \underline{60.20} & \underline{49.40} & \textbf{51.70} & \textbf{69.30} & \textbf{81.70} & 41.70 & \textbf{66.20} & \underline{49.60} & \textbf{62.30} & \textbf{72.00} \\
 & CS & \underline{47.50} & \textbf{51.20} & \underline{47.80} & \underline{57.80} & \underline{27.20} & 28.00 & 30.50 & \underline{44.40} & 31.10 & \underline{47.20} & \textbf{68.90} & \textbf{72.90} & 22.80 & \underline{68.20} & 57.30 & \textbf{49.60} & \underline{55.50} \\
 & Avg & \underline{49.50} & \underline{56.80} & \underline{53.00} & \underline{62.10} & \underline{32.20} & 32.40 & 27.10 & \underline{45.80} & 33.60 & 46.10 & \textbf{69.70} & \textbf{76.10} & 27.10 & \underline{65.40} & 56.70 & \textbf{52.50} & \underline{55.50} \\
\midrule
\rowcolor{line-blue}
\textbf{Open-source Models} & & & & & & & & & & & & & & & & & & \\
\multirow{3}{*}{CogVideoX1.5-5B-I2V} & RS & 28.46 & 11.11 & 26.77 & 30.26 & 18.70 & 13.33 & 29.24 & 35.95 & 46.37 & 11.83 & 8.52 & 43.68 & 37.57 & 43.07 & 46.94 & 20.94 & 19.75 \\
 & CS & 30.07 & 38.33 & 15.29 & 35.26 & 20.00 & 26.00 & 25.79 & 18.75 & 26.47 & 24.44 & 24.44 & 40.34 & 30.00 & 40.56 & 46.11 & 33.91 & 24.50 \\
 & Avg & 30.13 & 39.00 & 13.76 & 30.42 & 18.33 & 27.40 & 23.16 & 20.25 & 25.76 & 29.00 & 23.33 & 37.52 & 32.44 & 38.11 & 46.78 & 39.57 & 27.30 \\
\cmidrule(lr){1-19}
\multirow{3}{*}{LTX-2.3-I2AV} & RS & 34.05 & \underline{42.59} & 38.39 & 39.39 & 25.46 & 3.33 & 46.08 & 27.71 & 42.75 & 14.42 & 27.84 & 38.33 & \underline{43.90} & 56.20 & 42.31 & 19.86 & 25.00 \\
 & CS & 42.20 & 21.11 & 40.00 & 40.00 & 26.67 & \underline{31.00} & 36.32 & 21.25 & \textbf{63.53} & 42.22 & \underline{64.12} & \underline{58.97} & \underline{45.00} & 61.11 & \underline{58.89} & 34.35 & 21.90 \\
 & Avg & 44.26 & 22.33 & 41.56 & 43.47 & 29.33 & \underline{46.60} & 35.89 & 21.00 & \textbf{65.88} & 43.56 & 61.53 & 60.41 & \underline{43.75} & 60.89 & \underline{61.67} & 40.09 & 25.62 \\
\cmidrule(lr){1-19}
\multirow{3}{*}{Wan2.2-I2V-A14B} & RS & 24.76 & 15.93 & 38.07 & 22.11 & 13.24 & 13.33 & 30.99 & 23.20 & 14.91 & 13.03 & 7.96 & 29.48 & 38.23 & 34.85 & 46.02 & \underline{32.83} & 13.56 \\
 & CS & 26.91 & 29.44 & 27.78 & 40.53 & 17.22 & 9.00 & 27.00 & 12.50 & 12.22 & 26.11 & 23.89 & 32.07 & 28.33 & 47.78 & 43.89 & 29.57 & 12.27 \\
 & Avg & 29.84 & 32.89 & 31.67 & 41.47 & 18.56 & 15.20 & 25.20 & 16.88 & 27.22 & 31.56 & 22.22 & 36.00 & 31.89 & 43.33 & 44.11 & 31.30 & 18.27 \\
\cmidrule(lr){1-19}
\multirow{3}{*}{VBVR-Wan2.2} & RS & 34.13 & 29.26 & 42.08 & 33.07 & 22.13 & 15.83 & 39.90 & 39.29 & 32.04 & \underline{31.03} & 15.97 & 25.06 & 43.70 & 53.97 & 41.85 & 25.51 & 52.14 \\
 & CS & 36.57 & 41.67 & 36.67 & 44.74 & 23.33 & 22.00 & \underline{38.50} & 19.38 & 26.67 & 38.33 & 32.78 & 35.52 & 36.11 & 57.78 & 53.89 & 31.74 & 38.57 \\
 & Avg & 39.48 & 45.22 & 41.67 & 48.11 & 24.78 & 35.60 & \underline{36.10} & 21.13 & 42.44 & \underline{48.67} & 38.00 & 37.38 & 35.00 & 56.33 & 47.56 & 35.91 & 36.67 \\
\midrule
\rowcolor{line-blue}
\textbf{Ours} & & & & & & & & & & & & & & & & & & \\
\multirow{3}{*}{\textbf{VBVR-Pro-Wan2.2-I2V-A14B}} & RS & \underline{48.39} & 36.57 & \textbf{67.67} & \underline{40.96} & \textbf{57.22} & \textbf{35.83} & \underline{47.02} & \textbf{66.98} & \textbf{57.50} & 27.20 & \underline{30.93} & \underline{45.40} & \textbf{50.58} & \underline{60.13} & \textbf{52.50} & 31.16 & \underline{68.33} \\
 & CS & \textbf{61.66} & \underline{48.24} & \textbf{78.33} & \textbf{75.26} & \textbf{51.11} & \textbf{57.00} & \textbf{57.89} & \textbf{58.67} & \underline{60.56} & \textbf{58.33} & 62.22 & 54.48 & \textbf{70.56} & \textbf{70.00} & \textbf{74.44} & \underline{48.70} & \textbf{64.50} \\
 & Avg & \textbf{62.20} & \textbf{57.88} & \textbf{77.56} & \textbf{70.32} & \textbf{57.67} & \textbf{55.60} & \textbf{53.58} & \textbf{55.47} & \underline{63.00} & \textbf{60.89} & \underline{63.33} & \underline{61.66} & \textbf{65.67} & \textbf{67.00} & \textbf{69.89} & \underline{50.87} & \textbf{63.90} \\
\bottomrule
\end{tabular}
}
\label{tab:appendix_mme_cof_pro}
\end{table*}

\begin{table*}[th!]
\centering
\scriptsize
\setlength{\tabcolsep}{3pt}
\caption{Results on \textbf{IntelligentVBench} (Implicit I2V). IF: Instruction Following; CP: Condition Preserving; VQ: Visual Quality. AVG $=$ (IF$+$CP$+$VQ)$/3$, MIN $=\min($IF, CP, VQ$)$. All scores are on a 1--5 scale.}
\begin{tabular}{l|ccccc}
\toprule
\textbf{Models} & \textbf{IF $\uparrow$} & \textbf{CP $\uparrow$} & \textbf{VQ $\uparrow$} & \textbf{MIN} & \textbf{AVG} \\
\midrule
\rowcolor{gray!20}
\multicolumn{6}{l}{\textbf{Video Generation Models}} \\
\midrule
\rowcolor{line-blue}
\textbf{Proprietary Models} & & & & & \\
Veo-3.1 & 4.12 & 3.58 & 3.52 & 3.15 & 3.74 \\
\midrule
\rowcolor{line-blue}
\textbf{Open-source Models} & & & & & \\
CogVideoX1.5-5B-I2V & 3.70 & 3.74 & 3.07 & 2.82 & 3.50 \\
LTX-2.3-I2AV & 4.15 & 3.92 & 3.58 & 3.13 & 3.88 \\
Wan2.2-I2V-A14B & 4.30 & 3.68 & 3.80 & 3.30 & 3.93 \\
VBVR-Wan2.2 & \underline{4.36} & \underline{3.97} & \underline{3.91} & \underline{3.52} & \underline{4.08} \\
\midrule
\rowcolor{line-blue}
\textbf{Ours} & & & & & \\
\textbf{VBVR-Pro-Wan2.2-I2V-A14B} & \textbf{4.38} & \textbf{4.02} & \textbf{3.94} & \textbf{3.55} & \textbf{4.11} \\
\bottomrule
\end{tabular}
\label{tab:appendix_intelligentvbench}
\end{table*}

\newpage

\section{Reinforcement Learning Details}
\label{sec:rl_details}

\subsection{Training Pipeline}

\begin{figure}[t]
\centering
\includegraphics[width=\columnwidth]{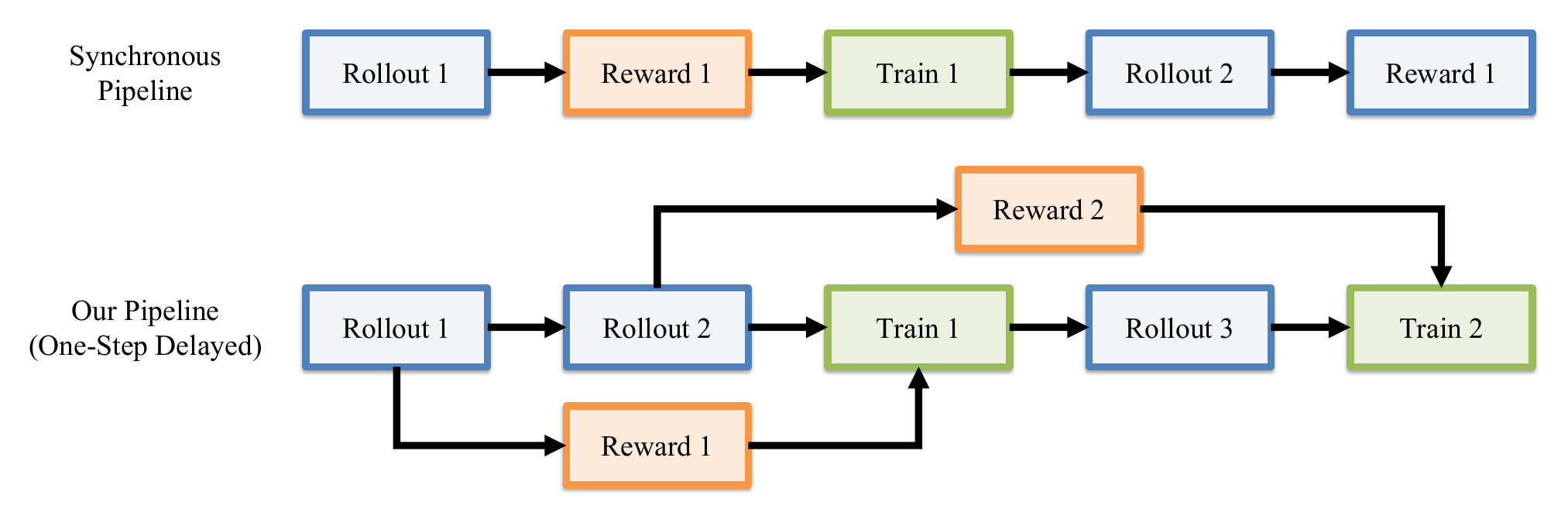}
\caption{
Comparison between synchronous training and our one-step-delayed pipeline.
Our pipeline overlaps asynchronous preprocessing and reward evaluation with rollout generation and policy optimization, reducing GPU idle time while bounding policy lag to one rollout iteration.
}
\label{fig:pipelined_training}
\end{figure}

To improve training efficiency, we adopt a one-step-delayed rollout pipeline that overlaps reward computation with GPU-side rollout generation and policy optimization. As illustrated in \cref{fig:pipelined_training}, the synchronous pipeline requires policy optimization to wait for rollout preprocessing and reward evaluation to complete, resulting in substantial GPU idle time.

Inspired by DeepCoder~\cite{luo2025deepcoder} and AReaL~\cite{fu2026areal}, our pipeline introduces a fixed one-step delay between rollout generation and optimization. After rollout $k$ is generated, its preprocessing and reward computation are asynchronously dispatched to reward workers, while the GPUs continue generating rollout $k+1$. Once the rewards for rollout $k$ become available, the GPUs perform policy optimization using rollout $k$ while the reward workers process rollout $k+1$. This design removes reward computation from the critical execution path and improves GPU utilization.

Unlike fully asynchronous reinforcement learning systems, our approach preserves a deterministic execution order and bounds policy staleness to at most one rollout iteration. Therefore, it achieves efficient overlap between reward computation and policy execution while avoiding excessive policy lag.

To further reduce rollout overhead, we generate training samples at $512 \times 512$ resolution and only upscale them to $1024 \times 1024$ during reward computation. This strategy reduces the cost of trajectory generation while maintaining high-resolution inputs for reward evaluation.

How effectively this overlap can be exploited also depends on the systems cost of reward evaluation. The verifiable scorer used by RLVR runs on CPU reward workers and can therefore overlap with GPU-side rollout generation and policy optimization without occupying accelerator memory. In contrast, RLVLM requires serving Qwen-3.6-27B as a GPU-resident reward model; in our implementation, we reserve 50\% of the available GPU memory for judge inference. This difference is reflected in end-to-end training time: on 128 H800 GPUs, training for 2,200 steps takes approximately 5.1 days with the verifiable scorer, compared with 8.3 days with the VLM judge. RLVR therefore provides a $1.63\times$ increase in training throughput and reduces wall-clock time by 38.7\%.

\subsection{Training Optimizations}

We investigate KL regularization with respect to a reference policy, which is commonly adopted in group-relative policy optimization. Maintaining an explicit reference model requires additional forward computation and memory consumption. Although lightweight alternatives such as Shadow KL avoid explicit reference-model evaluation, we find that KL regularization does not provide consistent gains in our setting. Therefore, all experiments are conducted without KL regularization.

For efficient large-scale training, we replace the native RMSNorm implementation with the fused Liger Kernel~\cite{hsu2024liger}, employ FlashAttention-3~\cite{shah2024flashattention} for memory-efficient attention computation, and enable \texttt{torch.compile} whenever compatible. We additionally use hybrid-sharded FSDP2~\cite{zhao2023pytorchfsdpexperiencesscaling}, which shards model states within each node while maintaining data-parallel synchronization across nodes.

\subsection{Training and Inference Configuration}
\label{sec:rl_inference}

During RL training, we use 30 sampling steps with the CPS sampler. For policy optimization, 60\% of the denoising transitions are replayed, following the formulation described in \cref{sec:rl:algo}. Each optimization batch contains 32 prompts, with 32 trajectories sampled per prompt. We set the CPS stochasticity parameter to $\eta=0.7$ and use a learning rate of $5 \times 10^{-6}$. Unless otherwise specified, all RL experiments use a CFG scale of 1.0.

For the controlled comparison in the main paper, both Verifiable-RL and VLM-Judge RL are evaluated using CPS with $\eta=0.7$. This matched configuration isolates the effect of the reward objective from that of the inference sampler. All RL evaluations use 30 denoising steps, matching the rollout configuration used during training.

We additionally evaluate deterministic ODE sampling with Euler and UniPC in the sampler-sensitivity study. The CPS reference implementation~\cite{wang2025coefficients} uses an Euler discretization as its deterministic ODE backbone, whereas the official Wan inference pipeline adopts UniPC by default. Euler is a first-order single-step solver, while UniPC~\cite{NEURIPS2023_9c2aa1e4} uses a higher-order predictor--corrector formulation. We refer to these configurations as ODE (Euler) and ODE (UniPC), respectively. Both use the same 30-step schedule and CFG scale as the RL evaluation, ensuring that the solver is the only inference variable being changed.

For the supervised fine-tuning (SFT) baseline, we retain the default inference configuration of the official Wan implementation: UniPC with 50 sampling steps and a CFG scale of 5. This configuration is kept unchanged from the original model inference pipeline.

\subsection{Sampler Sensitivity}
\label{sec:rl_sampler_sensitivity}

\begin{table*}[t]
\centering
\scriptsize
\setlength{\tabcolsep}{3pt}
\caption{
Full comparison of inference samplers on \textbf{\benchname}.
We evaluate CPS at different stochasticity levels and deterministic ODE sampling using either the Euler or UniPC solver.
Baseline and SFT use the default UniPC inference configuration.
Detailed inference settings are provided in \cref{sec:rl_inference}.
We report the overall, In-Domain (ID), and Out-of-Domain (OOD) scores together with category-wise performance.
Higher is better.
\textbf{Bold}: best; \underline{underline}: second best.
Rankings are computed before rounding.
}
\label{tab:vbvr_rl_results_full}

\resizebox{\linewidth}{!}{
\begin{tabular}{ll|c|c|ccccc|c|ccccc}
\toprule
\multicolumn{2}{c|}{}
& \multicolumn{1}{c|}{}
& \multicolumn{6}{c|}{\textbf{In-Domain}}
& \multicolumn{6}{c}{\textbf{Out-of-Domain}} \\
\cmidrule(lr){4-4}
\cmidrule(lr){5-9}
\cmidrule(lr){10-10}
\cmidrule(lr){11-15}
\textbf{Model}
& \textbf{Inference}
& \textbf{Overall}
& \textbf{Avg.}
& \textbf{Abst.}
& \textbf{Know.}
& \textbf{Perc.}
& \textbf{Spat.}
& \textbf{Trans.}
& \textbf{Avg.}
& \textbf{Abst.}
& \textbf{Know.}
& \textbf{Perc.}
& \textbf{Spat.}
& \textbf{Trans.} \\
\midrule


Baseline
& UniPC
& 0.470
& 0.641
& 0.578
& 0.511
& 0.476
& 0.480
& 0.724
& 0.300
& \textbf{0.585}
& 0.153
& 0.129
& 0.311
& 0.352 \\

SFT
& UniPC
& 0.503
& 0.679
& 0.671
& 0.637
& 0.524
& 0.687
& 0.754
& 0.328
& 0.452
& 0.185
& 0.214
& \underline{0.555}
& \textbf{0.450} \\

\midrule

\multirow{6}{*}{RLVR}
& CPS ($\eta=0.1$)
& 0.509
& 0.675
& 0.699
& 0.649
& 0.586
& 0.644
& 0.776
& 0.342
& 0.489
& 0.277
& 0.246
& 0.499
& 0.390 \\

& CPS ($\eta=0.3$)
& 0.526
& 0.698
& \underline{0.723}
& 0.676
& 0.618
& 0.680
& 0.783
& 0.354
& 0.502
& 0.289
& 0.254
& 0.471
& \underline{0.439} \\

& CPS ($\eta=0.7$)
& \textbf{0.548}
& \textbf{0.719}
& \textbf{0.743}
& \textbf{0.689}
& \textbf{0.657}
& \textbf{0.709}
& \textbf{0.815}
& \textbf{0.377}
& \underline{0.511}
& 0.282
& \textbf{0.288}
& 0.551
& 0.388 \\

& CPS ($\eta=0.9$)
& \underline{0.539}
& \underline{0.706}
& 0.709
& \underline{0.679}
& \underline{0.650}
& \underline{0.697}
& 0.799
& \underline{0.372}
& 0.501
& 0.273
& \underline{0.282}
& \textbf{0.556}
& 0.382 \\

& ODE (Euler)
& 0.522
& 0.697
& 0.718
& 0.666
& 0.596
& 0.688
& 0.806
& 0.347
& 0.499
& 0.266
& 0.263
& 0.495
& 0.387 \\

& ODE (UniPC)
& 0.522
& 0.696
& 0.723
& 0.664
& 0.596
& 0.687
& \underline{0.815}
& 0.347
& 0.493
& 0.269
& 0.268
& 0.493
& 0.383 \\

\midrule

\multirow{6}{*}{RLVLM}
& CPS ($\eta=0.1$)
& 0.482
& 0.621
& 0.630
& 0.602
& 0.512
& 0.572
& 0.768
& 0.342
& 0.463
& \textbf{0.302}
& 0.259
& 0.455
& 0.371 \\

& CPS ($\eta=0.3$)
& 0.493
& 0.650
& 0.656
& 0.644
& 0.539
& 0.643
& 0.750
& 0.335
& 0.462
& 0.283
& 0.235
& 0.485
& 0.393 \\

& CPS ($\eta=0.7$)
& 0.508
& 0.671
& 0.679
& 0.667
& 0.601
& 0.652
& 0.771
& 0.345
& 0.468
& 0.260
& 0.256
& 0.467
& 0.389 \\

& CPS ($\eta=0.9$)
& 0.509
& 0.665
& 0.668
& 0.654
& 0.573
& 0.673
& 0.773
& 0.352
& 0.469
& 0.269
& 0.266
& 0.481
& 0.385 \\

& ODE (Euler)
& 0.488
& 0.643
& 0.666
& 0.618
& 0.529
& 0.626
& 0.735
& 0.333
& 0.476
& 0.286
& 0.245
& 0.474
& 0.384 \\

& ODE (UniPC)
& 0.497
& 0.653
& 0.694
& 0.609
& 0.553
& 0.634
& 0.754
& 0.342
& 0.482
& \underline{0.295}
& 0.256
& 0.475
& 0.386 \\

\bottomrule
\end{tabular}
}
\end{table*}

The main results use CPS with $\eta=0.7$ for both RL models to provide a matched comparison. \cref{tab:vbvr_rl_results_full} extends this evaluation to CPS with $\eta\in\{0.1,0.3,0.7,0.9\}$ and deterministic ODE sampling with either Euler or UniPC.

For Verifiable-RL, increasing the CPS stochasticity from $\eta=0.1$ to $\eta=0.7$ improves the overall score from 0.509 to 0.548, with corresponding improvements on both the In-Domain and Out-of-Domain benchmarks. Performance decreases slightly at $\eta=0.9$, suggesting that moderate-to-high stochasticity provides useful exploration, whereas further noise offers no additional benefit. CPS with $\eta=0.7$ therefore provides the strongest aggregate result for Verifiable-RL.

The two deterministic solvers produce nearly identical results for Verifiable-RL: ODE (Euler) and ODE (UniPC) both achieve an overall score of 0.522, with only minor category-level differences. Their close agreement indicates that the deterministic results are not sensitive to the particular ODE solver. Nevertheless, both remain below CPS with $\eta=0.7$, supporting the importance of stochastic exploration for tasks that require discrete visual decisions.

VLM-Judge RL exhibits smaller and less consistent differences across samplers. Its strongest aggregate results are obtained with CPS at $\eta=0.7$ or $\eta=0.9$, while ODE (UniPC) performs slightly better than ODE (Euler). However, the gaps are substantially smaller than for Verifiable-RL. Moreover, Verifiable-RL remains stronger than VLM-Judge RL under every matched sampler, showing that the advantage of the verifiable reward is not specific to the inference configuration used in the main paper.

\begin{figure}[t]
\centering
\includegraphics[width=\columnwidth]{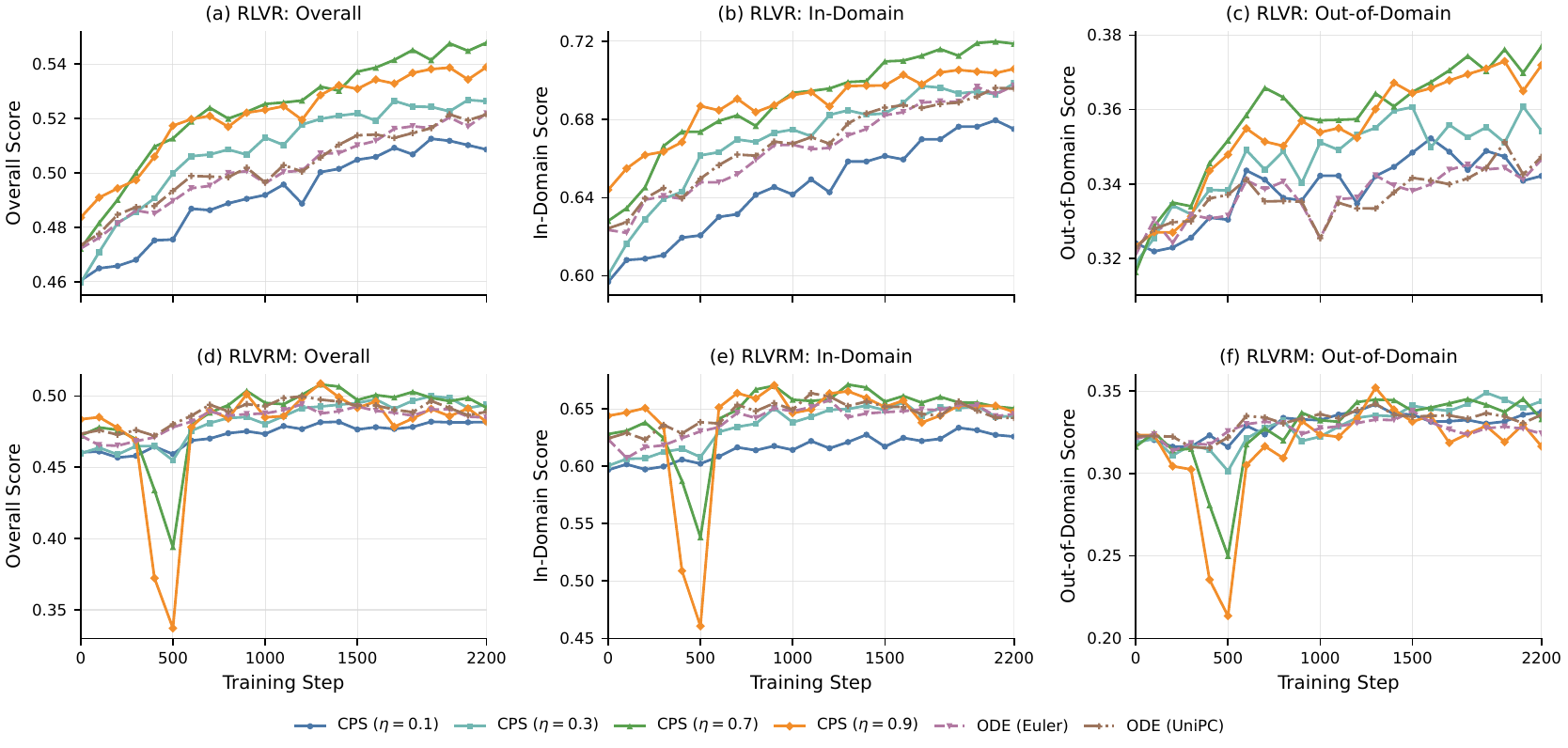}
\caption{Evaluation performance throughout training under different inference samplers. The top and bottom rows show RLVR and RLVRM, respectively, while the columns report overall, in-domain, and out-of-domain scores. Each panel compares CPS with $\eta\in\{0.1,0.3,0.7,0.9\}$, ODE (Euler), and ODE (UniPC). Checkpoints are evaluated every 100 training steps.}
\label{fig:rl_sampler_sensitivity_training}
\end{figure}

\cref{fig:rl_sampler_sensitivity_training} complements the aggregate results by showing how performance evolves during training. For Verifiable-RL, performance generally improves more rapidly as the CPS stochasticity increases from $\eta=0.1$ to $\eta=0.7$, while $\eta=0.9$ provides no further improvement. The ODE (Euler) and ODE (UniPC) trajectories closely overlap and remain below the best CPS configurations across the three metrics. For VLM-Judge RL, the differences among samplers are smaller and less consistent. Higher-stochasticity CPS reaches a higher performance ceiling but also exhibits a pronounced temporary drop around Steps 400--500, whereas the deterministic ODE trajectories are smoother. Thus, stochastic sampling provides useful exploration when paired with a reliable verifiable reward, but can amplify errors that are not adequately penalized by the VLM judge. We analyze this behavior below.

\subsection{Analysis of VLM-Judge Instability}
\label{sec:vlm_judge_instability}

As observed in \cref{fig:rl_training_eval,fig:rl_sampler_sensitivity_training}, VLM-Judge RL exhibits a temporary performance drop around Steps 400--500 under CPS with $\eta=0.7$. The same behavior becomes more pronounced with $\eta=0.9$. We investigate this instability by comparing inference samplers at Checkpoint 500 and tracing the corresponding failure mode across nearby training checkpoints.

\begin{figure}[t]
\centering
\includegraphics[width=\columnwidth]{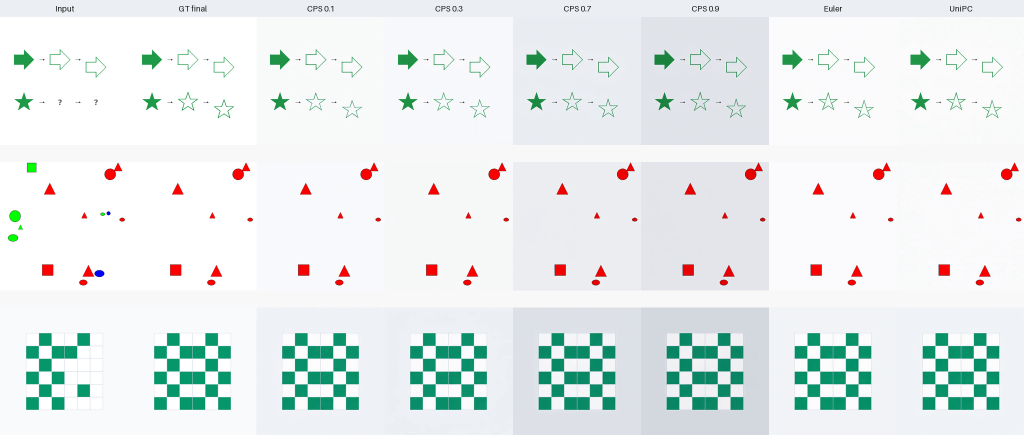}
\caption{Sampler sensitivity of VLM-Judge RL at Step 500. Across three representative tasks, the foreground transformations remain largely correct, while the output background becomes increasingly gray as the CPS stochasticity increases. Low-stochasticity CPS and deterministic ODE sampling better preserve the target background.}
\label{fig:rl_training_checkpoint500_sampler_sensitivity}
\end{figure}

\begin{figure}[t]
\centering
\includegraphics[width=\columnwidth]{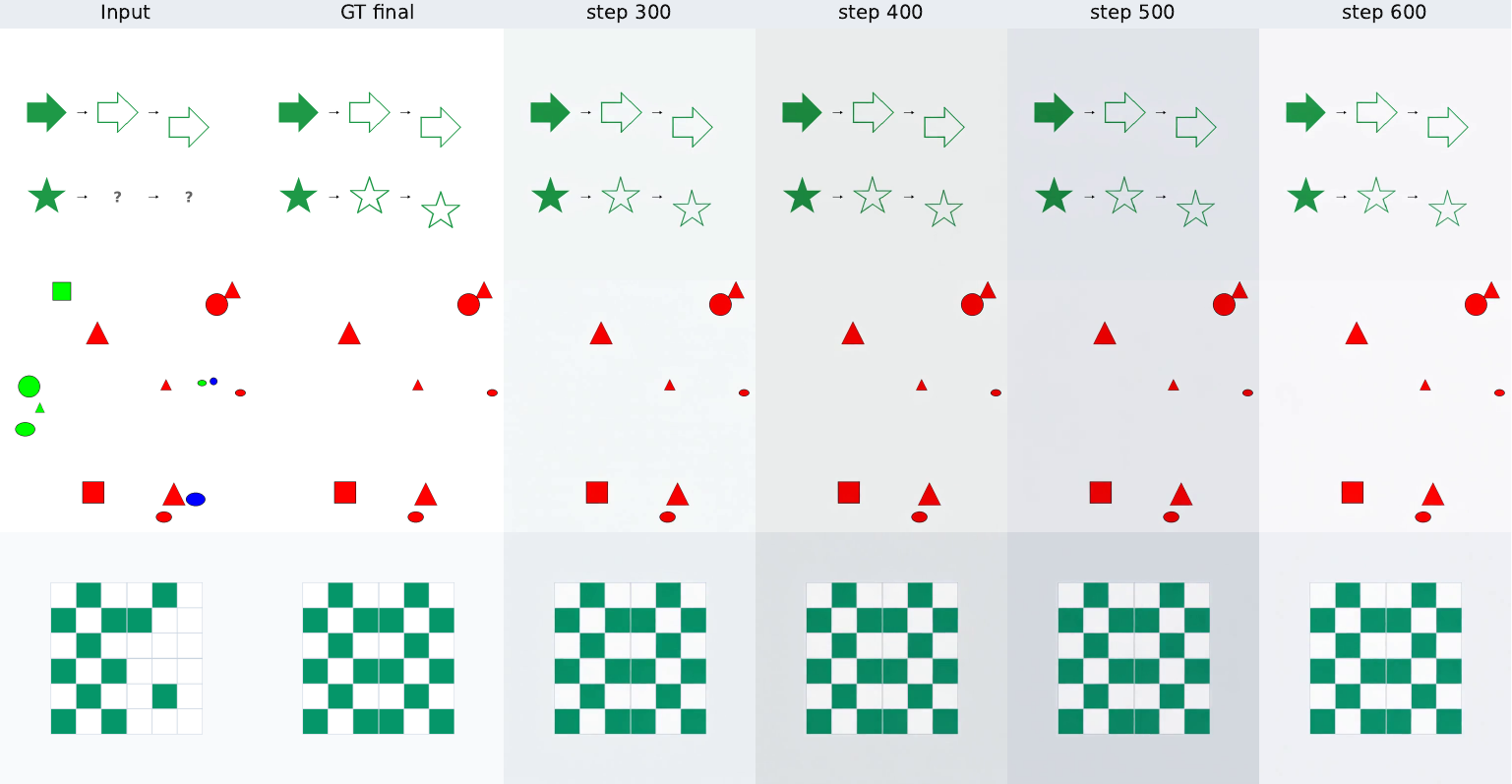}
\caption{Evolution of the background-color failure under CPS with $\eta=0.9$. The foreground task outcomes remain largely correct, while a gray background tint emerges at Step 400, becomes most pronounced at Step 500, and largely disappears by Step 600.}
\label{fig:rl_training_cps09_checkpoint_drop_examples}
\end{figure}

\cref{fig:rl_training_checkpoint500_sampler_sensitivity} compares inference samplers at the checkpoint exhibiting the largest performance drop. The foreground transformations remain largely correct across samplers, but the generated backgrounds become progressively grayer as the CPS stochasticity increases. In contrast, low-stochasticity CPS and deterministic ODE sampling largely preserve the target white background. The drop therefore reflects a sampler-dependent degradation in visual fidelity rather than a complete failure of the underlying reasoning capability.

\cref{fig:rl_training_cps09_checkpoint_drop_examples} further traces this failure mode across training checkpoints under CPS with $\eta=0.9$. The background-color deviation begins to emerge at Step 400, is most severe at Step 500, and largely recovers by Step 600, closely matching the temporary drop in evaluation performance.

These observations suggest that the general-purpose VLM judge does not reliably penalize subtle background-color deviations. Such errors can therefore persist during optimization and become amplified under more stochastic sampling, even when the foreground task is completed correctly. In contrast, the task-grounded verifiable scorer explicitly captures these deviations and provides a more stable optimization signal. This explains why increasing stochasticity benefits VLM-Judge RL less consistently than Verifiable-RL, despite using the same underlying policy optimization procedure.

\subsection{Directed Temporal Trajectory Visualization}
\label{sec:temporal_trajectory_visualization}

To summarize the complete navigation videos without displaying a sparse subset of frames, we render each 81-frame video as a stage-wise directed centerline trajectory. Let $I_t$ denote frame $t$, where $t\in\{0,\ldots,80\}$. Because the scene layout and task markers remain static, we first estimate a clean background by taking the per-pixel temporal median. We then construct a soft foreground mask $M_t(p)\in[0,1]$ using the known color of the moving agent---green in the maze task and orange in the waypoint task.

To recover a single agent location from each frame, we threshold the soft mask, retain its largest connected component $\mathcal{C}_t$, and use the component centroid:
\begin{equation}
B(p)=\operatorname*{median}_t I_t(p),
\qquad
\widehat{p}_t=
\frac{1}{\lvert\mathcal{C}_t\rvert}
\sum_{p\in\mathcal{C}_t}p.
\end{equation}
Keeping only the largest component suppresses small compression artifacts and residual colors from the static task markers. We linearly interpolate isolated single-frame detection failures when the neighboring locations are sufficiently close. Longer disappearances and large inter-frame jumps remain disconnected, so failures such as abrupt relocation in the pre-RL video are not artificially repaired.

We divide each RLVR trajectory at the frames where the agent reaches the task-specific intermediate targets and apply the same temporal boundaries to the corresponding pre-RL video for direct comparison. For \cref{fig:rl_appendix_waypoint_navigation}, we divide the trajectories into four stages: Start-to-1, 1-to-2, 2-to-3, and 3-to-Goal. For \cref{fig:rl_better_reasoning_maze}, we divide them into three stages: Start-to-Red, Red-to-Cyan, and Cyan-to-Goal. Each stage is assigned a distinct color, and arrowheads indicate the direction of motion. When two stages traverse the same spatial segment in opposite directions, their centerlines are slightly offset so that both passes remain visible.

Finally, we render the directed trajectories over the median background and restore the static task markers from a clean reference image. All 81 frames contribute to the recovered trajectories; no temporal subsampling is used. Compared with direct RGB averaging, which attenuates a moving object that occupies each location for only a few frames, the foreground-based centerline preserves its motion. Compared with a temporal foreground union, the stage-wise representation additionally exposes traversal direction, repeated visits, and discontinuities.

\subsection{Additional Qualitative Examples}
\label{sec:rl_makes_reasoning_step_better}

We provide additional examples to illustrate how Verifiable-RL changes the model's reasoning trajectory. The first three examples compare predictions at different denoising steps, while the final example examines how the generated behavior evolves across video frames. We use \emph{pre-RL} to denote the SFT-initialized policy from which Verifiable-RL is trained.

\begin{figure}[t]
\centering
\includegraphics[width=\columnwidth]{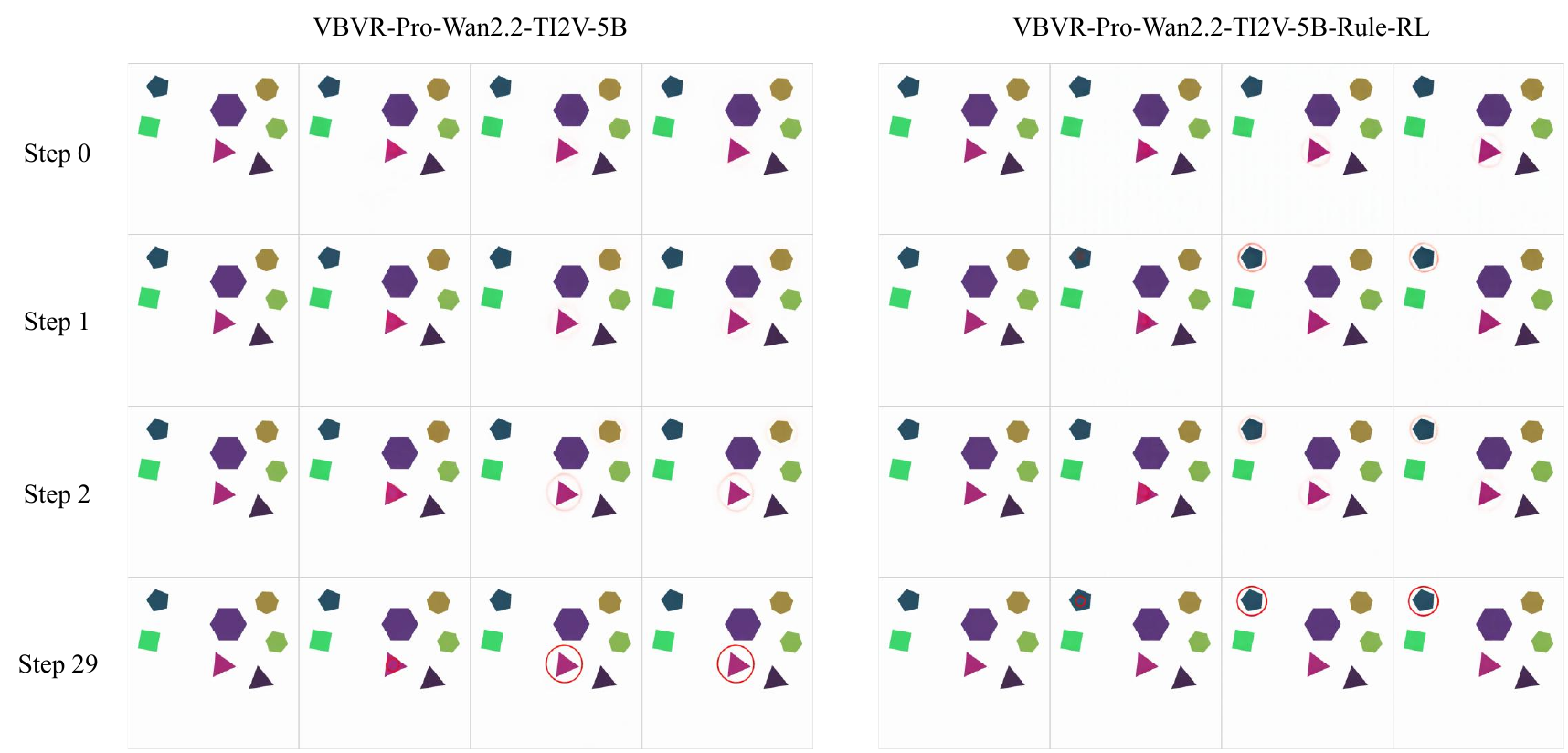}
\caption{Revision of an object-selection hypothesis. The correct target is the blue pentagon. The pre-RL model commits to an incorrect triangle and retains this choice, whereas Verifiable-RL revises its initial prediction and ultimately selects the correct object. Rows show predictions at different denoising steps.}
\label{fig:rl_appendix_hypothesis_revision}
\end{figure}

\cref{fig:rl_appendix_hypothesis_revision} shows that Verifiable-RL is more capable of revising an intermediate decision. Once the pre-RL model begins marking the triangle, its prediction remains unchanged. Verifiable-RL also considers the triangle initially, but subsequently redirects its prediction toward the correct pentagon. Since both models use the same 30-step sampling schedule, this difference reflects more effective use of the denoising trajectory rather than additional inference computation.

\begin{figure}[t]
\centering
\includegraphics[width=\columnwidth]{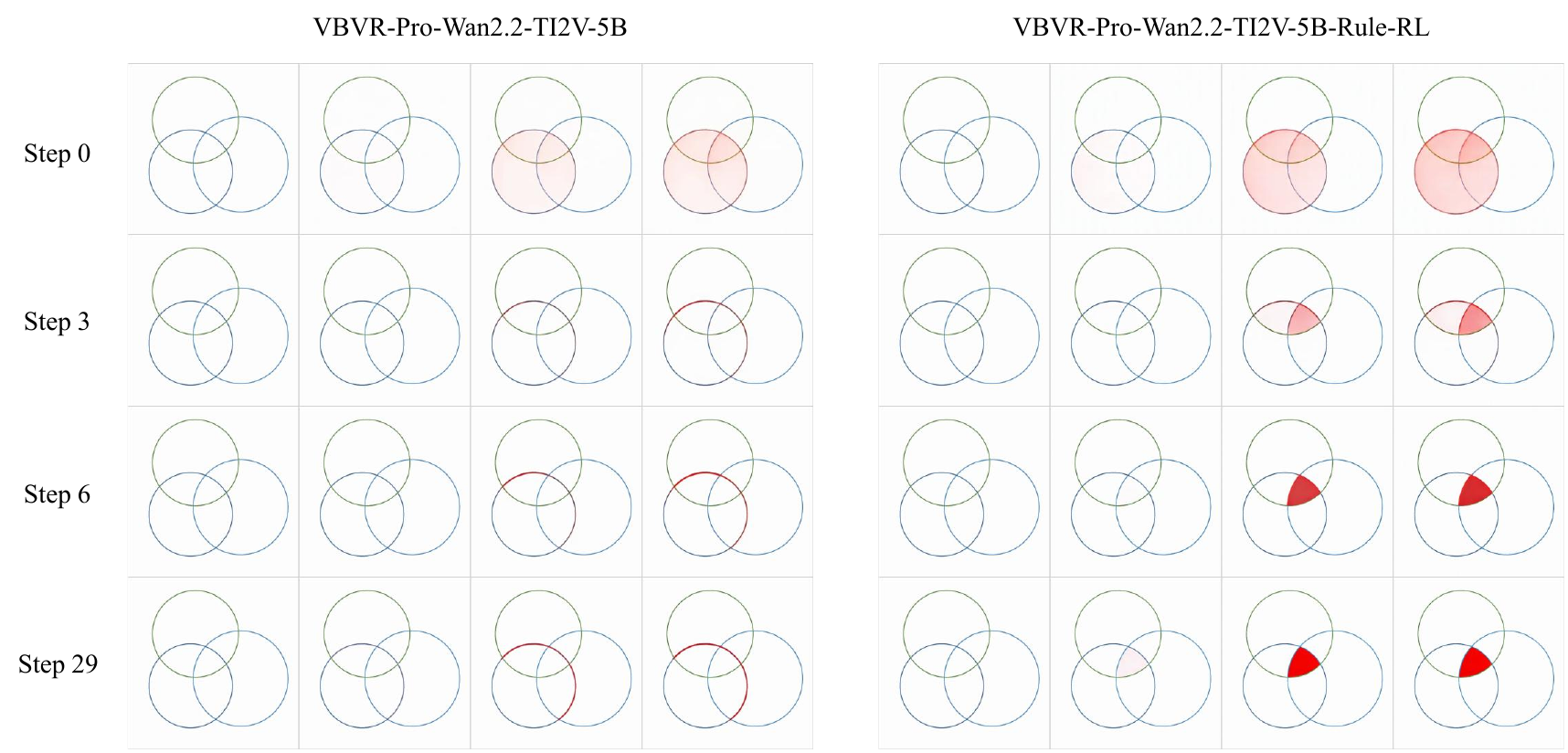}
\caption{Progressive localization in a set-intersection task. The pre-RL model fails to isolate the target region and produces incomplete boundary traces. Verifiable-RL gradually narrows a broad initial prediction to the common intersection of the three circles.}
\label{fig:rl_appendix_region_localization}
\end{figure}

\begin{figure}[t]
\centering
\includegraphics[width=\columnwidth]{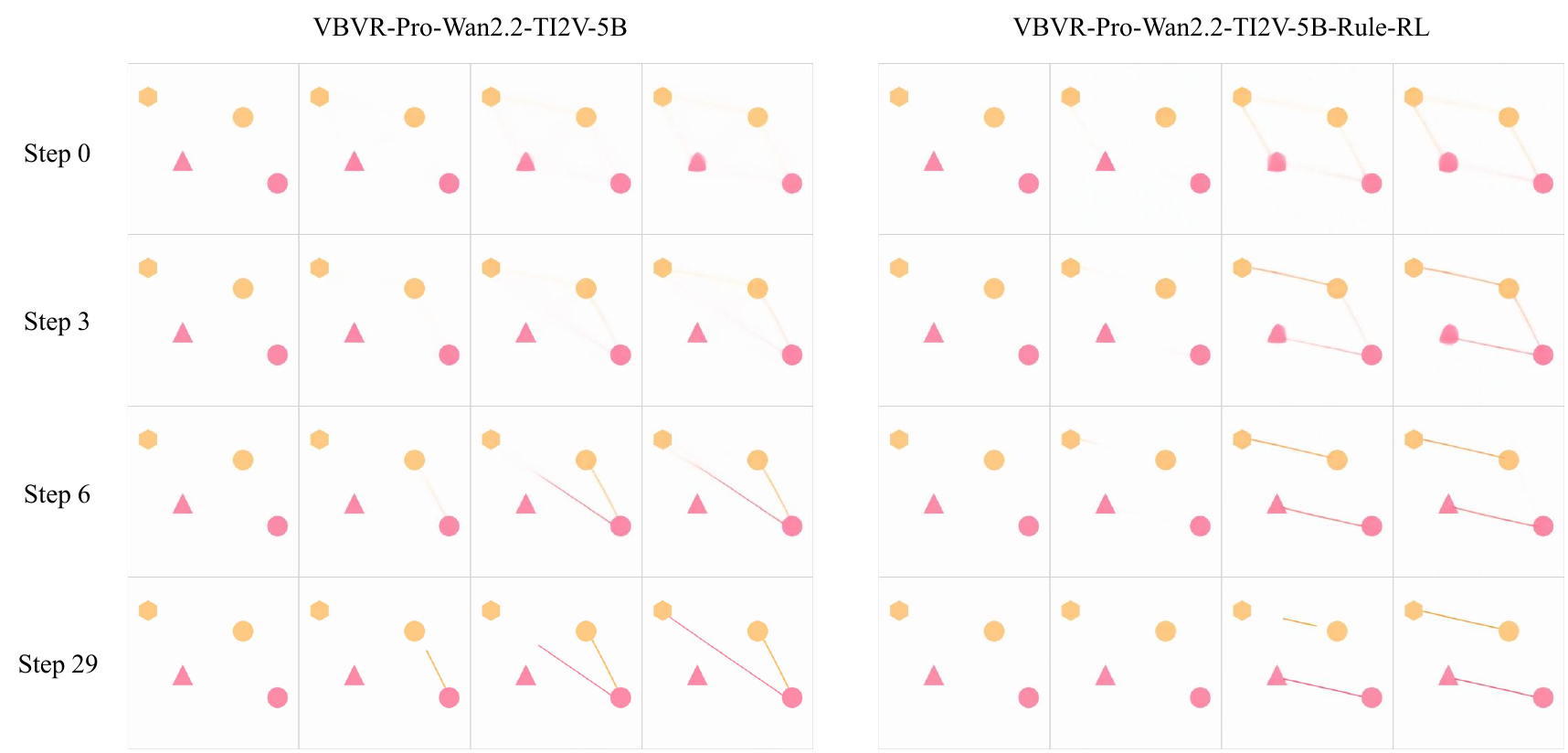}
\caption{Progressive construction of object relations. The task requires connecting the two objects of each color. The pre-RL model produces inconsistent cross-color connections, whereas Verifiable-RL gradually forms the two correct same-color pairings.}
\label{fig:rl_appendix_relation_consistency}
\end{figure}

The examples in \cref{fig:rl_appendix_region_localization,fig:rl_appendix_relation_consistency} involve different tasks but exhibit a similar progression. In the set-intersection task, the model must identify the correct region. Verifiable-RL progressively reduces a broad predi  ction to the intended overlap, while the pre-RL model remains focused on irrelevant boundaries. In the object-matching task, the model must determine which objects should be connected. Verifiable-RL gradually replaces uncertain or inconsistent connections with the two correct pairings. Thus, one example concerns finding the correct region and the other concerns connecting the correct objects, but both show a more orderly transition toward a task-consistent solution after RL.

\begin{figure}[t]
\centering
\includegraphics[width=1.0\columnwidth]{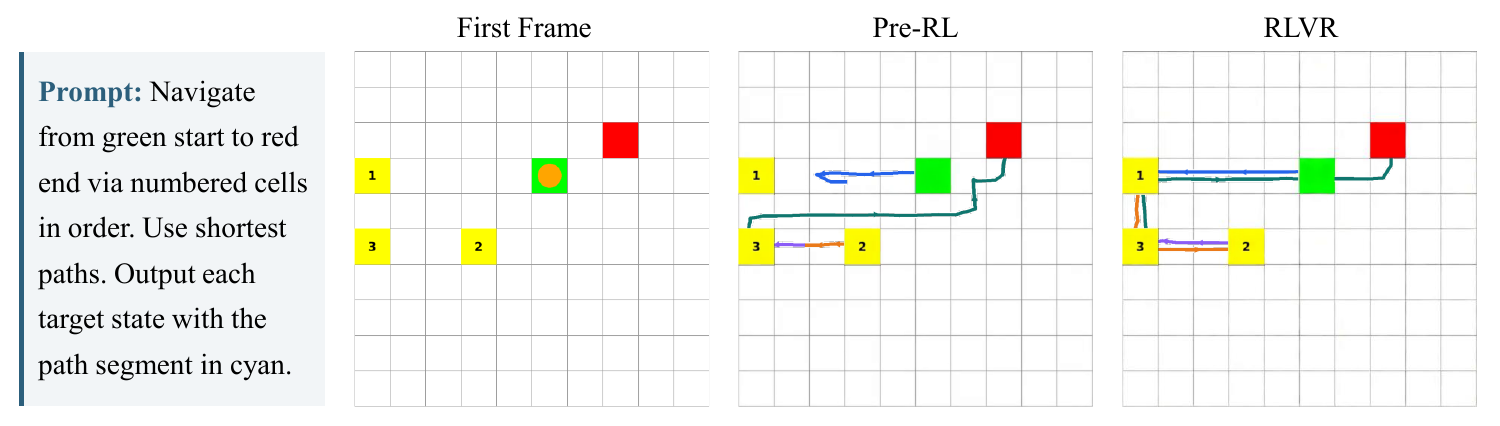}
\caption{Stage-wise visualization of a waypoint-navigation task. The orange agent must start from the green square, visit Waypoints 1, 2, and 3 in order, and finally reach the red square. Each panel summarizes all 81 video frames as a directed centerline trajectory: blue, orange, purple, and teal denote the Start-to-1, 1-to-2, 2-to-3, and 3-to-Goal stages, respectively, while arrowheads indicate motion direction. Overlapping paths are slightly offset to reveal repeated traversal in opposite directions. The pre-RL model fails to complete the first stage and subsequently exhibits discontinuous relocation, whereas Verifiable-RL completes the required waypoint sequence and reaches the destination. The trajectory-rendering procedure is described in \cref{sec:temporal_trajectory_visualization}.}
\label{fig:rl_appendix_waypoint_navigation}
\end{figure}

The preceding examples compare predictions across denoising steps, whereas \cref{fig:rl_appendix_waypoint_navigation} examines the temporal evolution of the final video. The pre-RL model places the agent at inconsistent locations and fails to maintain a coherent sequence of intermediate goals. Verifiable-RL produces more continuous state transitions and eventually reaches the destination. This suggests that RL improves not only how a solution is refined during denoising, but also how a multi-stage behavior is executed over time.

Overall, these examples show that Verifiable-RL improves two aspects of the visual reasoning trajectory: intermediate predictions become more effectively refined toward task-consistent outcomes, and the resulting behavior becomes more coherent across video frames.

\section{Additional Details and Analysis of the \name{}-Dataset}
\label{app:details_of_dataset}

\subsection{Additional Examples of the Multi-modality Design}
\label{app:multimodal_examples}

\begin{figure}[t]
  \centering
  \includegraphics[width=\columnwidth]{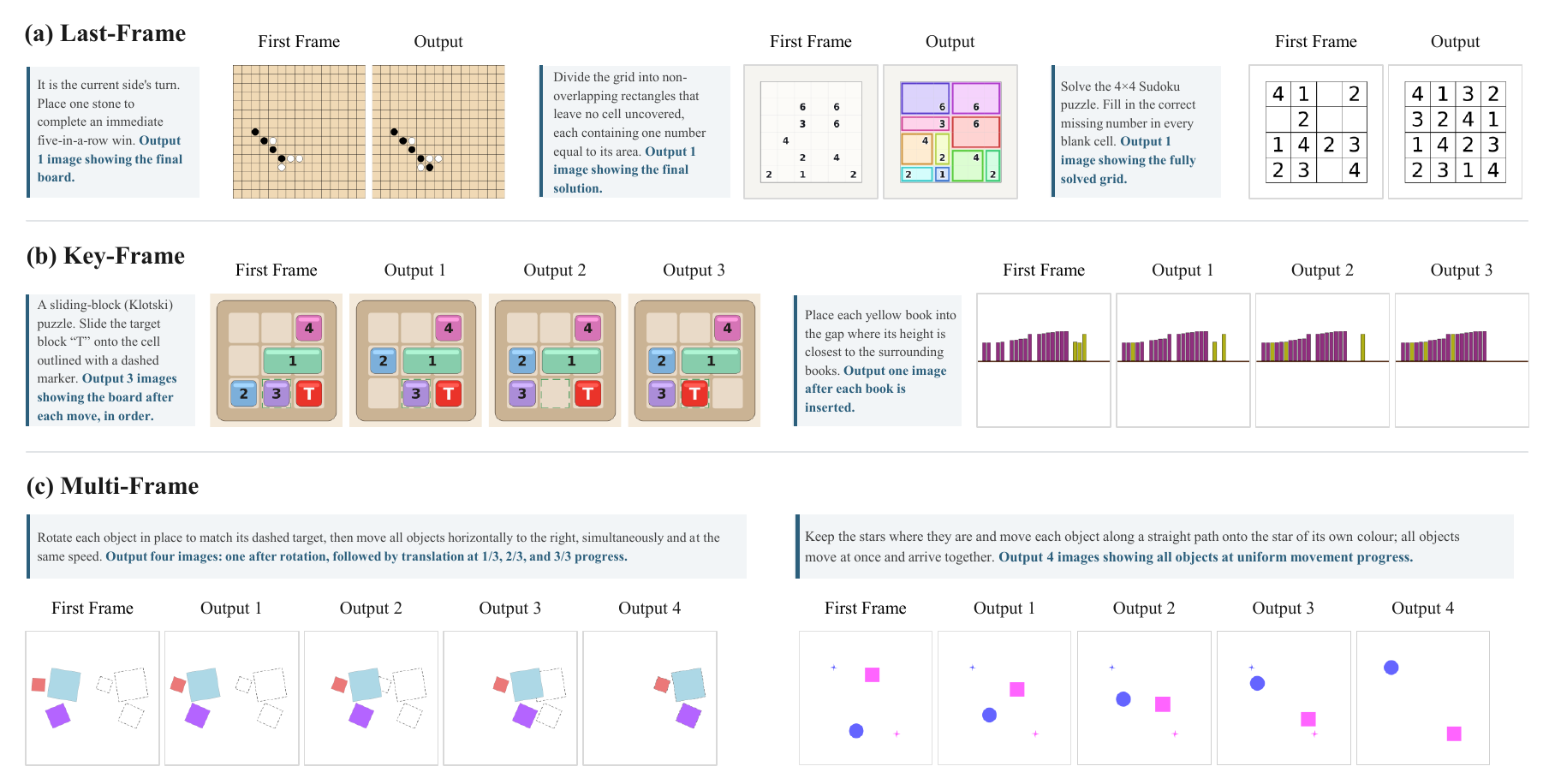}
  \caption{Additional examples illustrating the three output regimes of the \name{} image modality.}
  \label{fig:app_image_modality_examples}
\end{figure}

Every instance in \name{} is released in two modalities, video and image, which are two renderings of the same solved instance rather than two independently collected corpora. This appendix describes how the image modality is derived from a solved instance and provides additional examples of the three output regimes introduced in~\cref{sec:dataset:implementation}. \cref{fig:app_image_modality_examples} shows multiple tasks from each regime.

\paragraph{Choice of output regime.}
The regime is a property of the task generator, so all instances of a task share the same regime. It is determined by how much of the solution trajectory has to be made explicit in order to specify the answer. \emph{Last-Frame} is used when the solution is fully determined by its final state and no intermediate state carries information the model must produce; the task then has exactly one output image, as in completing a Sudoku grid or placing the winning stone in Gomoku. \emph{Key-Frame} is used when the solution consists of semantically meaningful discrete steps whose intermediate states form part of the answer, such as one board state per move in a sliding-block puzzle or one shelf state per inserted book. The number of output images is therefore generally instance-dependent, following the number of steps in that instance's solution. Path-based tasks, such as mazes, are treated as a special case: the complete solution path is overlaid on a selected frame, allowing the trajectory to be represented in a single image. \emph{Multi-Frame} is used for continuous processes that have no natural discrete decomposition, such as rotating and translating an object to a target pose. In this regime, intermediate states are sampled at uniform progress along a task-specific process parameter.

\paragraph{Image prompts.}
The video and image modalities share the same task description. The image prompt appends an explicit output specification stating how many images to produce and what each should depict, as shown in bold in~\cref{fig:app_image_modality_examples}. The video prompt omits this specification because the video represents the process continuously. This keeps the task conditions identical across modalities while making the discretisation used by the image modality explicit to the model.

\paragraph{Distribution over tasks.}
Of the 300 tasks in \name{}, 209 (69.7\%) use the Last-Frame regime, 58 (19.3\%) use Key-Frame, and 33 (11.0\%) use Multi-Frame. Last-Frame accounts for approximately seventy percent of the tasks because, for many reasoning problems, the final state alone constitutes a complete answer. The remaining thirty percent require intermediate solution states to be represented explicitly, and thus provide a controlled setting for studying a broader question: whether dynamic video or discretized image sequences offer a more effective representation for visual reasoning.

\subsection{Interleaved Data Generation}
\label{app:interleaved_generation}

The interleaved representation pairs each output image with a textual reasoning step placed immediately before it. The images are taken directly from the image modality described in \cref{app:multimodal_examples}. An instance containing $K$ output images therefore produces $K$ reasoning-image pairs.

\paragraph{Reasoning text generation.}
For each instance, Gemini-3.1-Pro-Preview receives the task question, input image, and complete sequence of solver-generated output images, and produces one reasoning step per image. The complete trajectory is provided because many tasks are difficult to solve reliably from the input alone. Gemini therefore annotates the known state transitions rather than independently discovering the solution. Each step is phrased prospectively and placed before its corresponding image.

\paragraph{Task-specific prompts.}
We manually design one prompt for each task, specifying its rules, image order, required reasoning, and output format. We then inspect five generated annotations from each task to verify that they are consistent with the solver-generated transitions. In particular, we check that each step explains the decision leading to its corresponding output image rather than merely captioning the image. As a representative example, we show the annotation prompt for the Klotski sliding-block task, shown as the leftmost example in the Key-Frame row of \cref{fig:app_image_modality_examples}.

\begin{promptbox}
\noindent\textbf{Example annotation prompt: Klotski sliding-block task}

\medskip
\noindent
You are given the complete sequence of images for a Klotski sliding-block task, including the initial tray and the tray after each successive slide. Use the complete sequence as the ground-truth solution trajectory for generating the reasoning steps.

\medskip
\noindent\textbf{Scene.}
A tray is packed with rectangular wooden blocks and a small amount of empty space. Blocks may slide horizontally or vertically into the open space, but cannot be lifted, rotated, or overlapped. The goal is to move the highlighted target block to its designated exit. Each output image shows the tray after one slide.

\medskip
\noindent\textbf{Image order.}
\begin{itemize}
    \item Input image: the initial layout with the highlighted target block.
    \item Output images 1 to $K$: the layout after each successive slide.
\end{itemize}

\noindent
For each output image, write 2 to 3 sentences of reasoning that can be placed before that image:
\begin{enumerate}
    \item Identify the next block that should slide into the empty space and explain how this move contributes to the solution.
    \item State the layout that this move should produce, with all other blocks unchanged.
    \item Phrase the step prospectively as what should happen next, rather than as a caption describing an image that has already been produced.
\end{enumerate}

\noindent\textbf{Output format.}

\noindent
Step 1: \textless reasoning before output image 1\textgreater

\noindent
\ldots

\noindent
Step $K$: \textless reasoning before output image $K$\textgreater
\end{promptbox}

\subsection{Visual Complexity Measurements}
\label{app:visual_complexity}

\cref{sec:dataset:analysis} reports that the 150 tasks newly designed for
\name{} are visually richer than the 150 tasks reworked from VBVR. Here, we provide the measurements supporting this claim. Because there is no single accepted definition of image complexity, we consider four complementary measures. Our conclusion is based on their consistent ordering of the two groups rather than on any individual measure.

\paragraph{Protocol.}
All measurements are computed on the question frame (\texttt{first\_frame.png}), which is provided to the model as input. For each of the 300 tasks, we select 50 instances at evenly spaced positions in the instance list. We average each measure across these instances to obtain one task-level value, yielding 150 values for the reworked tasks and 150 for the newly designed tasks. We then compare the distributions of these task-level values. This treats the task, rather than the individual instance, as the unit of analysis and avoids regarding multiple instances generated by the same task as independent observations.

\paragraph{Measures.}
We use four complementary measures, each capturing a different aspect of visual complexity:
\begin{itemize}
  \item \textbf{Distinct regions.}
        We quantise each frame to a 16-colour palette using median-cut quantisation without dithering. For each palette colour, we identify 8-connected components and retain those occupying at least $10^{-4}$ of the frame area. We remove the largest connected component, which corresponds to the background in our renderings, and count the remaining components. This is the measure reported in the main text.

  \item \textbf{Distinct colours.}
        We quantise each RGB channel to five bits and count the resulting colours that occupy at least $10^{-4}$ of the frame pixels.

  \item \textbf{Spatial information (SI).}
        We compute the standard deviation over pixels of the Sobel gradient  magnitude of the luminance plane, corresponding to the frame-level SI measure defined in ITU-T~P.910~\cite{itu-p910}.

  \item \textbf{Encoded size.}
        We re-encode each frame as JPEG at quality 90 and report the resulting file size in kilobytes. The same encoder and encoding settings are used for all frames.
\end{itemize}

For the first two measures, the minimum area thresholds are normalised by frame size, and the measures are computed at the native rendering resolution. Because SI and encoded size are sensitive to image resolution, we compute both after resizing each frame to $512\times512$.

\paragraph{Results.}
Table~\ref{tab:visual_complexity} reports the results. The newly designed tasks have higher means and medians under all four measures. Based on the median task-level values, they contain $6.9\times$ as many distinct regions and $2.4\times$ as many distinct colours as the reworked tasks, while exhibiting $1.7\times$ higher spatial information and a $2.0\times$ larger encoded size. The consistent ordering across these complementary measures supports the conclusion that the newly designed tasks are visually richer than the reworked tasks.

\begin{table}[th]
\centering
\small
\caption{Visual complexity of the question frames. For each task, each measure is first averaged over 50 instances; the table reports statistics over 150 task-level values per group. The ratio divides the median for the newly designed tasks by that for the reworked tasks.}

\begin{tabular}{@{}lrrrrr@{}}
\toprule
 & \multicolumn{2}{c}{Reworked}
 & \multicolumn{2}{c}{Newly designed} & \\
\cmidrule(lr){2-3}\cmidrule(lr){4-5}
Measure & Mean & Median & Mean & Median & Ratio \\
\midrule
Distinct regions      & 25.6 & 11.6 & 171.4 & 80.4 & 6.9$\times$ \\
Distinct colours      &  6.7 &  5.3 &  29.3 & 12.7 & 2.4$\times$ \\
Spatial information   & 79.8 & 68.8 & 134.3 & 119.0 & 1.7$\times$ \\
Encoded size (KB)     & 14.4 & 12.1 &  28.4 & 24.7 & 2.0$\times$ \\
\bottomrule
\end{tabular}
\label{tab:visual_complexity}
\end{table}

\subsection{Reasoning Depth: Blinded Pairwise Judgement}
\label{app:reasoning_depth}

\cref{sec:dataset:analysis} reports that the new half of \name{} requires deeper reasoning than the reworked half. Unlike visual complexity, reasoning depth cannot be measured directly from a rendered frame. We therefore assess it through a blinded comparison from human annotation.  We additionally conduct a binary per-task assessment that distinguishes direct single-rule application from multi-step reasoning.

\paragraph{Protocol.}
Each of the 150 reworked tasks is paired with one of the 150 new tasks using a fixed random one-to-one matching. GPT-5.5 receives only the starting frame and instruction for each task; task origins and group labels are withheld. Each pair is evaluated twice, with the two tasks exchanged between the first and second slots. This controls for positional preference and allows us to measure its magnitude.

For the pairwise assessment, annotator is asked to select the task that requires more reasoning before its answer can be determined. A pair is counted as consistent only when the same task is selected under both orderings. In a separate per-task assessment, annotator classifies each task according to whether its answer follows from a direct application of a single rule to the starting frame or instead requires multiple dependent inference steps, such as comparing alternatives, searching over possibilities, or eliminating candidate answers.

\paragraph{Results.}
The first slot was selected in 45.8\% of pairwise judgements with a decisive outcome, indicating little systematic position bias. The two orderings selected the same task for 133 of the 150 pairs (88.7\%). Among all 150 pairs, the new task was selected as requiring more reasoning under both orderings in 113 cases (75.3\%), whereas the reworked task was selected under both orderings in 20 cases (13.3\%); the remaining 17 pairs (11.3\%) produced inconsistent outcomes. In the per-task assessment, 10 of the 150 reworked tasks (6.7\%) were classified as requiring multi-step reasoning, compared with 70 of the 150 new tasks (46.7\%). These results show that the new tasks are more likely to require both greater relative reasoning depth and multiple dependent inference steps than the reworked tasks.
\section{Details of Verifiable Scorer}
\label{app:details_of_scorer}

\subsection{Human Preference Alignment}
\label{app:human_preference_alignmrnt}

\paragraph{Annotation Protocol}
We evaluate eight video generation models on \name-Bench. To keep annotation
tractable, we sample one instance per task and form all $\binom{8}{2}=28$ model pairs per task. Each pair is shown to ten independent annotators, who rate the absolute quality of both videos on a four-level scale: $L_0$ denotes a fully correct and faithful result, $L_1$ minor errors that do not affect task success, $L_2$ major errors while still attempting the task, and $L_3$ a complete failure in which the task is not accomplished. Annotators judge each video independently, so a
pair receives ten level ratings per video.

\paragraph{From Ratings to Pairwise Preferences}
For each annotation vote, we convert the two level ratings into one of four outcomes: \emph{$A$ wins}, \emph{$B$ wins}, a \emph{tie above the failure level}, or a \emph{tie at the failure level}. If the two videos receive different levels, the one with the better level wins. If they receive the same level, the vote is a tie-above-failure when both are rated $L_0$--$L_2$, and a tie-at-failure when both
are rated $L_3$. Aggregating the ten votes for a pair yields a human preference vector $\mathbf{v}=[\,n_A,n_{\text{tie-above}},n_{\text{tie-fail}},n_B\,]$ with $\sum_i v_i = 10$. Splitting ties is important because the tasks are difficult and most generated videos fall at $L_3$; collapsing all ties into one class would conflate pairs in which both videos fail with pairs in which both are of high
quality.

\paragraph{Mapping Evaluator Scores to Preferences} 
Each evaluator (our scorer or a VLM judge) produces a scalar quality score in $[0,1]$ per video. For a pair with scores $s_1,s_2$, if $|s_1-s_2|\le\theta$ (with $\theta=0.05$) we declare a tie and assign it to tie-above- or tie-at-failure according to whether the mean $(s_1+s_2)/2$ exceeds $0.5$, matching the $L_2$--$L_3$ boundary of the human protocol; otherwise the higher-scoring video wins. The same mapping is applied to every evaluator, ensuring a consistent comparison protocol.

\paragraph{Agreement Metric}
We score an evaluator by how well its single predicted outcome agrees with the human vote vector of each pair. Let $c$ be the outcome predicted by the evaluator; the per-pair agreement is $v_c/10$, i.e., the fraction of the ten annotators whose vote matches the prediction. We report the mean of this quantity over all evaluated pairs (``per-vote agreement'').

\paragraph{Cost Estimation}
For proprietary APIs, cost per evaluation is computed from token usage and the provider's price. For open-source judges, it is the total wall-clock evaluation time on a self-hosted Lambda H100~SXM node multiplied by its rental price. Our scorer is a lightweight rule-based computation requiring no API calls and only minimal compute (optionally a GPU for OCR) is required. However, for a conservative estimate, we charge its entire wall-clock runtime at the same H100 node rental rate used for the open-source judges. Its reported cost therefore represents an upper bound.

\subsection{Detailed Scoring Logic for Selected Tasks}
\label{app:detailed_scoring_logic}

For each selected task, we present its prompt and ground-truth key frames, followed by the corresponding scoring logic. These examples illustrate how the task-specific scorers jointly assess the correctness of the final state, the validity of the intermediate process, and the preservation of task-irrelevant scene content.

\paragraph{Greedy Ball Eating (Task O-31)}
\begin{figure}[t]
\centering
\includegraphics[width=\linewidth]{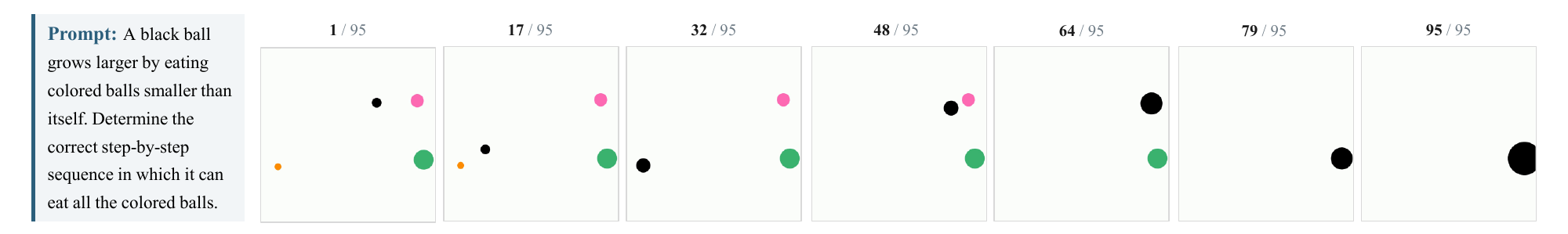}
\caption{Greedy Ball Eating (Task O-31): prompt and ground-truth frames. The black ball absorbs the colored balls one at a time in ascending size order until only the grown black ball remains.}
\label{fig:app_o31}
\end{figure}
As shown in Figure~\ref{fig:app_o31}, a black ball must absorb the smaller colored balls one at a time in ascending size order until only the grown black ball remains. From the first frame, we detect the black ball and each colored target $i$ via HSV segmentation and contour analysis, recording its area $A_i$. Across frames, we track a presence ratio $\rho_i(t)\in[0,1]$ and register an \emph{eat} event for target $i$ at the frame where $\rho_i$ drops from ${\approx}1$ to ${\approx}0$ while the target overlaps the dilated black-ball mask. The final score equally weights a final-state term and an eating-process term:
\begin{equation}
S = 0.5\cdot S_{\text{final}} + 0.5\cdot S_{\text{eat}}.
\end{equation}
\emph{Final-state score.} We first compute a size-match score $b\in[0,1]$ that measures how closely the final black ball's area matches its expected grown size, i.e., the total area of all absorbed balls. The last-frame score is then gated by $b$:
\begin{equation}
S_{\text{final}} = b\cdot\big(0.5 + 0.5\cdot(0.5\cdot q_{\text{tgt}} + 0.5\cdot q_{\text{clean}})\big),
\end{equation}
where $q_{\text{tgt}}=\max(0,\,1-r/N)$ penalizes the $r$ colored balls still present among the $N$ targets, and $q_{\text{clean}}$ penalizes stray colored pixels. Because $b$ multiplies the entire term, a missing or incorrectly sized black ball collapses the score. \emph{Process score.} Each ground-truth eat event is evaluated using four checks:
\begin{equation}
S_{\text{eat}} = \tfrac{1}{N}\sum_{k=1}^{N}(0.25\cdot o_k + 0.25\cdot a_k + 0.30\cdot g_k + 0.20\cdot s_k),
\end{equation}
where $o_k$ rewards the correct size rank, $a_k$ the black ball's arrival at and contact with the target, $g_k$ its growth by the expected factor, and $s_k$ the stillness of the not-yet-eaten balls. A target that is never eaten contributes $0$.

\paragraph{Ball-Cluster Merging (Task O-29)}
\begin{figure}[t]
\centering
\includegraphics[width=\linewidth]{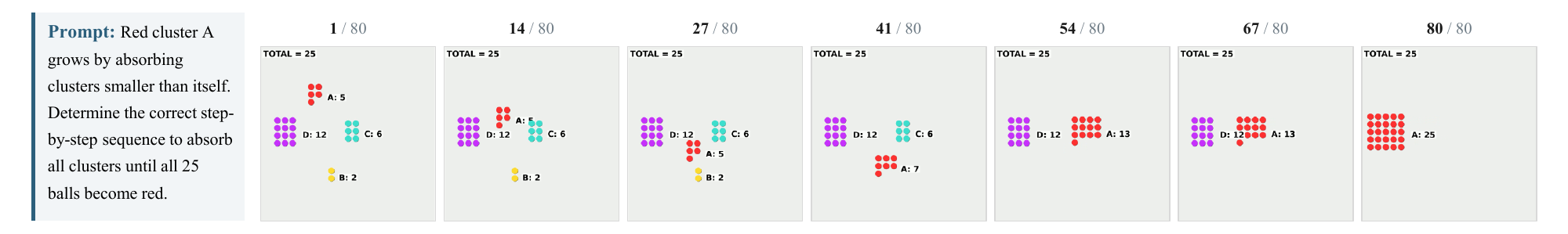}
\caption{Ball-Cluster Merging (Task O-29): prompt and ground-truth frames. A survivor cluster sequentially absorbs the other clusters until only it remains, and its label must show the total count.}
\label{fig:app_o29}
\end{figure}
As illustrated in Figure~\ref{fig:app_o29}, a survivor cluster sequentially absorbs the other colored clusters until only it remains, and its label must show the total count. We track each cluster's pixel presence across frames to detect merge events and use OCR to read the final count. The score is a weighted sum of a final-state and a merge-process term, which is then controlled by a process gate $g_\text{proc}=0.2 + 0.8\cdot S_\text{proc}$:
\begin{equation}
S = (0.6\cdot S_\text{final} + 0.4\cdot S_\text{proc})\cdot g_\text{proc}.
\end{equation}
so a poor merging process suppresses the overall score. \emph{Final-state score.} The last frame is evaluated in terms of the survivor count, cluster compactness, and scene cleanliness:
\begin{equation}
S_\text{final} = (0.5\cdot q_\text{surv} + 0.5\cdot q_\text{cluster})\cdot(0.6 + 0.4\cdot q_\text{pen})\cdot(0.8 + 0.2\cdot q_\text{text}),
\end{equation}
where $q_\text{surv}$ measures whether the survivor count is correct, $q_\text{cluster}$ whether the survivors form a single compact group, $q_\text{pen}=0.5\cdot q_\text{gone}+0.5\cdot q_\text{clean}$ whether the other colors have disappeared without leaving stray pixels, and $q_\text{text}$ whether the label matches the total count. \emph{Merge-process score.} The process score combines the average quality of individual merge events with a survivor-stability term:
\begin{equation}
S_\text{proc} = 0.7\cdot\tfrac{1}{M}\sum_{k=1}^{M}(0.2\cdot o_k + 0.3\cdot a_k + 0.3\cdot c_k + 0.2\cdot s_k) + 0.3\cdot q_\text{stab},
\end{equation}
where $o_k$, $a_k$, $c_k$, and $s_k$ score the order, arrival, resulting count, and stillness of the $k$-th merge, respectively, and $q_\text{stab}$ rewards a stable survivor count between consecutive merges.

\paragraph{Key-Door Navigation (Task G-45)}
\begin{figure}[t]
\centering
\includegraphics[width=\linewidth]{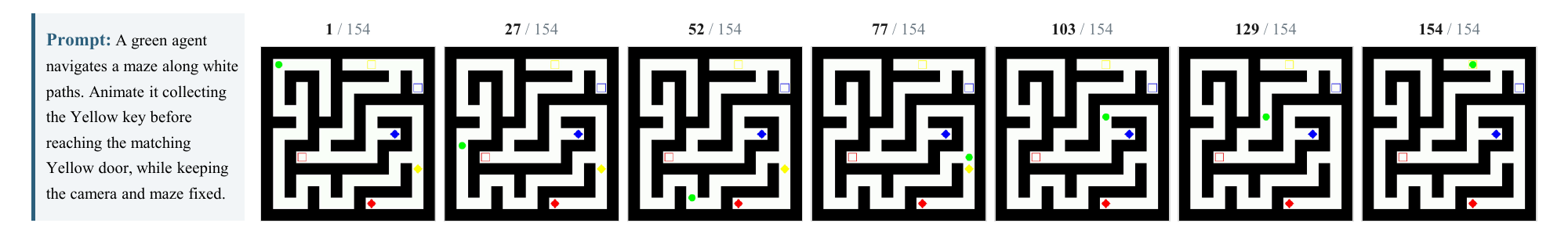}
\caption{Key-Door Navigation (Task G-45): prompt and ground-truth frames. A colored agent must reach the key cell and then the door cell without crossing walls.}
\label{fig:app_g45}
\end{figure}
As shown in Figure~\ref{fig:app_g45}, a colored agent must move from its starting cell to the key cell and then to the door cell without crossing walls. We track the agent's per-frame centroid using color segmentation, reconstruct its path, detect wall crossings, and determine whether the key was visited. The score multiplies a trajectory term by two constraint gates:
\begin{equation}
S = \underbrace{\tfrac{1}{3}(\text{proximity}+\text{continuity}+\text{coverage})}_{\text{trajectory}}\cdot(0.4 + 0.6\cdot g_\text{wall})\cdot(0.4 + 0.6\cdot g_\text{key}).
\end{equation}
\emph{Trajectory score.} \emph{proximity} measures closeness to the optimal path, \emph{coverage} measures the fraction of that path traversed, and \emph{continuity} penalizes teleportation.
\emph{Constraint gates.} $g_\text{wall}=\max(0.4,\,1-0.15\cdot n_\text{hits})$ penalizes the $n_\text{hits}$ wall cells entered, while $g_\text{key}=1$ if the key is reached and $0.25$ otherwise. Each gate is linearly mapped into $[0.4,1]$ via $0.4+0.6\cdot g$, so violating either constraint sharply reduces the score; for example, skipping the key caps its multiplier at $0.4+0.6\cdot0.25=0.55$.

\paragraph{Rolling Ball (Task O-32)}
\begin{figure}[t]
\centering
\includegraphics[width=\linewidth]{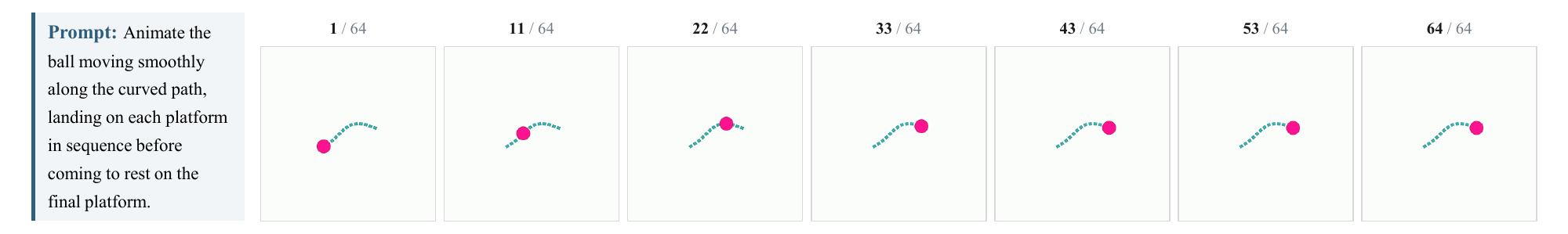}
\caption{Rolling Ball (Task O-32): prompt and ground-truth frames. A ball rolls along the fixed curved path to the target.}
\label{fig:app_o32}
\end{figure}
Figure~\ref{fig:app_o32} shows a ball rolling along a fixed curved path toward a target. We detect the ball in each frame and measure the fraction of frames in which the ball lies on the path. The score gates a completion term by a scene-consistency term:
\begin{equation}
S = S_\text{comp}\cdot(0.6 + 0.4\cdot q_\text{cons}).
\end{equation}
\emph{Completion score.} Final-position accuracy is gated by a path-following term:
\begin{equation}
S_\text{comp} = q_\text{pos}\cdot(0.2 + 0.8\cdot p),\qquad p = q_\text{on\text{-}path}\cdot q_\text{dir}\cdot q_\text{mid},
\end{equation}
where $q_\text{pos}$ rewards ending near the target, and $p$ is the product of $q_\text{on\text{-}path}$ (the fraction of frames in which the ball lies on the path), $q_\text{dir}$ (motion in the correct forward direction), and $q_\text{mid}$ (genuine intermediate motion). Because these factors are multiplied, a ball that rarely follows the path receives a low $S_\text{comp}$ even if it ends near the target. \emph{Consistency gate.} $q_\text{cons}$ averages scene- and background-preservation scores, scaling the completion score by a factor in $[0.6,1]$.

\paragraph{High-Density Liquid (Task G-273)}
\begin{figure}[t]
\centering
\includegraphics[width=\linewidth]{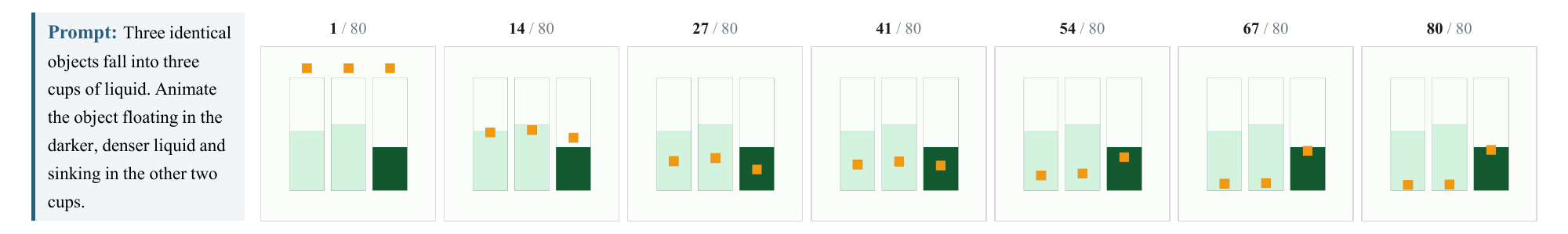}
\caption{High-Density Liquid (Task G-273): prompt and ground-truth frames. Three objects fall into cups; the object above the denser liquid should float rather than sink.}
\label{fig:app_g273}
\end{figure}
As illustrated in Figure~\ref{fig:app_g273}, three identical objects fall into three cups, one of which contains a denser liquid and should therefore cause its object to float rather than sink. We track each object's descent, classify its final state as floating or sinking based on its resting height, and check the liquid levels and background. The score is a weighted sum of final-state and process terms:
\begin{equation}
S = 0.6\cdot S_\text{final} + 0.4\cdot S_\text{proc}.
\end{equation}
\emph{Final-state score.} The final object positions are evaluated together with scene preservation:
\begin{equation}
S_\text{final} = q_\text{pos}\cdot(0.5 + 0.5\cdot q_\text{scene}),
\end{equation}
where $q_\text{pos}$ rewards each object for resting at its correct floating or sinking height, and $q_\text{scene}$ measures whether the cups, liquid, and background are preserved. \emph{Process score.} The descent must remain physically plausible:
\begin{equation}
S_\text{proc} = q_\text{ntp}\cdot(0.1 + 0.3\cdot q_\text{opc} + 0.3\cdot q_\text{stab} + 0.3\cdot q_\text{area}),
\end{equation}
where $q_\text{ntp}$ requires non-teleporting descent, $q_\text{opc}$ checks that each cup column contains at most one object, $q_\text{stab}$ measures the integrity of the intermediate frames, and $q_\text{area}$ rewards consistent object size throughout the video.

\paragraph{Move Objects to Targets (Task O-27)}
\begin{figure}[t]
\centering
\includegraphics[width=\linewidth]{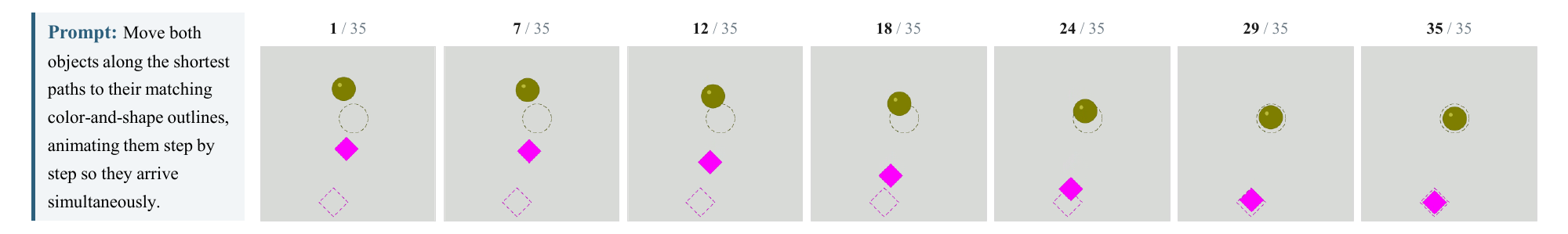}
\caption{Move Objects to Targets (Task O-27): prompt and ground-truth frames. Two objects move to their corresponding target outlines.}
\label{fig:app_o27}
\end{figure}
As shown in Figure~\ref{fig:app_o27}, two objects must be moved to their corresponding target outlines. We track each object's trajectory and evaluate its transport before checking whether the two objects move concurrently. The score gates the mean transport quality by a synchronization term:
\begin{equation}
S = S_\text{mov}\cdot(0.8 + 0.2\cdot q_\text{sync}).
\end{equation}
\emph{Transport score.} For each object, transport quality is computed as a product:
\begin{equation}
S_\text{mov} = \tfrac{1}{2}\sum_{i=1}^{2} q^{(i)}_\text{trans}\cdot q^{(i)}_\text{vis}\cdot q^{(i)}_\text{tp}\cdot q^{(i)}_\text{dur},
\end{equation}
where $q_\text{trans}$ combines path quality and endpoint accuracy, $q_\text{vis}$ is the fraction of frames in which the object is successfully tracked, $q_\text{tp}$ penalizes teleportation, and $q_\text{dur}$ rewards a realistic motion duration. \emph{Synchronization.} $q_\text{sync}=\sqrt{q_\text{start}\cdot q_\text{end}}$ rewards the two objects for starting and finishing at similar times. Moving them sequentially drives $q_\text{sync}\to0$ and limits the score to $0.8\cdot\text{movement}$.

\end{document}